\documentclass{article}
\usepackage[final, nonatbib]{neurips_2022}

\newcommand{\etal}{\emph{et al.}}
\usepackage{pgfplots} %
\pgfplotsset{every tick label/.append style={font=\scriptsize}}
\usepackage{libertine}
\usepackage[libertine]{newtxmath}
\usepackage{DejaVuSansMono}
\usepackage[format=plain,
            labelfont={bf},
            textfont=sf]{caption}

\definecolor{bg}{rgb}{0.95,0.95,0.95}

\usepackage{soul}
\sethlcolor{yellow}

\usepackage{inconsolata}
\usepackage{float}
\usepackage{titletoc}
\usepackage{graphicx}
\usepackage{colortbl}
\usetikzlibrary{patterns}

\DeclareFontFamily{U}{mathb}{\hyphenchar\font45}
\DeclareFontShape{U}{mathb}{m}{n}{
      <5> <6> <7> <8> <9> <10> gen * mathb
      <10.95> mathb10 <12> <14.4> <17.28> <20.74> <24.88> mathb12
      }{}
\DeclareSymbolFont{mathb}{U}{mathb}{m}{n}
\DeclareFontSubstitution{U}{mathb}{m}{n}

\let\dot\relax
\DeclareMathAccent{\dot}{0}{mathb}{"39}
\let\ddot\relax
\DeclareMathAccent{\ddot}{0}{mathb}{"3A}
\let\dddot\relax
\DeclareMathAccent{\dddot}{0}{mathb}{"3B}
\let\ddddot\relax
\DeclareMathAccent{\ddddot}{0}{mathb}{"3C}

\usepackage{hyperref}

\makeatletter
\let\NAT@parse\undefined
\makeatother
\usepackage[sort&compress, numbers]{natbib}
\usepackage{tabularx}
\usepackage{multicol}
\usepackage{multirow}
\usepackage{booktabs}
\usepackage{wrapfig}
\usepackage{listings}
\tikzset{every picture/.style={/utils/exec={\sffamily}}}

\definecolor{codegreen}{rgb}{0,0.6,0}
\definecolor{codegray}{rgb}{0.5,0.5,0.5}
\definecolor{codepurple}{rgb}{0.58,0,0.82}
\definecolor{backcolour}{rgb}{0.95,0.95,0.92}

\lstdefinestyle{mystyle}{
    backgroundcolor=\color{backcolour},   
    commentstyle=\color{codegreen},
    keywordstyle=\color{magenta},
    numberstyle=\tiny\color{codegray},
    stringstyle=\color{codepurple},
    basicstyle=\ttfamily\tiny,
    breakatwhitespace=false,         
    breaklines=true,                 
    captionpos=b,                    
    keepspaces=true,                 
    numbers=left,                    
    numbersep=5pt,                  
    showspaces=false,                
    showstringspaces=false,
    showtabs=false,                  
    tabsize=2
}

\definecolor{Gray}{gray}{0.9}
\definecolor{bargreen}{HTML}{9fffcb}
\definecolor{bargreenline}{HTML}{25a18e}
\definecolor{barorange}{HTML}{888888}
\definecolor{barorangeline}{HTML}{888888}
\definecolor{bargreenlight4}{HTML}{bbffda}
\definecolor{bargreenlight2}{HTML}{d5ffe8}
\definecolor{bargreenlight3}{HTML}{e2ffef}
\definecolor{bargreenlight}{HTML}{ceffff}

\usepackage[color=green!20, textsize=footnotesize]{todonotes} 

\definecolor{commentcolor}{HTML}{00e268}%
\newcommand{\pn}{$98^{\text{th}}$}
\newcommand{\ps}{$75^{\text{th}}$}

\definecolor{urlcolor}{rgb}{0,0.0,0.40}
\newcommand\numberthis[1]{\addtocounter{equation}{1}\tag{\theequation}\label{#1}}

\hypersetup{
    colorlinks=true,
    citecolor=black,
    filecolor=black,
    linkcolor=urlcolor,
    urlcolor=urlcolor,
}  
\newcommand{\algokw}{
    \SetKwInOut{KwInit}{Init}
\SetKwInOut{KwIn}{Param}
\SetKwInOut{KwProb}{Data}
}

\newcommand{\forloop}[3]{\upshape(#1 = #2;\ #1$<$ #3;\ #1++)}

\SetKwComment{tcp}{$\triangleright$\ }{}
\SetCommentSty{mycommfont}
\DeclareMathOperator*{\argmin}{arg\,min}

\newcommand{\website}{\href{https://curobo.org}{curobo.org}}

\usepackage{fancyhdr}
\usepackage{enumitem}

\title{{\LARGE\textbf{cuRobo:}} \\ {\Large \raggedright \textsf{Parallelized Collision-Free Minimum-Jerk Robot Motion Generation}}}

\author{
Balakumar Sundaralingam\footnotemark[1]
\quad
Siva Kumar Sastry Hari\thanks{Equal Contribution.}
\quad
Adam Fishman
\quad
Caelan Garrett
\And
Karl Van Wyk
\quad
Valts Blukis
\quad
Alexander Millane
\quad
Helen Oleynikova
\And
Ankur Handa
\quad
Fabio Ramos
\quad
Nathan Ratliff
\quad
Dieter Fox 
\\[12pt]
{\fontsize{12}{11}\textbf{\textsf{NVIDIA}}}
}
\pgfplotsset{compat=1.18}

\begin{document}

\makeatletter
\makeatother 
\maketitle

\begin{figure}[h]
    \centering
    \includegraphics[width=\textwidth,trim={0.1cm 0cm 0.1cm 0cm}, clip]{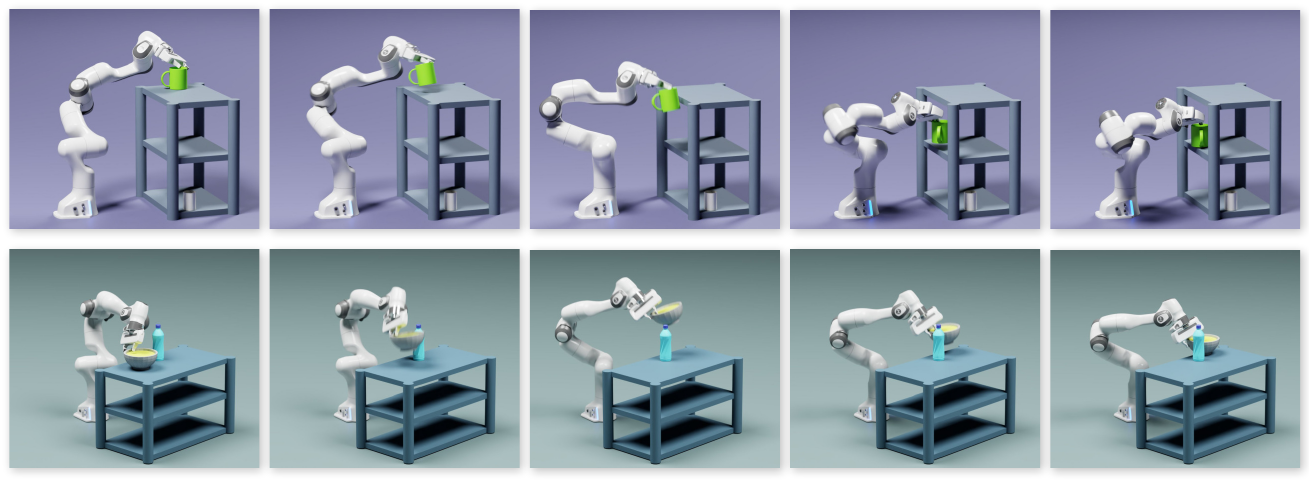}
    \caption{We present an approach to perform collision-free motion generation for a manipulator to move from a start configuration to a desired gripper pose. Few instances of using our approach is shown here, where the Franka Panda robot moves a grasped object around obstacles to place in a target location. We leverage parallel computing to generate motions within 30ms.}
    \label{fig:enter-label}
\end{figure}

\begin{abstract}
  This paper explores the problem of collision-free motion generation for manipulators by formulating it as a global motion optimization problem. We develop a parallel optimization technique to solve this problem  and demonstrate its effectiveness on massively parallel GPUs. We show that combining simple optimization techniques with many parallel seeds leads to solving difficult motion generation problems within 50ms on average, 60$\times$ faster than state-of-the-art (SOTA) trajectory optimization methods. We achieve SOTA performance by combining L-BFGS step direction estimation with a novel parallel noisy line search scheme and a particle-based optimization solver. To further aid trajectory optimization, we develop a parallel geometric planner that plans within 20ms and also introduce a collision-free IK solver that can solve over 7000 queries/s. We package our contributions into a state of the art GPU accelerated motion generation library,~\emph{cuRobo} and release it to enrich the robotics community. Additional details are available at \website.

\end{abstract}
\clearpage
\tableofcontents

\clearpage
\listofalgorithms
\listoftables
\clearpage
\section{Introduction}
\label{sec:intro}

Safe navigation is fundamental to robotics~\cite{medeiros2020trajectory}, requiring robots to have a robust global motion generation system to traverse any environment structure encountered at deployment. Motion generation for high-dimensional systems is extremely challenging as satisfying complex constraints and minimizing cost terms in a very large C-Space is computationally expensive. Manipulators, for instance, can have many articulations, complex link geometries, entire goal regions beyond a single configuration, task constraints, and nontrivial kinematic and torque limitations. There has been a long history of problem decomposition in this field to mitigate complexity, leading to standard approaches that often first plan collision-free geometric paths \cite{lavalle1998rapidly,lavalle2006PlanningAlgorithms} and then smooth those paths for dynamic efficiency \cite{lavalle2006PlanningAlgorithms,kunz2012time}. But increasingly, research into the interconnections between optimization and planning \citep{kunz2014probabilistically, toussaint2009trajopt, ratliff2009chomp, kalakrishnan2011stomp, schulman2014motion, toussaint2014komo, ratliff2015riemo, mukadam2018continuous} has shown that optimization can be a powerful tool well beyond trajectory smoothing, and trajectory optimization alone now has a breadth of applications \citep{posa2013direct,toussaint2015logic,apgar2018fast,sundaralingam2019relaxed}. Our modern understanding of this robot navigation problem is that it is a large \emph{global} motion optimization problem \cite{goldberg2020gomp,bertsekas2019RLAndOptControl}.

The global optimization literature suggests that finding the true global minimum is usually impractical, but strategies for robustly finding high-performing local minima can be effective \cite{hansen2016cmaTutorial}. Many strategies follow the simple pattern of selecting many seed candidates and performing a local optimization for each. This sample and optimize process can often realize substantial gains by leveraging distributed computation. However, most motion generation systems today remain sequential and slow, following a CPU-based design. State-of-the-art motion generation solutions take 0.5s to 10s depending on the task's complexity on modern CPUs~\cite{tesseract}. This run-time is even slower on edge devices that operate under limited power budgets. 
This slow and sequential process has resulted in pipelined systems that compute only a single best candidate seed, which is then passed to an optimizer for local optimization~\cite{lavalle2006PlanningAlgorithms}. Such systems fundamentally limit their ability to find better local optima by betting on a single seed. 

The insights used to improve the speed and quality of the solution for global optimization problems may apply well to the problem of global motion generation. In this work, we present a collection of techniques and implementations that leverage parallel processing to accelerate motion planning and optimization, and for running many optimization instances in parallel to robustly address these global optimization problems. Existing literature supports these algorithmic principles and has shown that (1) the heuristic initialization for the problem can be effective~\citep{schulman2014motion}, and (2) many restarts with randomized noise of the initial seed can dramatically improve performance~\citep{zucker2013chomp}. 

In the realm of global motion generation, massively parallel compute is already being used to accelerate Probabilistic Road Map (PRM) pruning by using an FPGA with special circuits, leading to many orders of magnitude speedup~\citep{konidaris2016fpga}. However, instead of designing special circuits for motion generation, we leverage the massively parallel compute available on graphical processing units (GPUs). GPUs have become pervasive in both high- and low-powered configurations as they offer energy-efficient and high-throughput computation platforms, an important requirement for solving parallelizable compute intensive problems.  We show how GPUs also offer programmability and flexibility to map sophisticated computations of motion optimization to hardware, allowing us to parallelize the entire motion generation pipeline. We achieve high speedups using GPUs compared to serial implementations in motion generation. While we demonstrate the benefits of using parallel compute for motion generation using NVIDIA GPUs, the approach is applicable to other parallel architectures.

Our effort to solve global motion generation starts with parallelizing the core blocks in a robotics stack -- robot kinematics, robot self signed distance (i.e., between a robot's links), and robot world signed distance~(i.e., world represented by cuboids, meshes, and a depth camera stream). We formalize these functions to use many threads per query, implement them efficiently in CUDA, and provide them as differentiable functions in pyTorch, enabling others to also use these functions as the backbone for their own robotic tasks. We then formulate a continuous collision checking algorithm that only requires a point signed distance function from the world representation. We then introduce parallel algorithms for numerical optimization and geometric planning, that aid in solving global motion generation. Our main contributions are summarized as follows: \\
\noindent\textbf{Performant Kinematics and Signed Distance Kernels:} We develop high-performance CUDA kernels for robot kinematics and signed distance computation which are up to 10,000x faster than existing CPU based methods.\\
\noindent\textbf{Differentiable Continuous Collision Checking:} Formulate continuous collision checking algorithm that only needs a point signed distance function (and closest point for gradients) to perform swept collision checks, enabling use across different world representations from primitives and meshes to occupancy maps.\\
\noindent\textbf{Parallel Optimization:} We develop a GPU batched L-BFGS optimizer, that uses an approximate parallel line search scheme, and a particle-based optimizer to solve difficult motion generation problems. Our solver is 23$\times$, 80$\times$, and 87$\times$ faster for inverse kinematics, collision-free inverse kinematics, and collision-free trajectory optimization respectively when compared to existing CPU based solvers.\\
\noindent \textbf{SOTA IK Solver:} Leveraging our performant kernels, we have developed a world-leading inverse kinematic solver, that can solve 37000 IK problems per second~(23$\times$ faster than TracIK~\cite{beeson2015trac}) and also solve 7600 collision-free IK problems per second~(80$\times$ faster than using TracIK + Bullet~\cite{coumans2021}).\\ 
\noindent \textbf{Parallel Geometric Planner:} We develop a geometric planner with a parallel steering algorithm to generate collision-free paths within 20 ms on a modern desktop machine with NVIDIA RTX 4090 and AMD Ryzen 9 7950x. \\
\noindent \textbf{Global Motion Generation:} Combining our above contributions, we have a global motion generation pipeline that can plan within 50ms, 60$\times$ faster than existing methods~(Tesseract). 
\\
\noindent \textbf{Validation on a Low-Power Device:} We evaluate our GPU-accelerated motion generation stack and existing CPU-based methods on an NVIDIA Jetson AGX Orin at different power budgets. Results show that our approach is 28$\times$ and 21$\times$ faster on average for motion generation problems when the device was set to 60W and 15W budgets, respectively.\\
\noindent \textbf{cuRobo Library:} We developed \emph{cuRobo}, a suite of GPU-accelerated robotics algorithms, providing SOTA implementations of robot kinematics, signed distance functions, optimization solvers, geometric planning, trajectory optimization, and model predictive control. We are releasing this library to enrich roboticists with the necessary tools to explore large-scale problems in robotics.

\section{Motion Generation as Optimization}
\label{sec:mopt}
We define the problem of motion generation as the task of moving from an initial joint configuration~$\theta_0$ to a final joint configuration~$\theta_T$, at which state a task cost~$C(\theta_T)$ is below a desired threshold. Additionally, the transition states from~$\theta_0$ to~$\theta_T$ must also satisfy system constraints. In this work, we focus on the task of collision-free motion generation to reach a goal Cartesian pose~$X_g\in\mathbb{SE}(3)$ with the robot's end-effector. Specifically, we want to obtain a joint-space trajectory~$\theta_{[0,T]}$ that satisfies the robot's joint limits~(position, velocity, acceleration, jerk), doesn't collide with itself or the environment, and reaches the goal pose~$X_g$ by the last timestep~$T$.

We formulate this continuous-time motion problem as a time discretized trajectory optimization problem,
    \begin{align*}
        \argmin_{\theta_{[1,T]}} \quad & C_{\text{task}}(X_g,\theta_T) + \sum_{t=1}^T C_{\text{smooth}}(\cdot) \numberthis{eq:cost_motion_opt}\\
    \text{s.t.,} \quad & 
    \theta^- \preceq \theta_t \preceq \theta^+, \forall t \in [1,T]\numberthis{eq:jl}\\
    & \dot{\theta}^- \preceq \dot{\theta}_t \preceq \dot{\theta}^+, \forall t \in [1,T] \numberthis{eq:jv}\\
    & \ddot{\theta}^-\preceq \ddot{\theta}_t \preceq \ddot{\theta}^+, \forall t \in [1,T] 
    \numberthis{eq:ja}\\
    & \dddot{\theta}^-\preceq \dddot{\theta}_t \preceq \dddot{\theta}^+, \forall t \in [1,T] 
    \numberthis{eq:jj}\\
    &\dot{\theta}_{T},\ddot{\theta}_{T},\dddot{\theta}_{T}  = 0 \numberthis{eq:vzero}\\
    & C_r(K_s(\theta_t)) \preceq 0, \forall t \in [1,T] \numberthis{eq:self-cost}\\
    &C_w(K_s(\theta_t)) \preceq 0, \forall t \in [1,T] \numberthis{eq:world-cost}
    \end{align*}
where $C_{smooth}(\cdot)$ is a cost term that encourages smooth robot behavior. Joint limit constraints are enabled by~Eq.\ref{eq:jl}-\ref{eq:jj}. We also constrain the robot to have zero velocity, acceleration and jerk at the final timestep by constraints in Eq.~\ref{eq:vzero}. A detailed discussion on this optimization problem and the formulation of cost terms is available in Appendix.~\ref{app:trajopt}. We discuss the collision avoidance constraints~Eq.\ref{eq:self-cost}, and Eq.\ref{eq:world-cost} in Sec.~\ref{sec:collision-avoidance}.

\begin{figure}
    \centering
    \includegraphics[width=\textwidth]{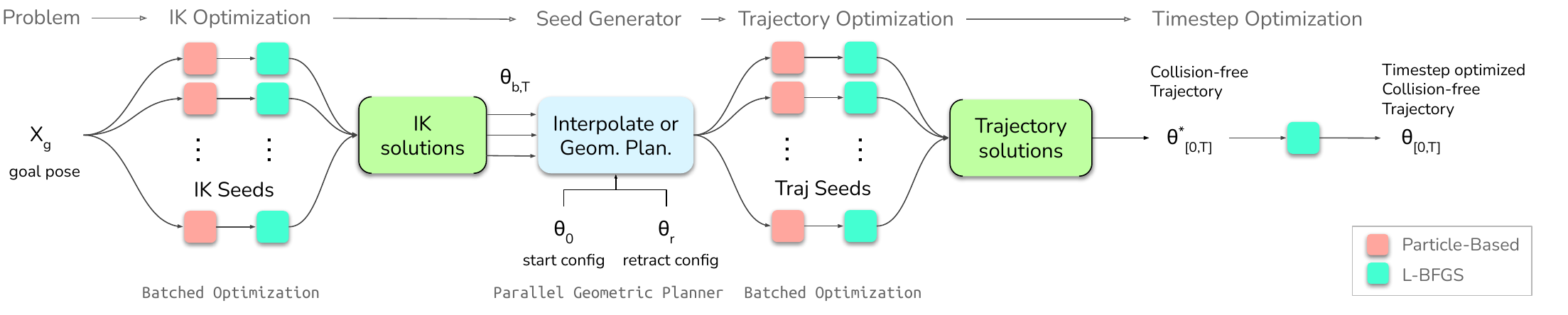}
    \caption{Our approach to global motion generation takes as input a goal pose~$X_g$, initial joint configuration~$\theta_0$, and outputs a timestep optimized, collision-free trajectory~$\theta_[0,T]$. We solve this by first running many instances of collision-free IK, followed by generating seeds for trajectory optimization by either linearly interpolating between start and IK solutions, passing through a retract config, or using our geometric planner. We then run trajectory optimization on many seeds in parallel to obtain a collision-free trajectory~$\theta^*_{[0,T]}$, which is then re-optimized with an estimated~$dt$ to get a timestep optimized collision-free trajectory~$\theta_{[0,T]}$. Our numerical optimization performs a few iterations of particle-based optimization to move the seed to a good region follow by L-BFGS to quickly converge to the minimum.
    }
    \label{fig:approach_outline}
\end{figure}

A good initial seed can speedup convergence in the above defined trajectory optimization problem. One common way~\cite{toussaint2015logic} to initialize the seed is to first optimize only for the terminal joint configuration~$\theta_T$ and then initialize the trajectory with a linear interpolation from the start configuration~$\theta_0$ to the solved terminal configuration (interpolating through a predefined waypoint has also shown to be helpful~\cite{schulman2014motion}).  In our problem setting of reaching a goal pose~$X_g$, the terminal state optimization problem boils down to a collision-free inverse kinematics~(IK) problem containing the pose cost, the collision constraints~Eq.~\ref{eq:world-cost}-Eq.~\ref{eq:self-cost} and the joint limit constraint~Eq.~\ref{eq:jl}. We hence first solve for collision-free IK, followed by seed generation, and then trajectory optimization. Once we run trajectory optimization, we find an optimal dt by scaling the trajectory's velocity, acceleration, or jerk to the robot's limits and rerun trajectory optimization with this new dt to get the final result (see~\ref{app:timeopt}). Our overall approach is illustrated in Figure~\ref{fig:approach_outline}.

\section{Kinematics \& Collision Avoidance}
\label{sec:collision-avoidance}
Collision avoidance is a critical component of motion generation as the robot needs to be able to avoid colliding with itself~(self-collision) and with the world for safe operation. A standard approach to computing collisions involves transforming the robot's geometries (often represented as meshes) based on the current joint configuration~(forward kinematics) and computing mesh-mesh distances~\cite{pan2012gpu,  pan2012fcl, Montaut-RSS-22, tracy2022differentiable, Montaut-RSS-22}. Since we know the geometry of the robot, we can reduce the computation required for collision checking by representing the robot's volume with a set of spheres~\cite{greenspan1996obstacle} as shown in Figure~\ref{fig:app_robot_sphere}. With this sphere representation for the robot, our collision avoidance cost terms only need to check the distance between the origin of each sphere and the world, then subtract the radius to get the sphere distance. Similarly, for self collisions, we only have to compute the distance between pairs of spheres~(i.e. compute point distance and subtract the radii of the two spheres). This enables our approach to scale to low-power edge devices and also accommodate very large batch queries. We discuss some techniques to approximate a mesh with spheres in Appendix~\ref{app:curobo-library}. We will next discuss how we map between the robot joint configuration and the location of the spheres.
\subsection{Robot Kinematics}
Robot kinematics~$K_s(\cdot)$ enables mapping between a robot's joint configuration and the Cartesian pose~($\mathrm{SE(3)}$) of all geometries attached to the robot. This mapping is done by computing a sequence of transformations from the base link of the robot to the different links attached through joints. Each actuated joint adds an additional transformation based on it's value and type. Hence,  
traversing the robot's kinematic tree by design is sequential for serial manipulators. To overcome the sequential nature of computation, we represent the transformations as homogeneous matrices~(4x4), enabling us to use four parallel threads to compute matrix multiplications. Once we build the pose of all links of the robot, we perform matrix vector products to compute the position of the spheres. We also output the pose of the end-effector as a position and quaternion. For computing the backward, we use 16 threads to read and project the gradients from the Cartesian space to the joint space. By using many threads for a single kinematics query, we overcome some of the memory overhead that comes with parallel compute devices. Our kinematics function supports single axis actuation across all three linear and three angular spaces. Extensive details are available in Appendix~\ref{app:kinematics-cuda}. Figure~\ref{fig:app_robot_sphere}-a shows the output of our forward kinematic function, given a joint configuration of the robot.
\subsection{Self-Collision Avoidance}
For avoiding self-collisions, we formulate a distance cost that computes the largest penetration distance between spheres from all links. Since most robots allow for safe contact between consecutive links and some link pairs will never be in collision due to kinematic limits, we build a set of sphere pairs~$S$ for which self-collision needs to be checked.  We empirically found only 50\% of the sphere pairs end up in this collision pair set across many widely used manipulators. We scale the largest penetration distance by a scalar weight~$\beta_1$. Our self-collision term can be written as,
\begin{align*}
    C_{r}(K_s(\theta_t)) &= \beta_{1} \max_{i,j\in S} (\max(0,s_{i,r} + s_{j,r} - ||s_{i,x} - s_{j,x}||)) \numberthis{eq:self-collision}
\end{align*}
where~$s_{i,x}, s_{j,x} \in \mathbb{R}^3$ are the positions, $s_{i,r}, s_{j,r} \in \mathbb{R}$ are the radii of of spheres~$i$ and $j$ respectively. 

\begin{figure}
    \centering
    \begin{tabular}{c c}
    \includegraphics[height=6cm, trim={11cm 1cm 9.5cm 4cm}, clip]{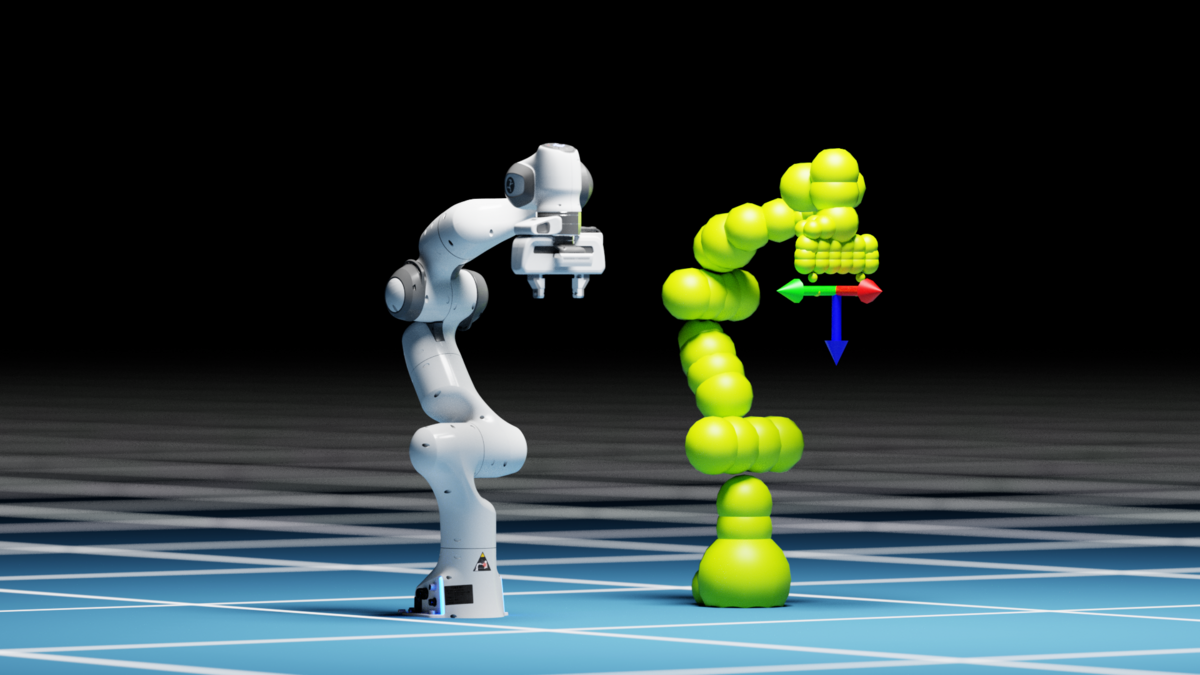}
         &  
 \includegraphics[height=6cm,  trim={0cm 0.6cm 0 0.4cm}, clip]{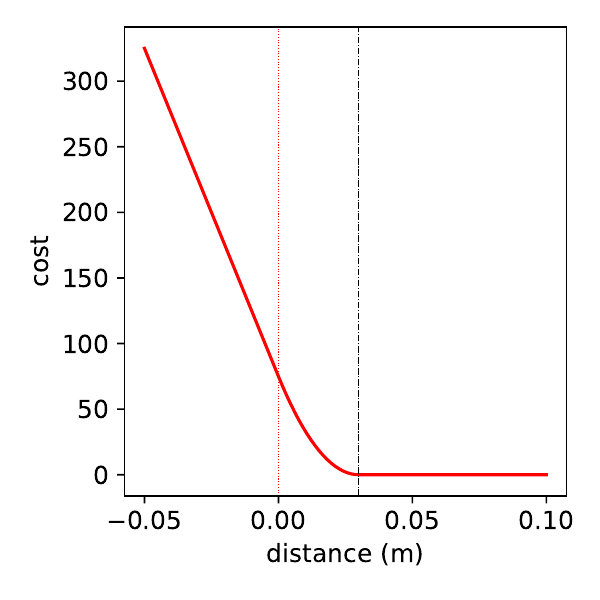}\\
    (a) Sphere representation of Franka Panda & (b) Quadratic collision cost profile\\
    \end{tabular}
    \caption{A sphere representation of the Franka Panda is shown in the left. We also visualize the end-effector frame by the axis near the gripper. cuRobo's kinematics function takes the joint configuration of a robot and outputs the location of these spheres and the pose of the end-effector. We generate bounding spheres for the robots using NVIDIA Isaac Sim's sphere generator~(\href{https://docs.omniverse.nvidia.com/prod_digital-twins/app_isaacsim/advanced_tutorials/tutorial_motion_generation_robot_description_editor.html\#adding-collision-spheres}{website}). On the right, we show the quadratic cost profile for smoothly transitioning from collision~(dotted vertical line at 0m) to non-collision space using an activation distance~$\eta$~(shown by dashed black vertical line at 0.03m).}
    \label{fig:app_robot_sphere}
\end{figure}
\subsection{World Collision Avoidance}
Generating smooth obstacle avoidance behavior has been studied extensively in the the literature~\cite{ratliff2009chomp, ratliff2015riemo, tracy2022differentiable,van2022geometric}. 
We highlight common pitfalls in collision-free motion optimization and how our approach overcomes them by leveraging existing techniques and introducing novel contributions below,
\\ \noindent \textbf{Discontinuity at surface boundary:} Discontinuity in the collision cost term near an obstacle surface leads to poor conditioning of the optimization problem, especially when collision cost term is non-convex. To mitigate this issue, we add a buffer distance~$\eta$ and change the cost to be quadratic when within~$\eta$ distance to the obstacle surface similar to~\cite{ratliff2009chomp} as shown in Fig.~\ref{fig:app_robot_sphere}-(b). This modification of the collision distance~$d_c$, given the signed distance~$d$ can be written as, 
\begin{align*}
    d_c &= \begin{cases}
        d  + 0.5\eta & \text{if } d > 0\\ 
        \frac{0.5}{\eta} (d + \eta)^2 & \text{if } -\eta < d < 0\\
        0 & \text{otherwise}
    \end{cases}\numberthis{eq:smooth-distance-cases}
\end{align*}

\noindent \textbf{Speeding through obstacles:} When a collision cost term only penalizes the position of the sphere, the optimization can attempt to move through obstacles (i.e., high penalty region) very fast to reach a lower cost region compared to being in collision for many timesteps~(see our website for a visualization of this phenomenon). To mitigate this issue, we implement a speed metric, similar to~\cite{ratliff2009chomp}, that scales the collision cost of a sphere by it's velocity~$\dot{s}$ (calculated through finite-difference). This encourages the optimization to move around the obstacle instead of speeding through an high penalty region. 

\noindent \textbf{Collision at real-robot execution:} 
Tuning a robot's control box to track a planned kinematic trajectory with millimeter accuracy at high speeds can be very time consuming. In addition, most manufacturers do not provide many parameters to tune their control box. Any Path deviation near obstacles could lead to catastrophic collisions with the world. To be robust to path deviations, we penalize the robot's velocity when within~$\eta$ distance to obstacles as robots can track with higher accuracy at slower speeds. This penalization is performed by enabling our speed metric when within~$\eta$ distance instead of only enabling at collision. This brings our collision term to~$d_s=\dot{s}d_c$.

\noindent \textbf{Collision with thin obstacles:} Motion optimization is commonly done by discretizing the trajectory by some timesteps and computing collisions at these timesteps. However, if the trajectory does not have a fine resolution of discretization, collisions with very thin obstacles could be missed. To overcome this issue, we develop a novel formulation of continuous collision checking that only requires a point query signed distance function from the world representation, enabling continuous collision checking with a variety of world representations. We discuss this formulation in the next Section~\ref{sec:continuous-coll}.

\subsection{Continuous Collision Checking}
\label{sec:continuous-coll}
Continuous collision checking is a well studied problem~\cite{redonsweptcoll, kim2003fast, redon2002fast}, with many methods building a swept volume followed by checking collision between the swept volume and the obstacles. However, collision checking using swept volumes requires the world representation to be able to compute signed distance between complex geometry~(i.e., the swept volume) and the obstacles in the world. This is only possible for worlds represented by primitive shapes or meshes. Worlds represented using neural networks~\cite{vandermerwe-icra2020-reconstruction-grasping, Ortiz-RSS-22} or voxels~\cite{oleynikova2017voxblox} will require special mechanisms to work with swept volumes. To avoid this complexity, we introduce a novel formulation of continuous collision checking that only requires a point signed distance query function from a world representation. We hope that this reduces the barrier for the perception community to deploy their world representations into cuRobo for global collision-free motion generation. Our method is related the iterative method from Bruce~\cite{bruce2006real} and loosely related to the bubbles concept from Quinlan and Khatib~\cite{quinlan1993elastic}. The difference between our approach and Bruce's approach is we not only compute the signed distance but also the gradients. We also formulate the computation to run in parallel threads for each time step in the trajectory.

\begin{figure}
    \centering
    \includegraphics[width=\textwidth, trim={0.4cm 0 0.4cm 0}, clip]{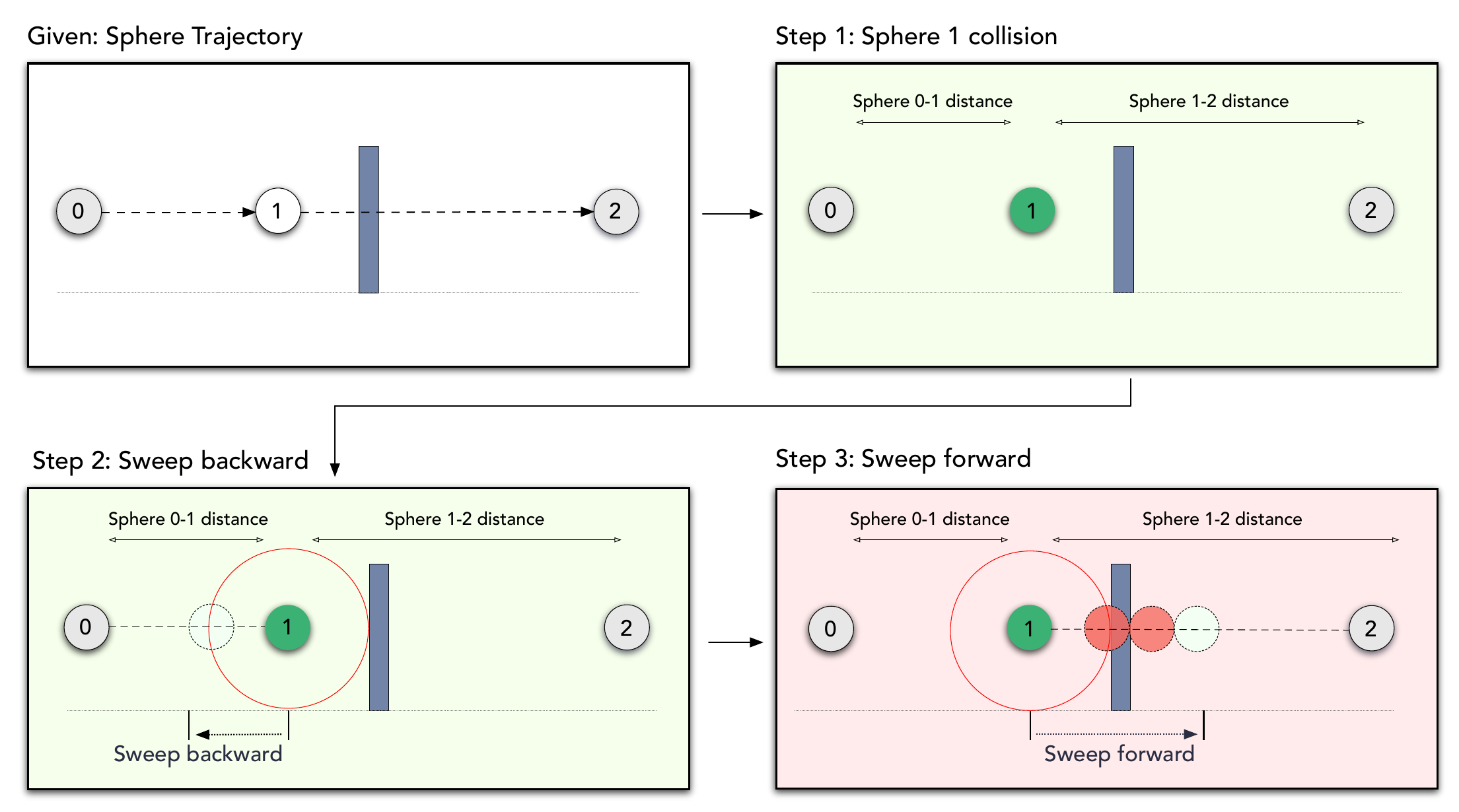}
    \caption{Our continuous collision checking algorithm is illustrated for a sphere moving from position~$0$ to~$2$ through~$1$. To compute the collision distance at a timestep~$S_1$, we first check if~$S_1$ is in collision~(step 1). If it is not in collision, we compute the distance to the closest obstacle~(shown as red circle in step 2) and~\emph{sweep backward} by checking for collision at this distance along the trajectory between~$S_0$ and~$S_1$. If no collision is found, we repeat until we reach midpoint between~$S_0$ and~$S_1$. We similarly do this iterative process between~$S_1$ and~$S_2$ in~\emph{sweep forward}~(step 3).}
    \label{fig:sweep_collision}
\end{figure}

Our continuous collision checking algorithm is illustrated in Fig.~\ref{fig:sweep_collision}. Given a trajectory of a sphere discretized by three timesteps~$S_{[0,1,2]}$, we first check if the sphere~$S_1$ is in collision. If it is in collision, we compute the collision cost and move by sphere radius. If it's not in collision, we compute the signed distance to the nearest obstacle and move this distance along the direction of motion between~$S_0$ and~$S_1$, which we term as \emph{sweep backward}. If we hit a collision, then we compute the collision cost and then continue sweeping until we reach the midpoint between~$S_0$ and~$S_1$. Similarly, we \emph{sweep forward} until midpoint between~$S_1$ and~$S_2$. For every sphere location in the trajectory, we \emph{sweep forward} and \emph{sweep backward} upto the mid distance as this enables our gradient computations to be parallelizable (i.e., gradient for a sphere location does not depend on the collisions at other sphere sweeps).

Our implementation of the algorithm assumes that the sphere is moving linearly between waypoints which is not true for links attached through a revolute joint. Empirically, we found that even with this assumption, the optimization was able to find paths in tight spaces. We leave incorporating exact movement path between sphere waypoints for future work and discuss the implementation of this algorithm in Appendix~\ref{app:signed_distance_cuda}.

\subsection{World Representation}
To compute the world collision cost term, we only require the ability to compute the closest point~$c_r\in \mathbb{R}^3$ to a query point and also if the query point is inside or outside an obstacle~$s_r \in [-1,1]$. These quantities can be obtained from a differentiable point query signed distance function or through geometric processing. We implement three different world representations that output these quantities through geometric processing,
\begin{description}
    \item[Oriented Bounding Box] We implement an efficient cuboid query function as we found cuboids to be a common representation for collision avoidance in many real-world deployments. For this world representation, we assume the world is made of only oriented bounding boxes~(i.e., cuboids with a~$\mathrm{SE}(3)$ pose).
    \item[Mesh] Leveraging NVIDIA warp's bounding volume hierarchy~(BVH), which stores mesh's faces and vertices in an accelerated framework for fast closest point and inside/outside queries. This representation assumes the world is represented by watertight meshes.
    \item[Depth Camera] Third, we write a wrapper to NVIDIA nvblox~\cite{millane2023nvblox} which integrates Euclidean Signed Distance (ESDF) from truncated Signed Distance Fields (TSDF) streaming from a depth camera.  This representation enables us to build a ESDF voxel representation of the world using a depth camera and use this for computing collision distance. This representation assumes that we have access to a depth camera and accurate pose of the camera with respect to the robot's base at each frame for integrating into nvblox. 
\end{description}
Our implementation of the collision cost also allows for using a combination of the three world representations as we sum over all collisions in the world. In addition to these representations, our robot sphere representation allows for interfacing with other methods that can output a differentiable signed distance~\cite{park2019deepsdf, vandermerwe-icra2020-reconstruction-grasping, Ortiz-RSS-22}. For example, Tang~\etal~\cite{tang2022rgb} integrated a learned SDF representation of the world~\cite{Ortiz-RSS-22} for collision avoidance in cuRobo. 

The overall world collision term can be written as,
\begin{align*}
    C_w(S_{t-1,t,t+1}) &=  \beta_2 \text{speed}(S_{t-1,t,t+1})  \text{smooth}(\text{sweep}(S_{t-1,t,t+1}))\numberthis{eq:world-collision-cost}
\end{align*}
where~$\beta_2$ scales the cost by a large penalty to act as a soft constraint, $\text{sweep}(\cdot)$ computes the collision distance using our continuous collision checking algorithm from Sec.~\ref{sec:continuous-coll}. The function~$\text{smooth}(\cdot)$ adds the quadratic smoothing over the collision distance using Eq.~\ref{eq:smooth-distance-cases}, which is then scaled by the velocity of the sphere~$\text{speed}(\cdot)$. We sum this term across all spheres that represent the robot.
\section{Parallel Optimization Solver}
\label{sec:opt_solver}
There are several techniques to solve the optimization problem defined in Section~\ref{sec:mopt}, from particle-based optimization~\citep{kalakrishnan2011stomp, Lambert2021EntropyRM} to gradient-based optimization~\citep{schulman2014motion,mukadam2018continuous,ratliff2009chomp, zucker2013chomp} methods. In particular many trajectory optimization methods~\citep{mukadam2018continuous,ratliff2009chomp, zucker2013chomp} have approximated hard constraints as soft constraints by treating them as cost terms with large weights to transform the optimization problem from one with nonconvex constraints to a box-constrained nonconvex optimization problem. Motivated by these successes, we also approximate our constraints as cost terms and implement a quasi-newton solver to solve this nonconvex optimization problem.

L-BFGS, a quasi-newton optimization method that can solve very large optimization problems, is a common method shown to achieve superlinear convergence by estimating the Hessian using evaluated gradients. Our optimizers are built around L-BFGS because of its combined performance and relative simplicity that aids parallelization.
Gauss-Newton solvers are also ubiquitous and important in robotics \cite{toussaint2014komo,schmidt2015dart,dellaert2021factorGraphs}, but after an initial exploration we decided to focus our experiments on L-BFGS. Many formulations of Gauss-Newton restrict their presentation to the nonlinear least-squares problem \cite{nocedal1999numerical,toussaint2014komo} where performance is best understood, but that's unduly restrictive in our setting. These methods can be generalized as a form of natural gradient descent \cite{ratliff2015riemo} and related formulations of iLQG \cite{todorov2005iLQG} demonstrate their empirical utility on more general costs using quadratic approximations. However, appropriately leveraging the problem structure within a GPU is not straightforward and the band-diagonal solve commonly used is inherently sequential leaving a number of open questions we would need to resolve. Our benchmarks indicate that even the simpler L-BFGS shows significant improvement over the state-of-the-art when GPU compute is properly leveraged; we leave a full exploration of Gauss-Newton to future work.
\subsection{Parallel L-BFGS Optimization}
\label{sec:gradient-solver}
Our L-BFGS optimizer has two steps, we first compute the step direction given the current optimization variables~$\Theta = \theta_{t\in[0,T]}$ and the gradient~$\Delta \Theta$ with respect to the sum of the cost terms using the standard L-BFGS steps as described in Nocedal and Wright\cite{nocedal1999numerical}. Given this step direction~$\Delta \Theta$, we perform line search by scaling the step direction with a discrete set of magnitudes~$\alpha\in \mathbf{R}^n$ and computing the best magnitude from this set using Armijo and Wolfe conditions as shown in Alg.~\ref{alg:linesearch}. Extensive details on our solver is available in Appendix~\ref{sec:solver-details}.

Our approach of trying a predefined discrete set of magnitudes instead of iteratively searching for the largest magnitude that satisfies the condition enables us to more effectively use parallel compute as the cost and gradient for the discrete set can be computed in parallel. For the case where none of the values in our discrete set satisfies the line search conditions, we use a very small magnitude~(0.01) which acts as a noisy step update. This noisy step update also prevents NaN values in the the step direction computation as there is always a perturbation in the optimization variables between iterations. After every optimization iteration, we update our best estimate as the optimization could diverge due to noisy perturbation in line search. Empirically, we found that the use of a noisy perturbation instead of stopping the optimization when line search fails to find a magnitude that satisfied the conditions greatly increased the convergence rate on trajectory optimization problems as shown in Section~\ref{sec:optimization-analysis}. 
 \begin{algorithm}
        \DontPrintSemicolon
        \algokw
        \caption{Parallel Noisy Line Search}
        \label{alg:linesearch}
        \KwIn{$\Theta, \Delta \Theta, \alpha=[0.01,...]$}
        $\Theta_l \gets \text{clip}(\Theta + \alpha \Delta \Theta)$ \tcp*{get bounded variables}
        $c_0, c_l\gets c(\Theta_{0}),  c(\Theta_l)$ \tcp*{compute cost for magnitudes}
        $\delta \Theta_l \gets \delta c_l$\tcp*{compute batched gradients}
        $a \gets c_l \leq c_0 + \eta_1  \alpha \Delta \Theta$ \tcp*{Armijo Condition}
        \If{Wolfe}
        {
        \leIf{Strong}{
            $a_2 \gets \text{abs}(\delta \Theta_l \alpha) \leq \eta_2 \Delta \Theta$\;
        }
        {
            $a_2 \gets \delta \Theta_l \alpha \geq \eta_2 \Delta \Theta$
        }
        $a \gets a$ $\&\&$ $a_2$\;
        }
       
        $i \gets \text{largest\_true}(a)$ \tcp*{returns 0 when none is true}
        $\hat{c}, \hat{\Theta}, \delta \hat{\Theta}  \gets c_l[i], \Theta_l[i],  \delta\Theta_l[i]$\tcp*{return best}
    \end{algorithm}

\subsection{Particle-Based Optimization}
\label{sec:particle-solver}
To encourage L-BFGS to reach a local optima, we devise a strategy combining particle and gradient-based optimization. This is inspired by strong theoretical results in stochastic gradient Markov Chain Monte Carlo~\cite{Welling2011Lan, Ma2015MCMC}, and sampling-based MPC controllers such as MPPI~\cite{wagener2019online}. In our method, we first run a few iterations of particle-based optimization over the initialization before sending to L-BFGS. Given an initial mean trajectory of joint configurations~$\Theta_\mu=\theta_{[1,T]}$ and a covariance~$\Theta_{\sigma}$, we sample~$n$ particles~$\theta_{n,[1,T]}$ from a zero mean Gaussian and then update~$\theta_{n, [1,T]} = \Theta_\mu + \sqrt{\Theta_{\sigma}} * \theta_s$. We compute the cost for these particles~$C(\theta_{n,[1,T]})\in \mathbb{R}^n$ and calculate the exponential utility~$w=\frac{e^{c_i}}{\sum_{i=0}^n e^{c_i}}$, where $c= \frac{-1.0}{\beta}C(\theta_{n,[1,T]})$. We then update the mean and covariance as,
\begin{align*}
    \Theta_\mu &= (1-k_\mu)\Theta_{\mu-1} + (k_\mu) w * \theta_{n,[1,T]}\numberthis{eq:particle_1},\\
    \Theta_{\sigma} &= (1-k_\sigma) \Theta_{\sigma -1} +  w * (\theta_{n,[1,T]}^2 - \Theta_{\mu-1}).\numberthis{eq:particle_2}
\end{align*}
We found that the use of particle-based optimization to initialize L-BFGS led to better convergence as empirically validated in Sec.~\ref{sec:optimization-analysis}. To tackle very hard problems and further reduce the number of seeds required to converge, we develop a parallelized geometric planner that generates collision-free geometric paths between start and goal in the next section.

\section{Parallel Geometric Planner}
\label{sec:graph}
We develop a geometric planner to generate a collision-free path from the start configuration~$\theta_0$ to the goal configuration~$\theta_{T}$. This generated path is specified by a list of $w$~waypoints~$\theta_{[0,w]}$ through which the robot passes in a linear fashion. By studying common geometric planning methods~\cite{lavalle2006PlanningAlgorithms}, we found three main components in graph building that can benefit from parallel compute. Specifically, sampling collision-free nodes, finding k nearest nodes in graph, and steering from each sampled node to k nearest nodes. We implement algorithms to perform these tasks in parallel on the GPU in our geometric planner.

\begin{algorithm}
    \DontPrintSemicolon
    \algokw
    \caption{Parallel Geometric Planner}\label{alg:graphsearch}
    \KwProb{$\theta_{b,0}$, $\theta_{b,g}$}
    \KwIn{$g_{max}, g_{refine}$, $c_{max}$,$c_{default}$, $k_{refine}$, $k_{explore}$,  $p_{init}$, $p_{refine}$, $p_{explore}$}
    \KwResult{path\_found, path($\Theta_{b,[0,w]}$)}
    \KwInit{$k_n\gets k_{explore}$, $c_{max} \gets c_{default}$, $p_n\gets p_{explore}$, $i \gets 0$}
    e = [[$\theta_{b,0}$, $\theta_{b,g}$],
         [$\theta_{b,g}$, $\theta_{b,0}$],
         [$\theta_{b,0}$, $\theta_r$],
         [$\theta_{r}$, $\theta_{b,0}$],
         [$\theta_{b,g}$, $\theta_{r}$],
         [$\theta_{r}$, $\theta_{b,g}$]] \;
    steer\_connect(e)         \tcp*{Connect start, goal, and retract}
    path\_found, path, min\_len $\gets$ shortest\_path($\theta_{b,0} $, $\theta_{b,g}$)\;
    \If{path\_found}
    {
        path, min\_len, $ c_{max} \gets$ shortcut\_path($\theta_{b,0}$, $\theta_{b,g}$)\; %
        \lIf{min\_len $== 2$}{
            return path
        }
    }
    $c_{min} \gets dist(\theta_{b,0}, \theta_{b,g})$\;

\While{not path\_found or $i  < g_{refine}$}{
        $id \gets $random(!path\_found) \tcp*{Pick an index from the set of queries that do not have a path yet.}
        $\theta_{s,k} \gets$ sample\_nodes($\theta_{0},\theta_g, c_{max},p_n$)\tcp*{sample nodes within ellipse}
        $e \gets near(k_n, \theta_{s,k})$ \tcp*{Find~$k_n$ nearest samples~$\theta_{s,k}$ to existing nodes in graph}
        steer\_connect(e) \tcp*{Steer and connect to graph}    
        path\_found,  path, min\_len $\gets$ shortest\_path($\theta_{b,0}$, $\theta_{b,g}$)\;
        $i += 1$\;   
        \eIf{path\_found \&\& min\_len $>$ 3}{
            path, min\_len, $ c_{max} \gets$ shortcut\_path(path)\; 
        }
        {
                $c_{max}[id] \gets c_{max}[id] +  c_{min}[id] * \eta_{explore}$\;
                $p_n += \eta_{explore} * p_n$ \;
                $k_n += \eta_{explore} * k_n$ \;   
        }
    }
    return path\_found, path\;
 \end{algorithm}

Our geometric planner as shown in Alg.~\ref{alg:graphsearch}, first performs heuristic planning by checking if we can steer from start to goal configuration directly~\cite{srinivasa2016system} or through a predefined retract configuration~$\theta_r$ (lines 1-7). If this heuristic fails, we sample collision-free configurations~$v_{new}$ from an informed search region that samples within~$c_{max}$ of the straight line distance between start and goal similar to BIT$^*$~\cite{gammell2015batch} (line 11). We then find the~$k_n$ nearest neighbours from the existing graph and try to steer from the graph nodes to the new vertices~(lines 12-13). We repeat these steps until we find a path with only one waypoint~(line 16). Between re-attempts we grow the number of sampled nodes~$p_n$, the number of nearest neighbours~$k_n$, and the search region~$c_{max}$ to grow the exploration space of the geometric planner (lines 19-21). To efficiently leverage parallel compute in geometric planning, we develop an algorithm to steer from~$s$ vertices~$\theta_{s,0}$ in a graph with~$s$ sampled new configurations~$\theta_{s,k}$ in parallel as described in Alg.~\ref{alg:steer}. 

\begin{algorithm}
    \DontPrintSemicolon
    \caption{Parallel Steering}\label{alg:steer}
    \SetKwInOut{KwParam}{Parameters}
    \KwIn{$e=[\theta_{s,0}, \theta_{s,k}]$}
    \KwParam{r, $d_w$}
    $\vec{g} \gets distvec(\theta_{s,0}, \theta_{s,k})$ \tcp*{distance between nodes}
    $n \gets max(|\vec{g}|/r)+1$\tcp*{find largest distance}
    $\vec{d} \gets d[:n+1]/n$\tcp*{discretize based on largest distance}
    $\vec{l} \gets \theta_{b,0} + \vec{d} * \vec{g}/d_w$ \tcp*{get disretized edges}
    $mask \gets \text{mask\_samples}(\vec{l})$ \tcp*{check for validity}
    $h \gets \text{first\_false}(mask) - 1$ \tcp*{first collision index/edge}
    $h[h==-1] \gets n$\;
    $v_{new} \gets l[h]$ \tcp*{store last valid point/edge}
    $d \gets \text{dist}(\theta_{s,0}, v_{new})$ \tcp*{store distance value in edge}
    graph\_add($\theta_{b,0}, v_{new}, d$)\;
\end{algorithm}

We also implement a \texttt{shortcut\_path()} function that tries to connect each waypoint with every other waypoint in the path to try to find a shorter path. We leverage this geometric planner to also find paths between a batch of start and goal configurations or from a single start to a goal set. We achieve this by randomly choosing a query index (for which a path does not exist yet) and sampling in this query region to expand the graph (lines 10,11). 
\section{Results}
\label{sec:benchmark}

\begin{figure}
    \centering
    \includegraphics[width=0.99\textwidth, trim= 0 0 0 4.5cm, clip]{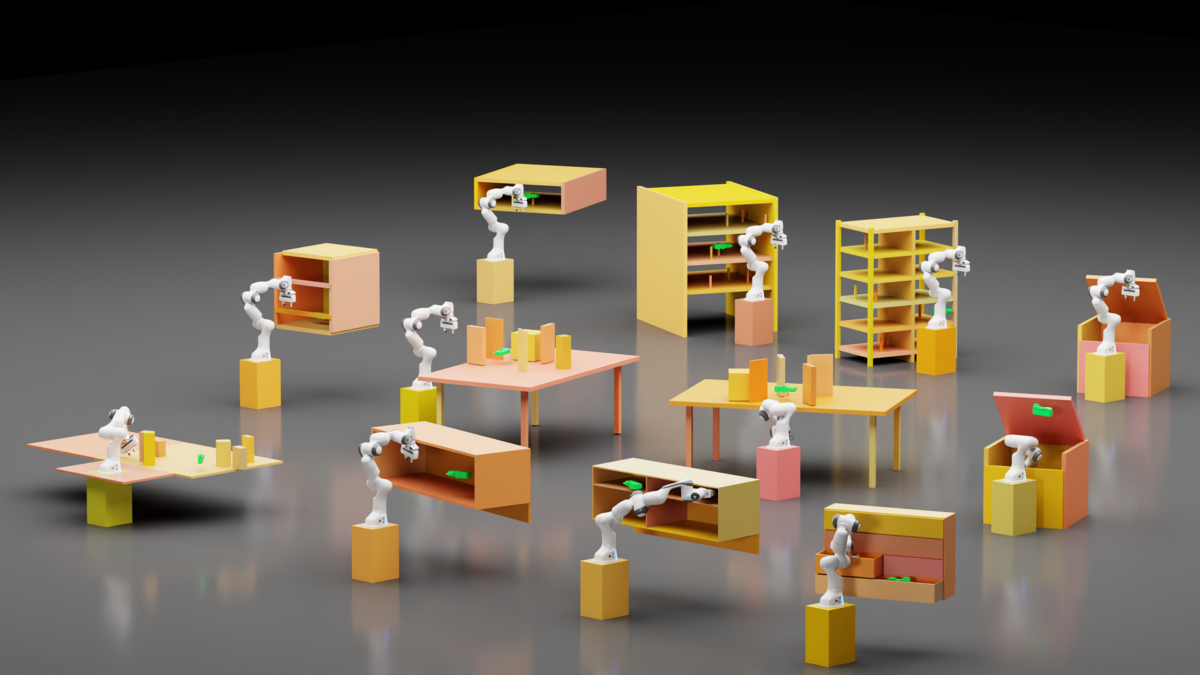}\vspace{0.1cm}
    \includegraphics[width=0.99\textwidth, trim= 0 0 0 4.5cm, clip]{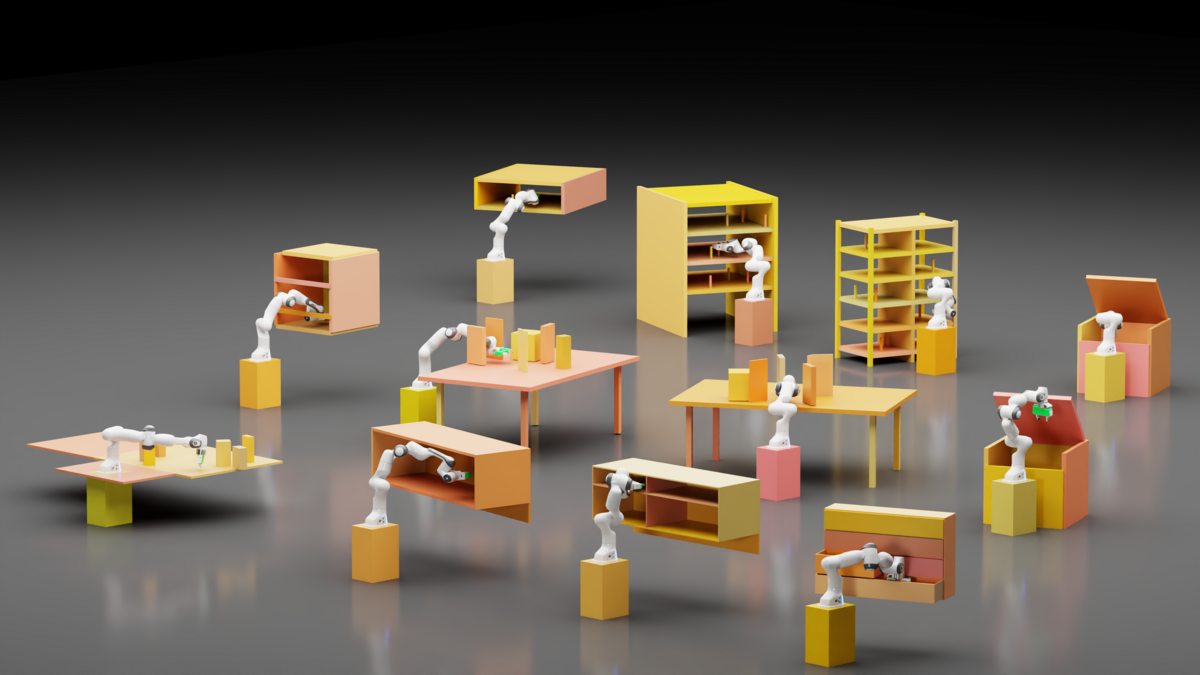}
   \caption{Motion planning problems across the 12 different scenes are visualized here. The top image shows the robot in the start configuration and the bottom image shows the robot at a final configuration after running cuRobo's motion generation. The top two rows show scenes from the motion benchmaker dataset~\cite{chamzas2021motionbenchmaker} and the bottom row shows the scenes from motion policy networks~\cite{fishman2022mpinets}}
    \label{fig:evaluation_dataset}
\end{figure}

We validate and compare our approach to existing methods on two motion planning datasets, the motionbenchmaker dataset~\cite{chamzas2021motionbenchmaker} containing 800 problems and the mpinets dataset~\cite{fishman2022mpinets} containing 1800 problems. Both datasets contain motion planning problems for the Franka Emika Panda robot, which has 7 actuated joints. The datasets span 12 unique scene types, with each problem starting the robot at a collision-free joint configuration and defining the goal as a desired end-effector pose. A few instances of the motion planning problems from this set is shown in Fig.~\ref{fig:evaluation_dataset}. We provide the planning problems along with code to compute different metrics at \href{https://github.com/fishbotics/robometrics}{github.com/fishbotics/robometrics}. 

We first analyze the quality of solutions in Section~\ref{sec:motion-gen-benchmark}, then compare compute times in Section~\ref{sec:mb-time}. We also analyze our collision-free inverse kinematics solver in Section~\ref{sec:inv-kin-benchmark}, followed by an analysis of our 
kinematics and distance query modules in Section~\ref{sec:kin-benchmark}, as they can also be used independently in other manipulation tasks.

\subsection{Motion Generation Quality}
\label{sec:motion-gen-benchmark}
We focus analysis of motion generation quality to metrics that affect the success rate, and execution time of the computed motions. We compute six different metrics that capture geometric and temporal qualities of the generated trajectories. First, we compute the standard metrics from geometric planning methods, specifically \emph{Success} on a dataset within a given time and~\emph{C-Space Path Length} which is the distance traveled by the robot's joints to reach the target pose. In addition, we introduce four metrics that evaluates time parameterization of the generated motions. The \emph{Motion Time} metric compares the trajectory times given the number of trajectory points~$n$ and the time~$dt$ between waypoints~(i.e., $(n-1)*dt$). If a robot perfectly tracks the planned trajectory, then this~\emph{Motion Time} would be the execution time. When moving manipulators at high-speed, they become very sensitive to jerk profiles, especially when starting from or ending at zero velocity~(i.e., idle). This has prevented motion generation approaches from executing at the full rated speed of a robot. We hence introduce a metric that measures the maximum jerk across the trajectory, which we call~\emph{Maximum Jerk} metric. We also measure the \emph{Maximum Acceleration} and \emph{Mean Velocity} across the trajectory to draw further comparisons between methods.

We compare our method to Tesseract~\cite{tesseract} which uses Bullet's continuous collision detector~\cite{coumans2021}, Open Motion Planning Library (OMPL)~\cite{sucan2012open} for geometric planning, and TrajOpt~\cite{schulman2014motion} for trajectory optimization. We use two motion planning methods from Tesseract, one that only performs geometric planning with RRTConnect~\cite{kuffner2000rrt} which we call~\emph{Tesseract-GP} and one that uses the geometric plan as a seed to Trajopt for trajectory optimization which we call~\emph{Tesseract}.\footnote{We also tried Tesseract's TrajOpt integration with a linear seed but it failed to find solutions on most problems.} We compare these baselines to two versions of motion generation from cuRobo, one that only does geometric planning~\emph{cuRobo-GP} using our algorithm from Section~\ref{sec:graph} and the other that does Trajectory Optimization~\emph{cuRobo}. For all methods, we timeout at 60 seconds and allow random restarts until this timeout is reached. More details on the baselines and the evaluation methods are available in Appendix~\ref{app:baseline}. 

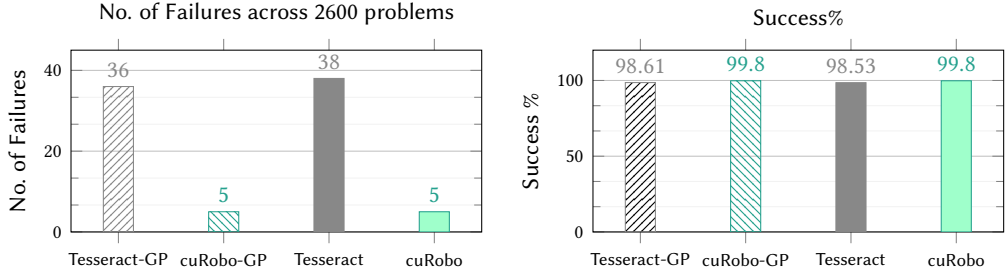
\begin{figure}
    \centering
    \begin{tabular}{c c}

    \begin{tikzpicture}  
     \tikzstyle{every node}=[font=\footnotesize]
        \begin{axis}  
        [
            ybar,  
            grid style={line width=.1pt, draw=gray!10},
            major grid style={line width=.2pt,draw=gray!50},
            ymajorgrids,
            yminorgrids,
            minor y tick num=2,
            bar width=0.4cm,
            x=1.4cm,
            title={No. of Failures across 2600 problems},
            enlarge x limits=0.15,
            every node near coord/.append style={/pgf/number format/fixed},
            ylabel={No. of Failures},
            symbolic x coords={Tesseract-GP, cuRobo-GP, Tesseract, cuRobo}, 
            xtick={Tesseract-GP, cuRobo-GP, Tesseract, cuRobo},  
            nodes near coords,   
            nodes near coords align={vertical},  
            ymin=0,
            ymax=45,
            height=4cm,
            width=0.3\textwidth,
            xticklabel style={align=center},
            bar shift = 0pt,
            ]  
        \addplot[barorangeline, fill=barorange,  pattern color=barorange, pattern = north east lines] coordinates{(Tesseract-GP, 36)};
        \addplot[bargreenline, fill=bargreen,
         pattern color=bargreenline, pattern = north west lines] coordinates{(cuRobo-GP, 5)};
        \addplot[barorangeline, fill=barorange] coordinates{(Tesseract, 38)};
        \addplot[bargreenline, fill=bargreen] coordinates{(cuRobo, 5)};
         \end{axis}  
    \end{tikzpicture}
    &
    \begin{tikzpicture}  
     \tikzstyle{every node}=[font=\footnotesize]
        \begin{axis}  
        [
            ybar,  
            grid style={line width=.1pt, draw=gray!10},
            major grid style={line width=.2pt,draw=gray!50},
            ymajorgrids,
            yminorgrids,
            minor y tick num=2,
            bar width=0.4cm,
            x=1.4cm,
            enlarge x limits=0.15,
            title={Success\%},
            every node near coord/.append style={/pgf/number format/fixed},
            ylabel={Success \%},
            symbolic x coords={Tesseract-GP, cuRobo-GP, Tesseract, cuRobo}, 
            xtick={Tesseract-GP, cuRobo-GP, Tesseract, cuRobo},  
            nodes near coords,   
            nodes near coords align={vertical},  
            ymin=0,
            ymax=120,
            height=4cm,
            width=0.3\textwidth,
            xticklabel style={align=center},
            bar shift = 0pt,
            ]  
        \addplot[barorangeline, fill=barorange, pattern = north east lines] coordinates{(Tesseract-GP, 98.61)};
        \addplot[bargreenline, fill=bargreen, pattern color=bargreenline, pattern = north west lines] coordinates{(cuRobo-GP, 99.8)};
        \addplot[barorangeline, fill=barorange] coordinates{(Tesseract, 98.53)};
        \addplot[bargreenline, fill=bargreen] coordinates{(cuRobo, 99.8)};
         \end{axis}  
    \end{tikzpicture}
    \\
    \end{tabular}
    \caption{We plot the success of different methods on the motionbenchmaker and mpinets dataset in these plots. On the left we plot the number of failures on our evaluation dataset which contains 2600 problems.~\emph{cuRobo} only fails on 5 problems which are all invalid problems when evaluated with~\emph{cuRobo}'s collision representation. On the right plot, we see that \emph{cuRobo} has a higher success~\% than~\emph{Tesseract}.}
    \label{fig:mg-success}
\end{figure}

We report the success across the 2600 motion planning problems in Figure~\ref{fig:mg-success}. We count a trajectory as \emph{Success} when it is collision-free, it doesn't violate any joint limits, and is within 5mm and 5\% of desired position and orientation respectively. We found our geometric planner~\emph{cuRobo-GP} to find a path on 99.8\% of the dataset compared to \emph{Tesseract-GP}'s 98.6\%. Our geometric planner only failed on 5 problems compared to \emph{Tesseract-GP} failing on 36 problems. When we compare trajectory optimization methods,~\emph{cuRobo} only failed on 5 problems while~\emph{Tesseract} failed on 38 problems, giving a success rate of~99.8\% and 98.53\% respectively. 
\emph{cuRobo} and \emph{cuRobo-GP} failed on the same set of five problems. 
When we look at these problems in Fig.~\ref{fig:failure}, we see that 3 problems have their goal pose in collision with the world, 1 problem as the start configuration in collision, and one problem does not have a collision-free IK solution to the goal pose. The reason for these problems to be invalid for \emph{cuRobo} is because we evaluate~\emph{cuRobo} with our cuboid collision checker which approximates the obstacles represented by cylinders as cuboids. While~\emph{cuRobo} also has mesh-based collision checking and depth camera based collision checking, we leave evaluating these collision checkers for a future work. 

\begin{figure}
    \centering
    \begin{tabular}{c@{\hspace{0.1cm}}c@{\hspace{0.1cm}}c}
    \multicolumn{3}{c}{\sffamily {Goal Pose in Collision}}\\    
    \includegraphics[width=0.32\textwidth]{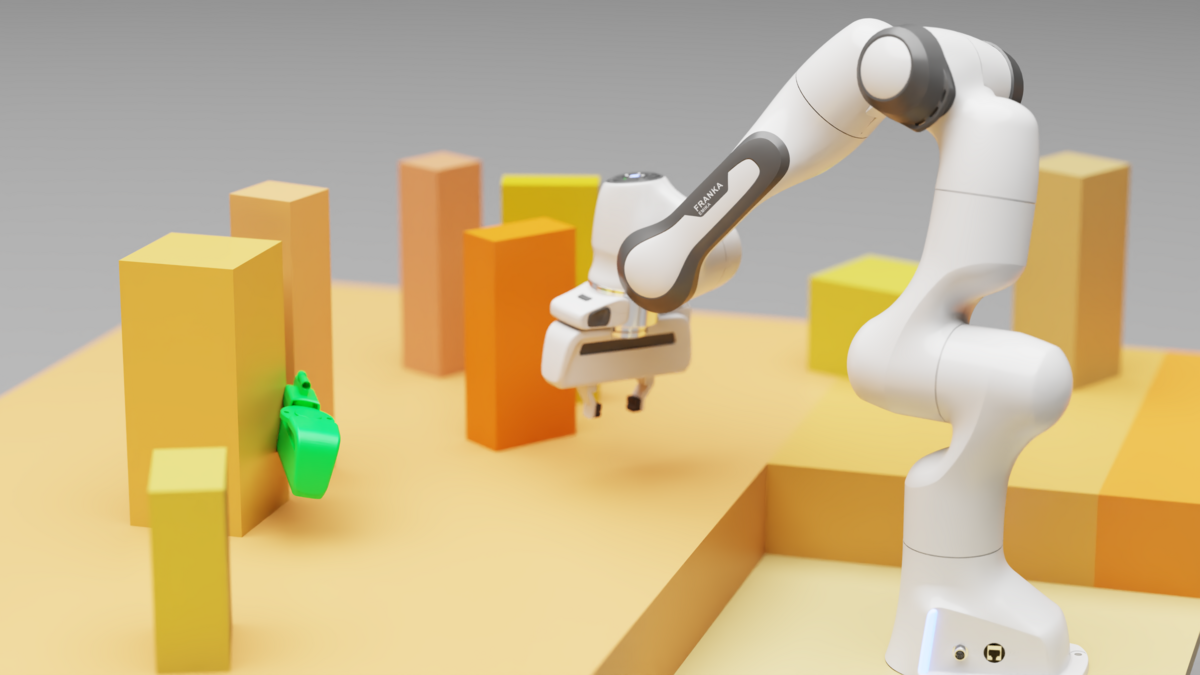} & 
    \includegraphics[width=0.32\textwidth]{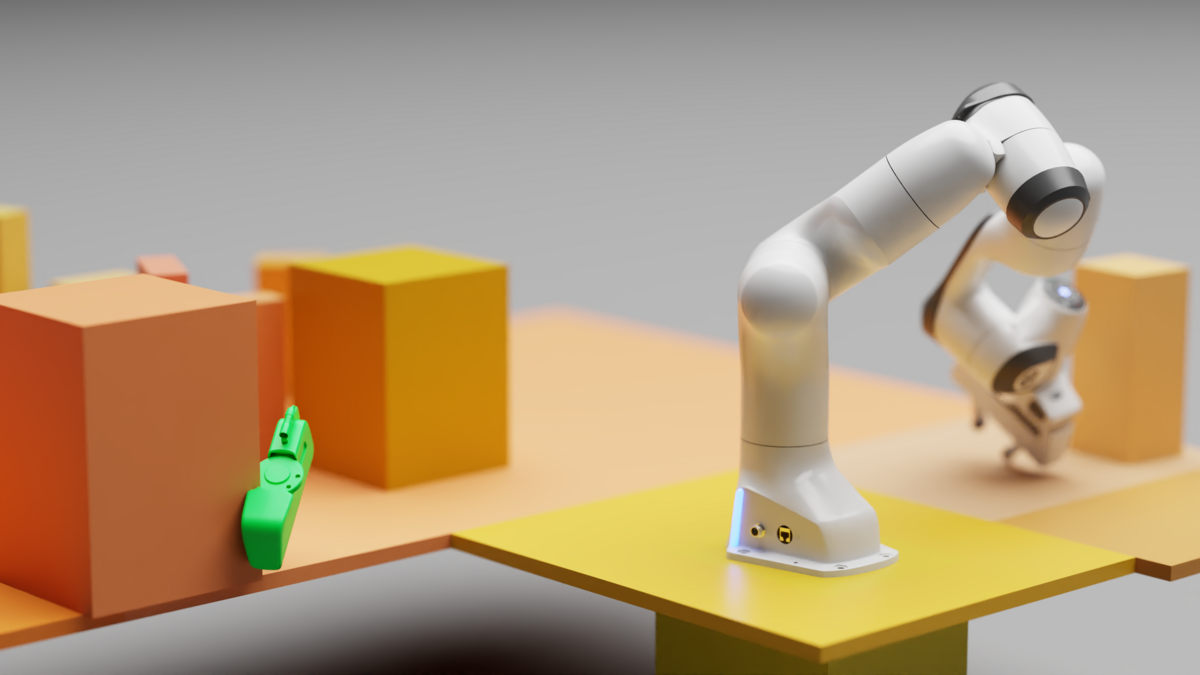} &  
    \includegraphics[width=0.32\textwidth]{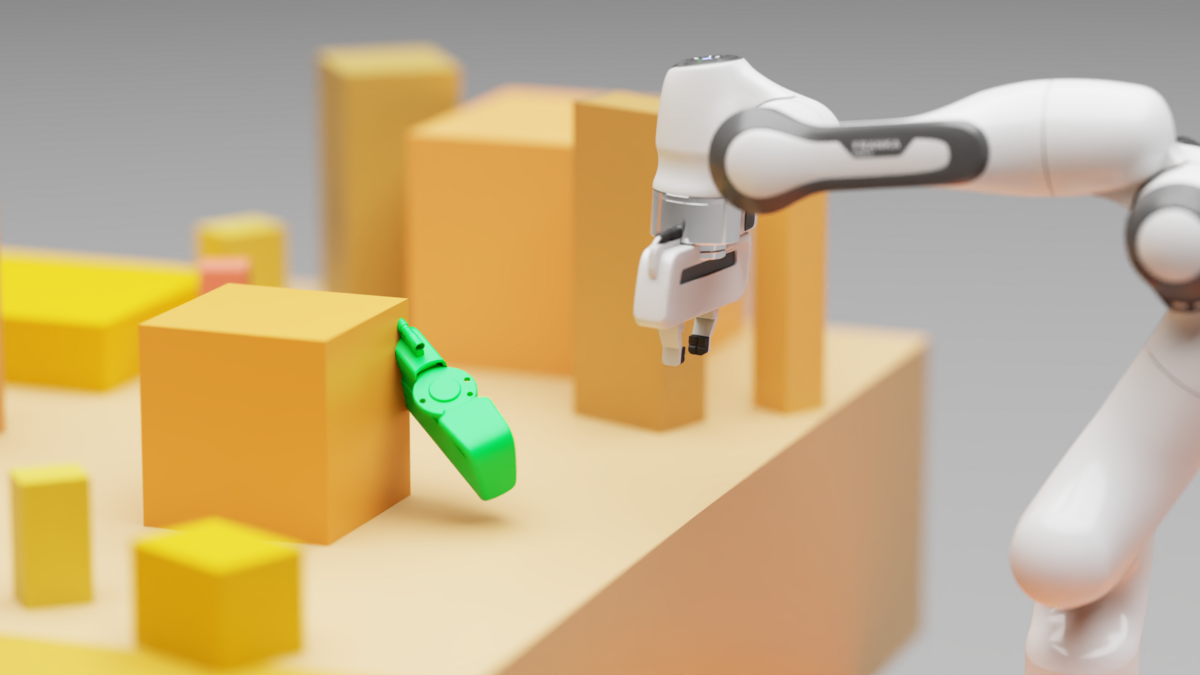}\\ 
    \end{tabular}
    \begin{tabular}{c@{\hspace{0.1cm}}c}
    \midrule
    {\sffamily {No Collision-free IK for Goal Pose}} & {\sffamily {Start Config in Collision}}\\
    \includegraphics[width=0.48\textwidth]{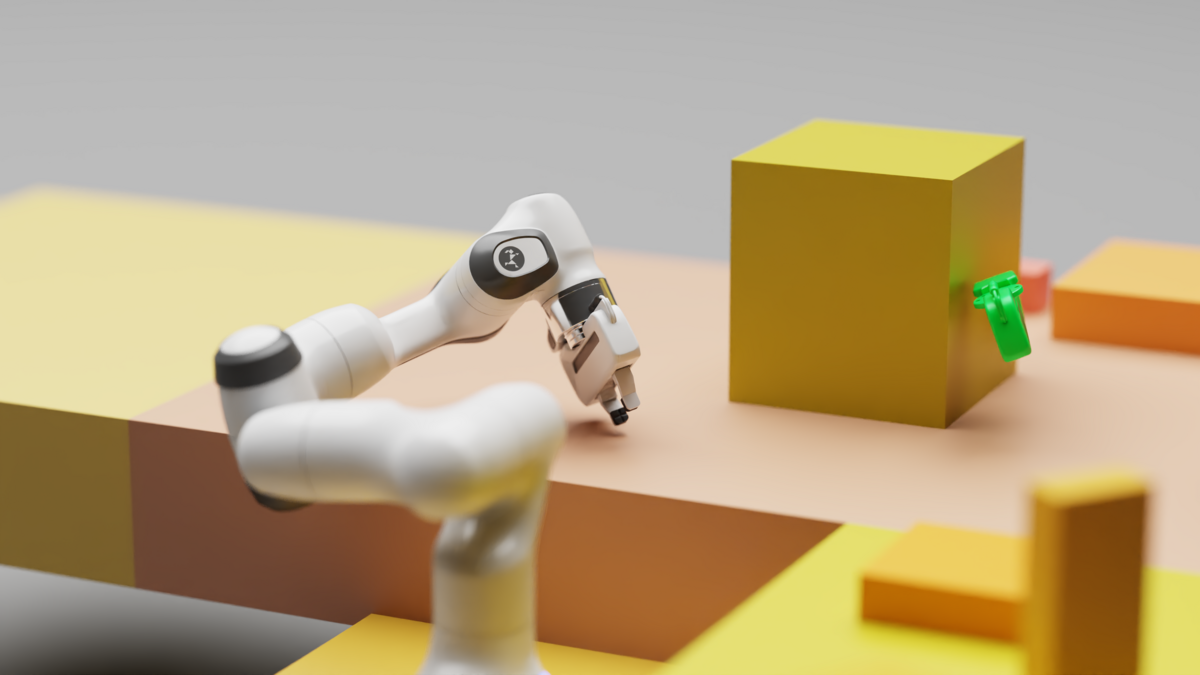} &  
    \includegraphics[width=0.48\textwidth]{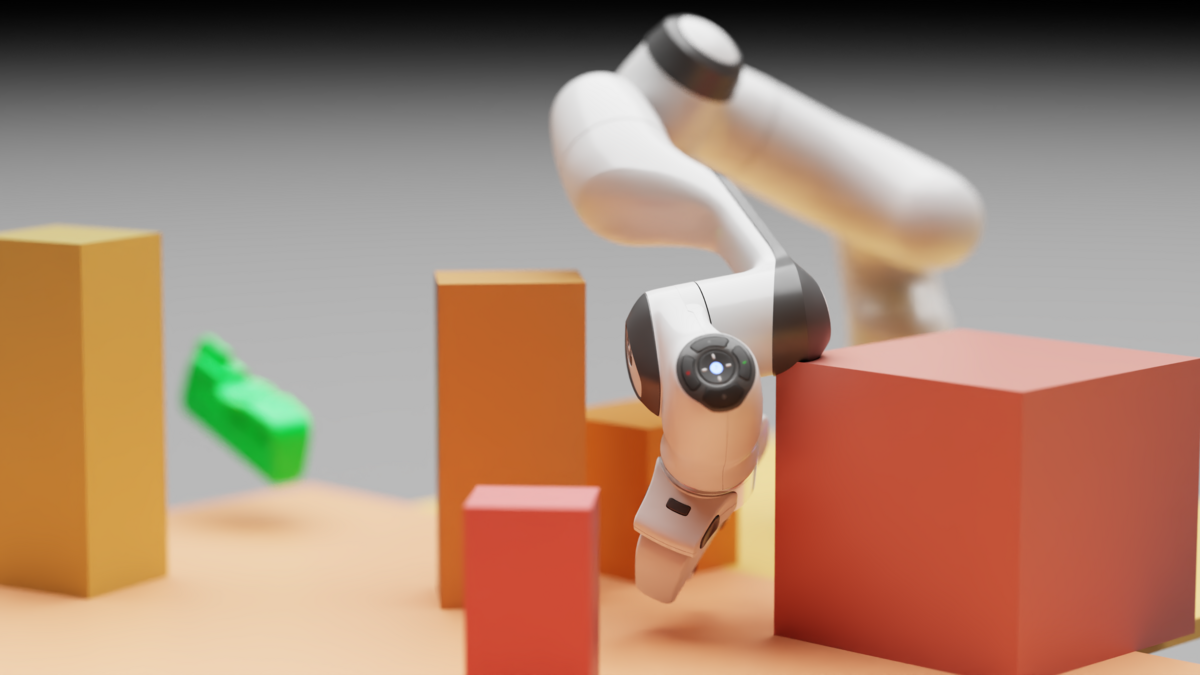}\\
    \end{tabular}
    \caption{\emph{cuRobo} failures in the dataset are visualized here. We found 3 instances the goal pose be in collision as shown in the top row. We found one instance where the all IK solutions to the goal pose was in collision as shown in the bottom left image and one instance where the start joint configuration was in collision as shown in the bottom right image.}
    \label{fig:failure}
\end{figure}
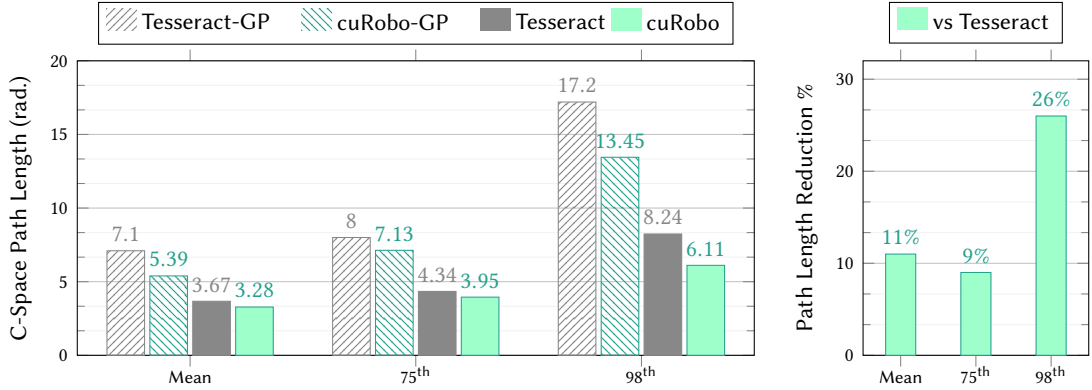
\begin{figure}
    \centering
    \begin{tabular}{c c}
\begin{tikzpicture}  
     \tikzstyle{every node}=[font=\footnotesize]
        \begin{axis}  
        [
            ybar,  
            grid style={line width=.1pt, draw=gray!10},major grid style={line width=.2pt,draw=gray!50},
            ymajorgrids,
            yminorgrids,
            minor y tick num=2,
            bar width=0.5cm,
            x=3cm,
            enlarge x limits=0.25,
            every node near coord/.append style={/pgf/number format/fixed},
            ylabel={C-Space Path Length (rad.)},
            symbolic x coords={Mean, 75$^{\text{th}}$, 98$^{\text{th}}$}, 
            xtick=data,  
             nodes near coords,   
            nodes near coords align={vertical},  
            ymin=0,
            ymax=20,
            height=5.5cm,
            width=0.9\textwidth,
            xticklabel style={align=center, text height=0.1cm},
            legend columns=4,
            legend style={at={(0.5,1.2)},anchor=north,
            /tikz/column 2/.style={
                column sep=10pt,
             },
             /tikz/column 4/.style={
                column sep=10pt,
            },
            /tikz/column 4/.style={
                column sep=10pt,
            },
            },
            legend image code/.code={%
                    \draw[#1, draw=none] (0cm,-0.1cm) rectangle (0.4cm,0.2cm);
            }, 
            legend cell align={left},
            ]  
        \addplot[barorangeline, fill=barorange,
         pattern color=barorange, pattern = north east lines] coordinates {(Mean, 7.1) (75$^{\text{th}}$, 8) (98$^{\text{th}}$, 17.2)}; 
        \addplot[bargreenline, fill=bargreenline,
         pattern color=bargreenline, pattern = north west lines] coordinates {(Mean, 5.39) (75$^{\text{th}}$, 7.13) (98$^{\text{th}}$, 13.45)};       
        \addplot[barorangeline, fill=barorange] coordinates {(Mean, 3.67)(75$^{\text{th}}$, 4.34) (98$^{\text{th}}$, 8.24)};       
        \addplot[bargreenline, fill=bargreen] coordinates {(Mean, 3.28)(75$^{\text{th}}$, 3.95) (98$^{\text{th}}$, 6.11)};       
        \legend{Tesseract-GP, cuRobo-GP, Tesseract, cuRobo}
        \end{axis}  
    \end{tikzpicture}
    &
    \begin{tikzpicture}  
     \tikzstyle{every node}=[font=\footnotesize]
        \begin{axis}  
        [
            ybar,  
            grid style={line width=.1pt, draw=gray!10},major grid style={line width=.2pt,draw=gray!50},
            ymajorgrids,
            yminorgrids,
            minor y tick num=2,
            bar width=0.4cm,
            x=1cm,
            enlarge x limits=0.25,
            ylabel style={align=center},
            ylabel={Path Length Reduction \%}, 
            symbolic x coords={Mean, 75$^{\text{th}}$, 98$^{\text{th}}$}, 
            xtick=data,  
            nodes near coords={\pgfmathprintnumber\pgfplotspointmeta \%},
            nodes near coords align={vertical},  
            ymin=0,
            ymax=32,
            height=5.5cm,
            width=0.3\textwidth,
            xticklabel style={align=center, text height=0.1cm},
            legend columns=3,
            legend style={at={(0.5,1.2)},anchor=north,
            /tikz/column 2/.style={
                column sep=10pt,
            },
            /tikz/column 4/.style={
                column sep=10pt,
            },},
            legend image code/.code={%
                    \draw[#1, draw=none] (0cm,-0.1cm) rectangle (0.4cm,0.2cm);
            }, 
            ]  
        \addplot[bargreenline, fill=bargreen] coordinates {(Mean, 11) (75$^{\text{th}}$, 9) (98$^{\text{th}}$, 26)}; 
        \legend{vs Tesseract}
        \end{axis}  
    \end{tikzpicture}\\
    
    \end{tabular}
    \caption{We compare the \emph{C-Space Path Length} across different methods in the above plots. On the left, we see that~\emph{cuRobo} has shorter path lengths than all other methods and~\emph{cuRobo-GP} has shorter path than~\emph{Tesseract-GP}. On the right, we plot the reduction in path length with~\emph{cuRobo} over~\emph{Tesseract} across our test dataset.}
    \label{fig:path-length}
\end{figure}

Next, we plot~\emph{C-Space Path Length} across the four methods in Figure~\ref{fig:path-length}. First, we observe that our geometric planner~\emph{cuRobo-GP} has shorter path length than~\emph{Tesseract-GP} which uses RRTConnect.~\emph{cuRobo-GP}'s path lengths are [5.39, 7.13, 13.45] radians on mean, 75$^\text{th}$, and 98$^\text{th}$ percentile of the dataset compared to~\emph{Tesseract-GP}'s [7.1, 8, 17.2] radians. We then observe methods that use trajectory optimization,~\emph{cuRobo} and~\emph{Tesseract} have shorter path length than geometric planning methods~\emph{cuRobo-GP} and~\emph{Tesseract-GP}. \emph{Tesseract} reduces paths on average by 48\% when compared to~\emph{Tesseract-GP}. \emph{cuRobo} reduces the path length by 53\% on average when compared with~\emph{Tesseract-GP}.~\emph{cuRobo} reduces path length by 53\%, 39\%, and 10\% on average when compared to~\emph{Tesseract-GP},~\emph{cuRobo-GP}, and~\emph{Tesseract} respectively. When we compare~\emph{cuRobo} with \emph{Tesseract}, we found that~\emph{cuRobo}'s paths are 26\% shorter than \emph{Tesseract} on the \pn percentile of the dataset.

From the geometric planning metrics, we find that trajectory optimization methods have comparable success percentage to geometric planning methods while also providing shorter path lengths than RRTConnect-like planners. While optimal geometric planners can generate shorter paths than~RRTConnect, they require either more planning time or task-specific heuristics as shown in Appendix~\ref{app:rrtstar}. Our goal is not only to get shorter paths, but also trajectories that can be executed on robots with minimal post-processing. Geometric planning methods require time parameterization as a post-processing step to be able to execute on robots. Kunz and Stilman developed a bounded velocity and acceleration time-parameterization technique~\cite{kunz2012time} that is extensively used in the robotics community, including MoveIt~\cite{coleman2014reducing}. However, this time parameterization technique does not bound the jerk along the trajectory and as such can have very large jerks. We could not find any accessible software library that can post process geometric paths while bounding jerks, making geometric path planning not directly deploy-able on jerk sensitive manipulators. We hence only compare between~\emph{Tesseract} and~\emph{cuRobo} on the time parameterization metrics. We additionally take the trajectories obtained from~\emph{Tesseract} and post process with Kunz and Stilman's method, which we call~\emph{Tesseract-TG}. We add this to our comparisons to highlight the improvements we can get with trajectory optimization techniques, especially with~\emph{cuRobo}'s minimum jerk formulation.

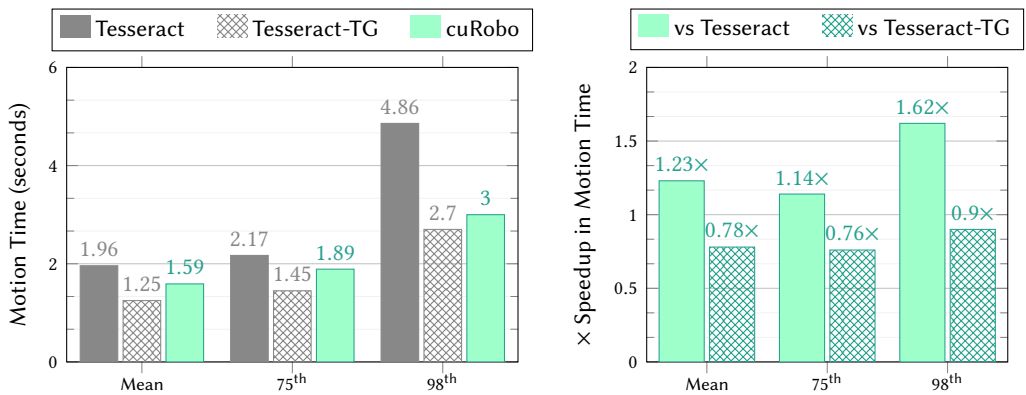
\begin{figure}[h]
    \centering
    \begin{tabular}{c c}
    \begin{tikzpicture}  
     \tikzstyle{every node}=[font=\footnotesize]
        \begin{axis}  
        [
            ybar,  
            grid style={line width=.1pt, draw=gray!10},major grid style={line width=.2pt,draw=gray!50},
            ymajorgrids,
            yminorgrids,
            minor y tick num=2,
            bar width=0.5cm,
            x=2cm,
            enlarge x limits=0.25,
            every node near coord/.append style={/pgf/number format/fixed},
            ylabel={Motion Time (seconds)},
            symbolic x coords={Mean, 75$^{\text{th}}$, 98$^{\text{th}}$}, 
            xtick=data,  
             nodes near coords,   
            nodes near coords align={vertical},  
            ymin=0,
            ymax=6,
            height=5.5cm,
            width=0.9\textwidth,
            xticklabel style={align=center, text height=0.1cm},
            legend columns=4,
            legend style={at={(0.5,1.2)},anchor=north,
            /tikz/column 2/.style={
                column sep=10pt,
             },
             /tikz/column 4/.style={
                column sep=10pt,
            },
            /tikz/column 6/.style={
                column sep=10pt,
            },
            },
            legend image code/.code={%
                    \draw[#1, draw=none] (0cm,-0.1cm) rectangle (0.4cm,0.2cm);
            }, 
            legend cell align={left},
            ]  
        \addplot[barorangeline, fill=barorange] coordinates {(Mean, 1.96)(75$^{\text{th}}$, 2.17) (98$^{\text{th}}$, 4.86)};       
        \addplot[barorangeline, fill=barorange,
        pattern color=barorange, pattern = crosshatch] coordinates {(Mean, 1.25)(75$^{\text{th}}$, 1.45) (98$^{\text{th}}$, 2.7)};       
        \addplot[bargreenline, fill=bargreen] coordinates {(Mean, 1.59)(75$^{\text{th}}$, 1.89) (98$^{\text{th}}$, 3.)};       
        \legend{Tesseract,Tesseract-TG, cuRobo}
        \end{axis}  
    \end{tikzpicture}
    &
    \begin{tikzpicture}  
     \tikzstyle{every node}=[font=\footnotesize]
        \begin{axis}  
        [
            ybar,  
            grid style={line width=.1pt, draw=gray!10},major grid style={line width=.2pt,draw=gray!50},
            ymajorgrids,
            yminorgrids,
            minor y tick num=2,
            bar width=0.6cm,
            x=1.6cm,
            enlarge x limits=0.25,
            ylabel style={align=center},
            ylabel={$\times$ Speedup in Motion Time}, 
            symbolic x coords={Mean, 75$^{\text{th}}$, 98$^{\text{th}}$}, 
            xtick=data,  
            nodes near coords={\pgfmathprintnumber\pgfplotspointmeta $\times$},
            nodes near coords align={vertical},  
            ymin=0,
            ymax=2,
            height=5.5cm,
            width=0.3\textwidth,
            xticklabel style={align=center, text height=0.1cm},
            legend columns=3,
            legend style={at={(0.5,1.2)},anchor=north,
            /tikz/column 2/.style={
                column sep=10pt,
            },
            /tikz/column 4/.style={
                column sep=10pt,
            },},
            legend image code/.code={%
                    \draw[#1, draw=none] (0cm,-0.1cm) rectangle (0.4cm,0.2cm);
            }, 
            ]  
        
        \addplot[bargreenline, fill=bargreen] coordinates {(Mean, 1.23) (75$^{\text{th}}$, 1.14) (98$^{\text{th}}$, 1.62)}; 
        \addplot[bargreenline, fill=bargreen,
        pattern color=bargreenline, pattern = crosshatch] coordinates {(Mean, 0.78) (75$^{\text{th}}$, 0.76) (98$^{\text{th}}$, 0.9)};
        \legend{vs Tesseract, vs Tesseract-TG}
        \end{axis}  
    \end{tikzpicture}\\
    \end{tabular}
    \caption{We compare the~\emph{Motion Time} (i.e., length of the trajectory in time) generated by~\emph{cuRobo} with~\emph{Tesseract} and also to~\emph{Tesseract-TG} which post processes the trajectory with a time-optimal reparameterization developed by Kunz and Stilman~\cite{kunz2012time}. On the left plot, we see that~\emph{cuRobo} generates trajectories that are within 3 seconds on the \pn percentile while~\emph{Tesseract} generates significantly slower trajectories at 5 seconds. On the right plot, we observe that~\emph{cuRobo} obtains trajectories that are 1.62$\times$ faster than \emph{Tesseract}. \emph{cuRobo} is within 0.9$\times$ of \emph{Tesseract-TG} motion time while having 9$\times$ lower jerk.}
    \label{fig:motion-time}
\end{figure}
\begin{figure}
    \centering
    \begin{tabular}{c c}
\begin{tikzpicture}  
     \tikzstyle{every node}=[font=\footnotesize]
        \begin{axis}  
        [
            ybar,  
            grid style={line width=.1pt, draw=gray!10},major grid style={line width=.2pt,draw=gray!50},
            ymajorgrids,
            yminorgrids,
            minor y tick num=2,
            bar width=0.5cm,
            x=2cm,
            enlarge x limits=0.25,
            every node near coord/.append style={/pgf/number format/fixed},
            ylabel={Max. Jerk (rad. s$^{-3}$)},
            symbolic x coords={Mean, 75$^{\text{th}}$, 98$^{\text{th}}$}, 
            xtick=data,  
             nodes near coords,   
            nodes near coords align={vertical},  
            ymin=0,
            ymax=1200,
            height=5.5cm,
            width=0.9\textwidth,
            xticklabel style={align=center},
            legend columns=4,
            legend style={at={(0.5,1.2)},anchor=north,
            /tikz/column 2/.style={
                column sep=10pt,
             },
             /tikz/column 4/.style={
                column sep=10pt,
            },
            /tikz/column 6/.style={
                column sep=10pt,
            },
            },
            legend image code/.code={%
                    \draw[#1, draw=none] (0cm,-0.1cm) rectangle (0.4cm,0.2cm);
            }, 
            legend cell align={left},
            ]  
        \addplot[barorangeline, fill=barorange] coordinates {(Mean, 200)(75$^{\text{th}}$, 239) (98$^{\text{th}}$, 501)};       
        \addplot[barorangeline, fill=barorange,  pattern color=barorange, pattern = crosshatch] coordinates {(Mean, 602)(75$^{\text{th}}$, 595) (98$^{\text{th}}$, 970)};       
        \addplot[barorangeline, fill=bargreen] coordinates {(Mean, 49)(75$^{\text{th}}$, 63) (98$^{\text{th}}$, 136)};       
        \legend{Tesseract,Tesseract-TG, cuRobo}
        \end{axis}  
    \end{tikzpicture}
    &
    \begin{tikzpicture}  
     \tikzstyle{every node}=[font=\footnotesize]
        \begin{axis}  
        [
            ybar,  
            grid style={line width=.1pt, draw=gray!10},major grid style={line width=.2pt,draw=gray!50},
            ymajorgrids,
            yminorgrids,
            minor y tick num=2,
            bar width=0.5cm,
            x=1.5cm,
            enlarge x limits=0.25,
            ylabel style={align=center},
            ylabel={$\times$ lower Max. Jerk}, 
            symbolic x coords={Mean, 75$^{\text{th}}$, 98$^{\text{th}}$}, 
            nodes near coords={\pgfmathprintnumber\pgfplotspointmeta $\times$},
            xtick=data,  
            nodes near coords align={vertical},  
            ymin=0,
            ymax=18,
            height=5.5cm,
            width=0.3\textwidth,
            xticklabel style={align=center},
            legend columns=3,
            legend style={at={(0.5,1.2)},anchor=north,
            /tikz/column 2/.style={
                column sep=10pt,
            },
            /tikz/column 4/.style={
                column sep=10pt,
            },},
            legend image code/.code={%
                    \draw[#1, draw=none] (0cm,-0.1cm) rectangle (0.4cm,0.2cm);
            }, 
            ]  
        
        \addplot[bargreenline, fill=bargreen] coordinates {(Mean, 4) (75$^{\text{th}}$, 3.8) (98$^{\text{th}}$, 3.7)}; 
        \addplot[bargreenline, fill=bargreen, 
        pattern color=bargreenline, pattern = crosshatch] coordinates {(Mean, 12) (75$^{\text{th}}$, 9.4) (98$^{\text{th}}$, 7)};
        \legend{vs Tesseract, vs Tesseract-TG}
        \end{axis}  
    \end{tikzpicture}\\
    \end{tabular}
    \caption{The maximum jerk along the trajectory across all joints is shown in the plots. On the left plot, we see that~\emph{cuRobo} has a maxumum jerk of 136~rad.$s^{-3}$ on the \pn percentile of the dataset while~\emph{Tesseract} has 501 rad.$s^{-3}$ a 7x higher value as shown by the plot on the right.}
    \label{fig:max-jerk}
\end{figure}
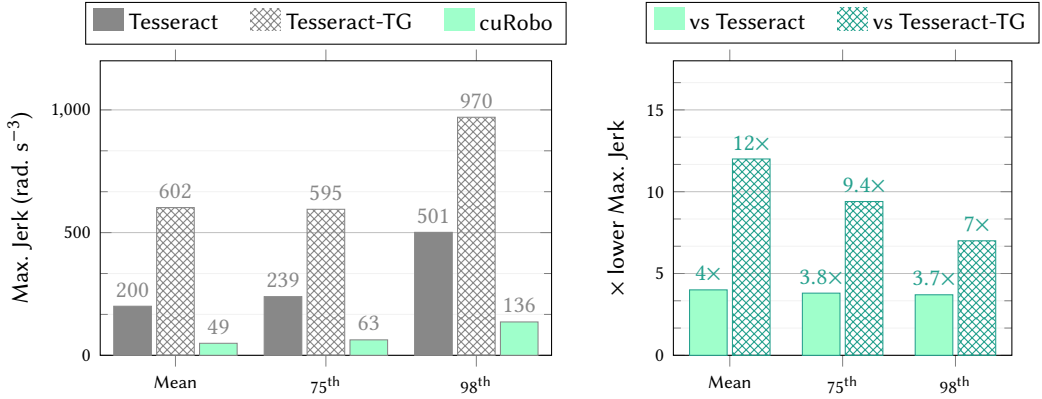

The first in time parameterization metrics is~\emph{Motion Time} which we plot in Figure~\ref{fig:motion-time}. \emph{cuRobo}'s trajectories take [1.59, 1.89, 3] seconds compared to Tesseract's [1.96, 2.17, 4.86] seconds on the mean, \ps, and \pn~percentiles. \emph{cuRobo} produces trajectories that have a 1.23x lower mean and 1.62x lower \pn~percentile motion time when compared to motions generated by~\emph{Tesseract}. This large reduction in motion time leads to \emph{cuRobo}'s \pn~percentile being 2.86 seconds quicker than~\emph{Tesseract}. When we compare~\emph{cuRobo}'s solutions to~\emph{Tesseract-TG} which uses time-optimal reparameterization,~\emph{cuRobo} generates trajectories that are 0.3 seconds slower both on average and \pn percentile of the dataset. This slow down is because~\emph{cuRobo} also optimizes for minimum-jerk, leading to trajectories with 12x lower jerk on average compared to~\emph{Tesseract-TG} as shown in Fig.~\ref{fig:max-jerk}. When comparing to~\emph{Tesseract}, we generate trajectories that have 4x lower jerk on average as~\emph{Tesseract} doesn't minimize jerk.

\begin{figure}
    \centering
    \begin{tabular}{c c}
    \begin{tikzpicture}  
     \tikzstyle{every node}=[font=\footnotesize]
        \begin{axis}  
        [
            ybar,  
            grid style={line width=.1pt, draw=gray!10},major grid style={line width=.2pt,draw=gray!50},
            ymajorgrids,
            yminorgrids,
            minor y tick num=2,
            bar width=0.45cm,
            x=1.9cm,
            enlarge x limits=0.25,
            every node near coord/.append style={/pgf/number format/fixed},
            ylabel={Max. Acceleration (rad. $s^{-2}$)},
            symbolic x coords={Mean, 75$^{\text{th}}$, 98$^{\text{th}}$}, 
            xtick=data,  
             nodes near coords,   
            nodes near coords align={vertical},  
            ymin=0,
            ymax=30,
            height=5.5cm,
            width=0.49\textwidth,
            xticklabel style={align=center},
            legend columns=4,
            legend style={at={(0.5,1.2)},anchor=north,
            /tikz/column 2/.style={
                column sep=10pt,
             },
             /tikz/column 4/.style={
                column sep=10pt,
            },
            /tikz/column 6/.style={
                column sep=10pt,
            },
            },
            legend image code/.code={%
                    \draw[#1, draw=none] (0cm,-0.1cm) rectangle (0.4cm,0.2cm);
            }, 
            legend cell align={left},
            ]  
        \addplot[barorangeline, fill=barorange] coordinates {(Mean,9.9)(75$^{\text{th}}$, 11.2) (98$^{\text{th}}$, 23.9)};       
        \addplot[barorangeline, fill=barorange, fill=barorange,  pattern color=barorange, pattern = crosshatch] coordinates {(Mean, 15)(75$^{\text{th}}$, 15) (98$^{\text{th}}$, 15)};       
        \addplot[bargreenline, fill=bargreen] coordinates {(Mean, 6.82)(75$^{\text{th}}$, 8) (98$^{\text{th}}$, 12)};       
        \legend{Tesseract,Tesseract-TG, cuRobo}
        \end{axis}  
    \end{tikzpicture}
    &
\begin{tikzpicture}  
     \tikzstyle{every node}=[font=\footnotesize]
        \begin{axis}  
        [
            ybar,  
            grid style={line width=.1pt, draw=gray!10},major grid style={line width=.2pt,draw=gray!50},
            ymajorgrids,
            yminorgrids,
            minor y tick num=2,
            bar width=0.45cm,
            x=1.9cm,
            enlarge x limits=0.25,
            every node near coord/.append style={/pgf/number format/fixed},
            ylabel={Mean Velocity (rad. $s^{-1}$)},
            symbolic x coords={Mean, 75$^{\text{th}}$, 98$^{\text{th}}$}, 
            xtick=data,  
             nodes near coords,   
            nodes near coords align={vertical},  
            ymin=0,
            ymax=1.4,
            height=5.5cm,
            width=0.49\textwidth,
            xticklabel style={align=center},
            legend columns=4,
            legend style={at={(0.5,1.2)},anchor=north,
            /tikz/column 2/.style={
                column sep=10pt,
             },
             /tikz/column 4/.style={
                column sep=10pt,
            },
            /tikz/column 6/.style={
                column sep=10pt,
            },
            },
            legend image code/.code={%
                    \draw[#1, draw=none] (0cm,-0.1cm) rectangle (0.4cm,0.2cm);
            }, 
            legend cell align={left},
            ]  
        \addplot[barorangeline, fill=barorange] coordinates {(Mean, 0.58)(75$^{\text{th}}$, 0.68) (98$^{\text{th}}$, 0.89)};       
        \addplot[barorangeline, fill=barorange,  pattern color=barorange, pattern = crosshatch] coordinates {(Mean, 0.87)(75$^{\text{th}}$, 0.98) (98$^{\text{th}}$, 1.2)};
        \addplot[bargreenline, fill=bargreen] coordinates {(Mean, 0.6)(75$^{\text{th}}$, 0.69) (98$^{\text{th}}$, 0.89)};       
        \legend{Tesseract,Tesseract-TG, cuRobo}
        \end{axis}  
    \end{tikzpicture}
    \\
    \end{tabular}
    \caption{The maximum acceleration and mean velocity across the dataset is plotted for the different methods.~\emph{cuRobo} has the smallest acceleration~(the left plot) across the methods while having similar mean velocity to~\emph{Tesseract} on the right plot.}
    \label{fig:mean-velocity}
\end{figure}
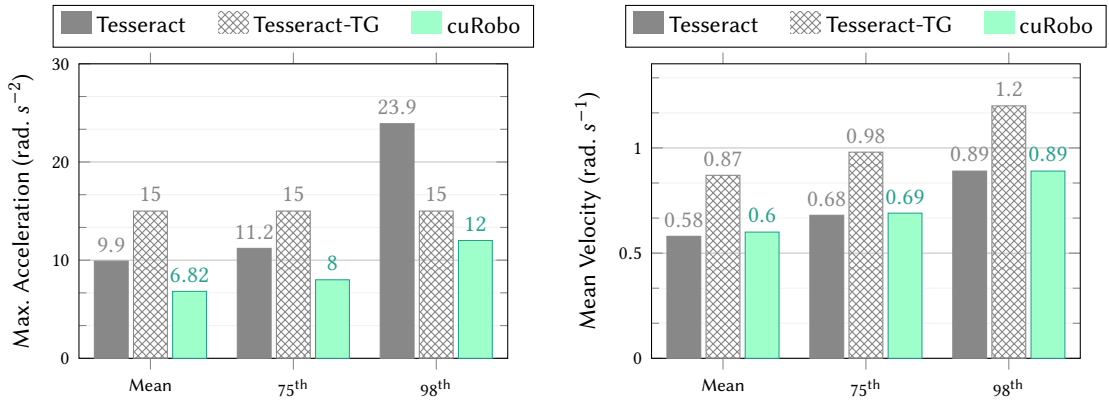

We compare the mean velocity and maximum acceleration between methods in Figure~\ref{fig:mean-velocity}. \emph{Tesseract-TG} has the largest values in mean velocity and maximum acceleration as Kunz and Stilman's time parameterization technique by design attempts to reach peak velocity by instantly jumping to maximum acceleration. \emph{Tesseract} has larger max acceleration when compared to \emph{cuRobo} as it doesn't have to optimize for jerk and as such can instantaneously change acceleration along the trajectory without any penalties. We also observed that~\emph{Tesseract} has a \pn~percentile maximum acceleration of 23.9 rad.s$^{-2}$ which is beyond the 15 rad.s$^{-2}$ acceleration limit we set for the Franka Panda robot. \emph{cuRobo} has the smallest maximum acceleration across the dataset as we minimize jerk across the trajectory, which penalizes instantaneous changes to acceleration. Even with the smallest maximum acceleration, \emph{cuRobo}'s mean velocity is comparable to \emph{Tesseract}, slightly higher in the mean by 0.02 rad.s$^{-1}$ and 0.01 rad.s$^{-1}$ in the \ps~percentile.

Summarizing across these metrics, \emph{cuRobo} produces paths that are shorter in path length than any other method while also succeeding on all feasible problems in the dataset. In addition,~\emph{cuRobo} generates trajectories that have the least jerk, 4x lower than existing trajectory optimization techniques, and 12x lower than existing time parameterization methods. Trajectories generated with \emph{cuRobo} also have motion times 1.23x lower than existing trajectory optimization techniques and is within 0.3 seconds of high-jerk time parameterization methods. We will next analyze the compute time taken by \emph{cuRobo} to obtain these trajectories.

\subsection{Compute Time}
\label{sec:mb-time}
We calculate the compute time on three platforms, a PC with an AMD Ryzen 9 7950x CPU and NVIDIA RTX 4090 GPU, and an NVIDIA Jetson AGX Orin 64GB system configured to operate at MAXN (60W) and 15W power budgets. We measure runtime using Python's {\it time} utility after synchronizing device and host. 

\subsubsection{Motion Generation}

\begin{figure}
    \centering
    \begin{tikzpicture}  
     \tikzstyle{every node}=[font=\footnotesize]
        \begin{axis}  
        [
            ybar,  
            grid style={line width=.1pt, draw=gray!10},major grid style={line width=.2pt,draw=gray!50},
            ymajorgrids,
            yminorgrids,
            minor y tick num=2,
            bar width=0.5cm,
            x=4.5cm,
            enlarge x limits=0.25,
            every node near coord/.append style={/pgf/number format/fixed},
            ylabel={Compute Time (s)},
            symbolic x coords={Mean, \ps, \pn}, 
            xtick=data,  
             nodes near coords,   
            nodes near coords align={vertical},  
            ymin=0,
            ymax=52,
            height=5.5cm,
            width=0.99\textwidth,
            xticklabel style={align=center},
            legend columns=2,
            legend style={at={(0.5,1.47)},anchor=north,
            /tikz/column 2/.style={
                column sep=10pt,
            },
            },
            legend image code/.code={%
                    \draw[#1, draw=none] (0cm,-0.1cm) rectangle (0.4cm,0.2cm);
            }, 
            legend cell align={left},
            ]  
        \addplot[barorangeline, fill=barorange] coordinates {(Mean, 2.95) (\ps, 2.47) (\pn, 22)}; 
       
        \addplot[bargreenline, fill=bargreen] coordinates {(Mean, 0.049) (\ps, 0.034) (\pn, 0.26)}; 
         
         \addplot[barorangeline, fill=barorange, fill=barorange,  pattern color=barorange, pattern = north east lines] coordinates {(Mean, 6.13) (\ps, 6.68) (\pn, 32.3)}; 
        \addplot[bargreenline, fill=bargreenlight,  pattern color=bargreenline, pattern = north east lines] coordinates {(Mean, 0.22) (\ps, 0.13) (\pn, 1.34)}; 
         \addplot[barorangeline, fill=barorange, pattern color=barorange, pattern = crosshatch] coordinates {(Mean, 10.3) (\ps, 12.6) (\pn, 44.7)}; 
         \addplot[bargreenline, fill=bargreen,  pattern color=bargreenline, pattern = crosshatch] coordinates {(Mean, 0.48) (\ps, 0.26) (\pn, 3.3)}; 
         \legend{Tesseract-PC, cuRobo-PC, Tesseract-ORIN-MAXN, cuRobo-ORIN-MAXN, 
                Tesseract-ORIN-15W, cuRobo-ORIN-15W}
                
        \end{axis}  
    \end{tikzpicture}
    \\ 
    \begin{tikzpicture}  
     \tikzstyle{every node}=[font=\footnotesize]
        \begin{axis}  
        [
            ybar,  
            grid style={line width=.1pt, draw=gray!10},major grid style={line width=.2pt,draw=gray!50},
            ymajorgrids,
            yminorgrids,
            minor y tick num=2,
            bar width=0.5cm,
            x=4.5cm,
            enlarge x limits=0.25,
            ylabel style={align=center},
            ylabel={$\times$ Speedup vs Tesseract}, 
            symbolic x coords={Mean, \ps, \pn}, 
            xtick=data,  
            nodes near coords={\pgfmathprintnumber\pgfplotspointmeta $\times$},
            nodes near coords align={vertical},  
            ymin=0,
            ymax=100,
            height=5.5cm,
            width=0.99\textwidth,
            xticklabel style={align=center},
            legend columns=3,
            legend style={at={(0.5,1.2)},anchor=north,
            /tikz/column 2/.style={
                column sep=10pt,
            },
            /tikz/column 4/.style={
                column sep=10pt,
            },},
            legend image code/.code={%
                    \draw[#1, draw=none] (0cm,-0.1cm) rectangle (0.4cm,0.2cm);
            }, 
            ]  
        \addplot[bargreenline, fill=bargreen] coordinates {(Mean, 60) (\ps, 72)(\pn, 83)};
        \addplot[bargreenline, fill=bargreen,  pattern color=bargreenline, pattern = north east lines] coordinates {(Mean, 28) (\ps, 53)(\pn, 23)};
        \addplot[bargreenline, fill=bargreen, fill=barorange,  pattern color=bargreenline, pattern = crosshatch] coordinates {(Mean, 21) (\ps, 49)(\pn, 14)};
        \end{axis}  
    \end{tikzpicture}
    \caption{We compare the compute time for motion generation between \emph{cuRobo} and \emph{Tesseract} across three compute platforms. On all of the 2600 motion planning problems, we found \emph{cuRobo} to take the least time, getting a 60$\times$ speedup on average on a desktop pc with NVIDIA RTX 4090 and AMD Ryzen 9 7950x, with a 83$\times$ speedup on the \pn~percentile.}
    \label{fig:motion-gen-speedup}
\end{figure}
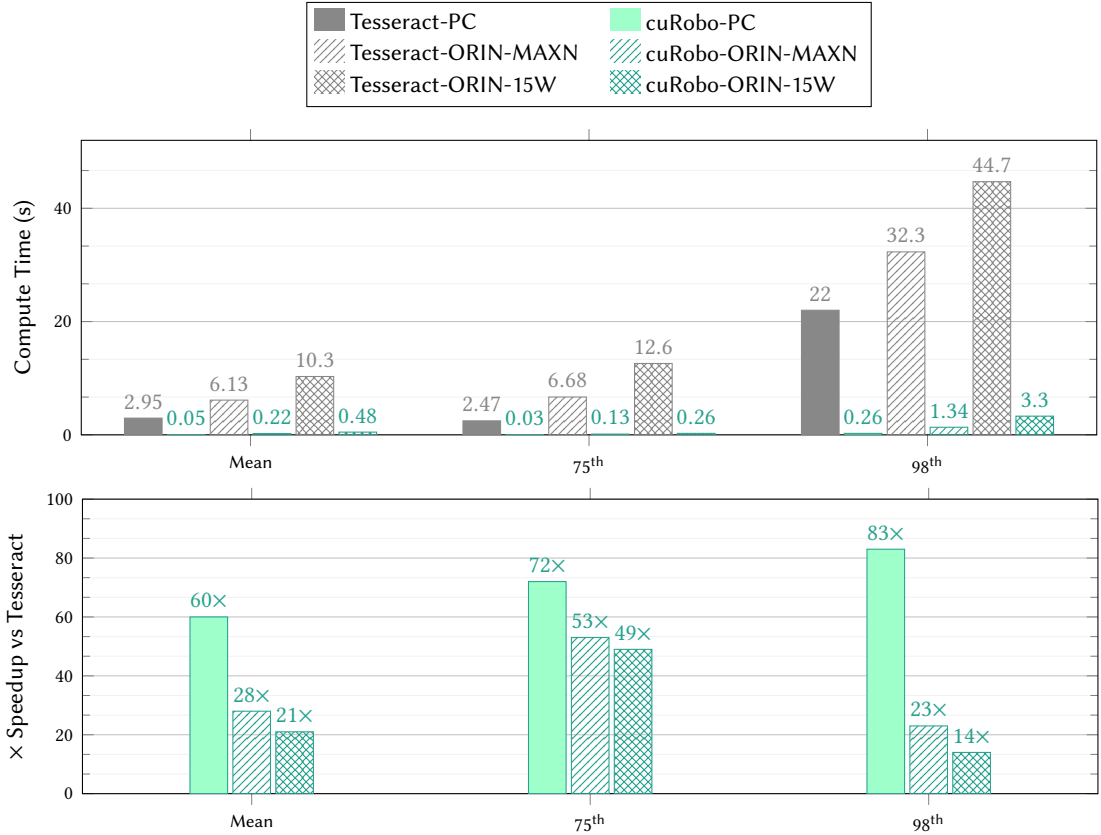

The time it takes to compute motions using~\emph{cuRobo} is compared to~\emph{Tesseract} in Figure~\ref{fig:motion-gen-speedup} across all three platforms. We observed that~\emph{Tesseract} takes on average 2.95 seconds compared to~\emph{cuRobo} taking 50ms, leading to a 60$\times$ speedup in motion planning. The gap in planning time increases at the \ps~percentile of evaluation set, \emph{cuRobo} taking 30ms to plan while \emph{Tesseract} takes 2.47 seconds, leading to a 72$\times$ speedup in planning with our proposed method. On the \pn~percentile of the dataset, \emph{cuRobo} takes 260 milliseconds while \emph{Tesseract} takes 22 seconds, giving \emph{cuRobo} 83$\times$ speedup in planning. The difference in planning time across the mean, \ps, and \pn is because \emph{cuRobo} calls the geometric planner only after three failed attempts with linear seeds.

\begin{figure}
    \centering
    \includegraphics[width=0.8\textwidth, trim=0 9.3cm 0 0, clip]{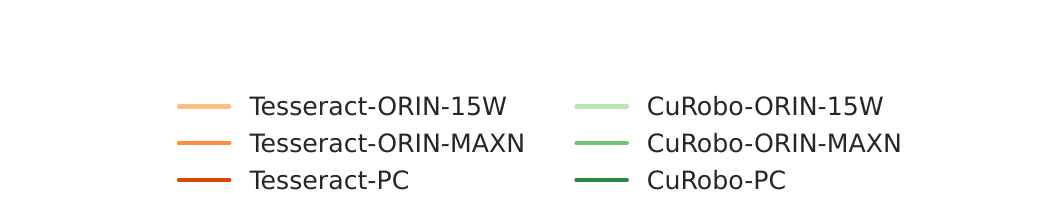}\\
    \includegraphics[width=\textwidth]{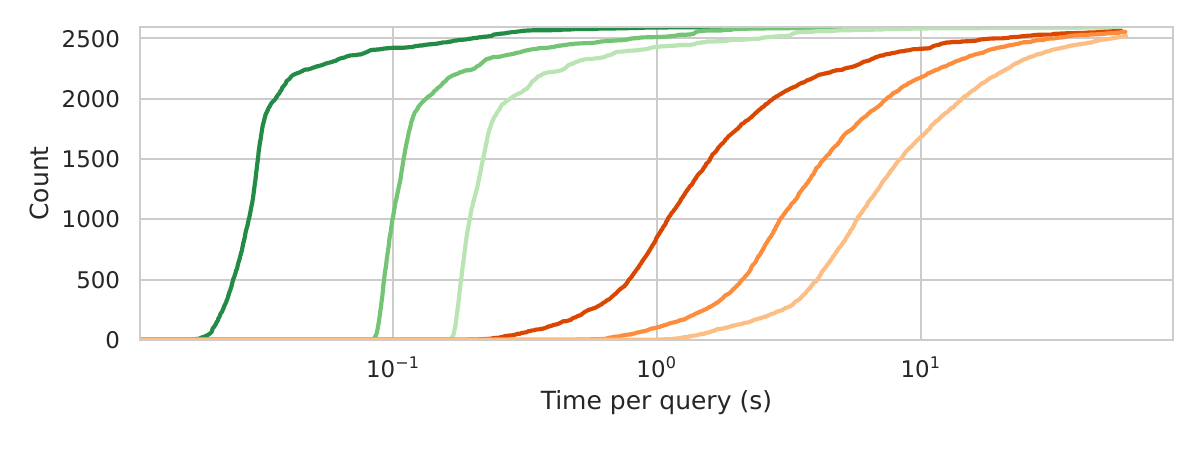}
    \caption{Motion generation time across compute platforms for \emph{Tesseract} and \emph{cuRobo} is shown with the x-axis in logarithmic scale to better highlight the difference in time across the dataset. \emph{cuRobo} at 15W on the NVIDIA Jetson ORIN AGX~(cuRobo-ORIN-15W) is faster than \emph{Tesseract} on a full desktop PC with an i7 processor~(Tesseract-i7) as seen by large gap in x-axis across the full dataset. In addition, we see that~\emph{cuRobo} slows down on the Jetson compared to a NVIDIA RTX 4090.} 
    \label{fig:compute_motion_opt}
\end{figure}

The speedup over~\emph{Tesseract} scales to the NVIDIA Jetson ORIN as well, in both power modes as shown in Figure~\ref{fig:motion-gen-speedup}. \emph{cuRobo} takes 0.22 seconds and 0.48 seconds on average on at MAXN and 15w while \emph{Tesseract} takes 6.13 seconds and 10.3 seconds respectively. On average, \emph{cuRobo} is 28$\times$ and 21$\times$ faster than \emph{Tesseract} at MAXN and 15W respectively. We also oberved that~\emph{cuRobo} is faster on a NVIDIA Jetson ORIN at 15W than \emph{Tesseract} running on a desktop PC as shown in Figure~\ref{fig:compute_motion_opt}.

\begin{figure}
    \centering
    \begin{tikzpicture} 
     \tikzstyle{every node}=[font=\footnotesize]
    \begin{axis}[%
    hide axis,
    xmin=10,
    xmax=50,
    ymin=0,
    ymax=0.4,
    legend columns=2,
    legend style={at={(0.5,0.5)},anchor=north,
            /tikz/column 2/.style={
                column sep=10pt,
            },
            },
            legend image code/.code={%
                    \draw[#1, draw=none] (0cm,-0.1cm) rectangle (0.4cm,0.2cm);
            }, 
            legend cell align={left},
    ]
    \addlegendimage{area legend,fill=bargreen,draw=bargreenline}
    \addlegendentry{Full Planning Time};
    \addlegendimage{area legend,fill=barorange,draw=barorangeline}
    \addlegendentry{Solve Time};
    \end{axis}
\end{tikzpicture}
\\
\begin{tabular}{c c}
\begin{tikzpicture}  
     \tikzstyle{every node}=[font=\footnotesize]
        \begin{axis}  
        [
            ybar,  
            grid style={line width=.1pt, draw=gray!10},major grid style={line width=.2pt,draw=gray!50},
            ymajorgrids,
            yminorgrids,
            minor y tick num=2,
            bar width=0.5cm,
            x=2cm,
            enlarge x limits=0.25,
            every node near coord/.append style={/pgf/number format/fixed},
            ylabel={Compute Time (ms)},
            symbolic x coords={Mean, \ps, \pn}, 
            xtick=data,  
             nodes near coords,   
            nodes near coords align={vertical},  
            ymin=0,
            ymax=300,
            height=4.5cm,
            width=\textwidth,
            xticklabel style={align=center},
            legend columns=2,
            legend style={at={(0.5,1.4)},anchor=north,
            /tikz/column 2/.style={
                column sep=10pt,
            },
            },
            legend image code/.code={%
                    \draw[#1, draw=none] (0cm,-0.1cm) rectangle (0.4cm,0.2cm);
            }, 
            legend cell align={left},
            ]  
        
        \addplot[bargreenline, fill=bargreen] coordinates {(Mean, 50) (\ps, 34) (\pn, 260)}; 
        \addplot[barorangeline, fill=barorange] coordinates {(Mean, 42) (\ps, 29) (\pn, 230)}; 
                
        \end{axis}  
\end{tikzpicture}
&
\begin{tikzpicture}  
     \tikzstyle{every node}=[font=\footnotesize]
        \begin{axis}  
        [
            ybar,  
            grid style={line width=.1pt, draw=gray!10},major grid style={line width=.2pt,draw=gray!50},
            ymajorgrids,
            yminorgrids,
            minor y tick num=2,
            bar width=0.4cm,
            x=2cm,
            enlarge x limits=0.25,
            every node near coord/.append style={/pgf/number format/fixed},
            ylabel={Speedup vs Tesseract},
            symbolic x coords={Mean, \ps, \pn}, 
            xtick=data,  
             nodes near coords,   
            nodes near coords align={vertical},  
            nodes near coords={\pgfmathprintnumber\pgfplotspointmeta $\times$},
            ymin=0,
            ymax=105,
            height=4.5cm,
            width=\textwidth,
            xticklabel style={align=center},
            legend columns=2,
            legend style={at={(0.5,1.4)},anchor=north,
            /tikz/column 2/.style={
                column sep=10pt,
            },
            },
            legend image code/.code={%
                    \draw[#1, draw=none] (0cm,-0.1cm) rectangle (0.4cm,0.2cm);
            }, 
            legend cell align={left},
            ]  
        
        \addplot[bargreenline, fill=bargreen] coordinates {(Mean, 60) (\ps, 72) (\pn, 83)}; 
        \addplot[barorangeline, fill=barorange] coordinates {(Mean, 69) (\ps, 84) (\pn, 93)}; 
                
        \end{axis}  
\end{tikzpicture}

\end{tabular}
    \caption{We measure the time taken by \emph{cuRobo} in the optimization iterations~\emph{Solve Time} and compare it to the time taken by the full pipeline. As our library is implemented in python, we take measurable amount of time outside of the solver iterations which could be reduced by rewriting in a compiled programming language.}
    \label{fig:motion-gen-opt-time}
\end{figure}
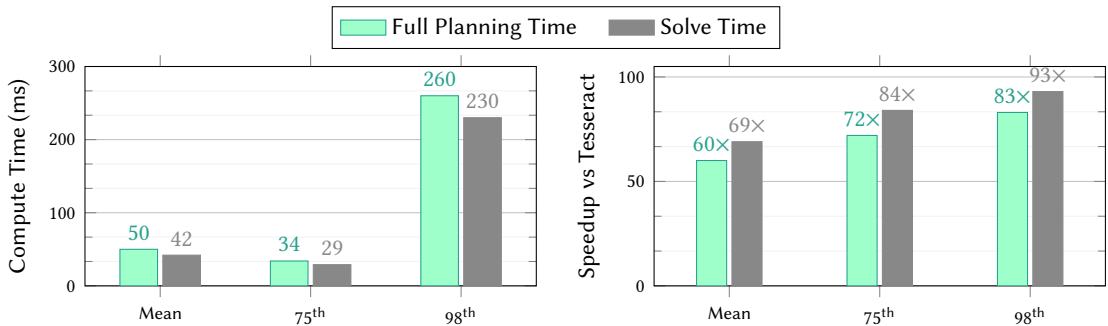
Our approach is implemented in python with key compute kernels in CUDA C++ called through python wrappers. We reduced the python overhead and the overhead of repeatedly launching CUDA kernels by recording optimization iterations and the \texttt{mask\_samples} function in geometric planning (see Algorithm~\ref{alg:steer}) in CUDA Graphs. We then replay the recorded CUDA Graphs with data from new planning problems. This use of CUDA Graphs reduced our planning time by 10x compared to calling the kernels individually from python. Our implementation still has some components in python, calling many small cuda kernels to setup the optimization problems, and also to get the final result from the many parallel seeds. We timed these parts and found that~\emph{cuRobo} spent 8ms, 5ms, and 30ms on the mean, \ps, and \pn~percentile as shown in Figure~\ref{fig:motion-gen-opt-time}. The percentage of time spent in these steps compared to the solver time was 15\%, 15\%, and 12\% on average, \ps, and \pn~percentiles. We leave speeding up these components by rewriting directly in C++ and fusing the CUDA kernels to future work. We do share the speedup that could be gained if these components are optimized by only comparing the iterations time and geometric planning time to \emph{Tesseract} in Figure~\ref{fig:motion-gen-opt-time}. We see that our speedup of 60$\times$ becomes 69$\times$ on mean, 72$\times$ becomes 84$\times$ on \ps~percentile, and 83$\times$ becomes 93$\times$ on the \pn~percentile.

\subsubsection{Geometric Planning}
\begin{figure}
    \centering
   \begin{tikzpicture}  
     \tikzstyle{every node}=[font=\footnotesize]
        \begin{axis}  
        [
            ybar,  
            grid style={line width=.1pt, draw=gray!10},major grid style={line width=.2pt,draw=gray!50},
            ymajorgrids,
            yminorgrids,
            minor y tick num=2,
            bar width=0.5cm,
            x=4.5cm,
            enlarge x limits=0.25,
            every node near coord/.append style={/pgf/number format/fixed},
            ylabel style={align=center},
            ylabel={Geometric Planning \\Time (s)},
            symbolic x coords={Mean, \ps, \pn}, 
            xtick=data,  
             nodes near coords,   
            nodes near coords align={vertical},  
            ymin=0,
            ymax=30,
            height=4.5cm,
            width=0.9\textwidth,
            xticklabel style={align=center},
            legend columns=2,
            legend style={at={(0.5,1.5)},anchor=north,
            /tikz/column 2/.style={
                column sep=10pt,
            },
            },
            legend image code/.code={%
                    \draw[#1, draw=none] (0cm,-0.1cm) rectangle (0.4cm,0.2cm);
            }, 
            legend cell align={left},
            ]  
         \addplot[barorangeline, fill=barorange] coordinates {(Mean, 1.5) (\ps, 0.34) (\pn, 20)}; 
        \addplot[bargreenline, fill=bargreen] coordinates {(Mean, 0.015) (\ps, 0.021) (\pn, 0.035)}; 
         \addplot[barorangeline, fill=barorange, fill=barorange,  pattern color=barorange, pattern = north east lines] coordinates {(Mean, 2.04) (\ps, 0.89) (\pn, 24.36)}; 
        \addplot[bargreenline, fill=bargreenlight,  pattern color=bargreenline, pattern = north east lines] coordinates {(Mean, 0.087) (\ps, 0.098) (\pn, 0.367)}; 
         \addplot[barorangeline, fill=barorange, pattern color=barorange, pattern = crosshatch] coordinates {(Mean,2.74) (\ps, 1.72) (\pn, 25.9)}; 
         \addplot[bargreenline, fill=bargreen,  pattern color=bargreenline, pattern = crosshatch] coordinates {(Mean, 0.165) (\ps, 0.185) (\pn, 0.707)}; 
         \legend{RRTConnect-PC, cuRobo-PC, RRTConnect-ORIN-MAXN, cuRobo-ORIN-MAXN, 
                RRTConnect-ORIN-15W, cuRobo-ORIN-15W}
        \end{axis}  
    \end{tikzpicture}
    \\ 
    \begin{tikzpicture}  
     \tikzstyle{every node}=[font=\footnotesize]
        \begin{axis}  
        [
            ybar,  
            grid style={line width=.1pt, draw=gray!10},major grid style={line width=.2pt,draw=gray!50},
            ymajorgrids,
            yminorgrids,
            minor y tick num=2,
            bar width=0.5cm,
            x=4.5cm,
            enlarge x limits=0.25,
            ylabel style={align=center},
            ylabel={$\times$ Speedup\\vs RRTConnect}, 
            symbolic x coords={Mean, \ps, 98},
            xtick=data,  
            nodes near coords={\pgfmathprintnumber\pgfplotspointmeta $\times$},
            nodes near coords align={vertical},  
            ymin=0,
            ymax=150,
            height=4.5cm,
            width=0.9\textwidth,
            xticklabel style={align=center},
            legend columns=3,
            legend style={at={(0.5,1.2)},anchor=north,
            /tikz/column 2/.style={
                column sep=10pt,
            },
            /tikz/column 4/.style={
                column sep=10pt,
            },},
            legend image code/.code={%
                    \draw[#1, draw=none] (0cm,-0.1cm) rectangle (0.4cm,0.2cm);
            },
            xticklabels={Mean, \ps, \pn},
            ]  

       \addplot[bargreenline, fill=bargreen] coordinates {(Mean,101) (\ps, 16)(98, 581)};
       \draw [->, thick, black, xshift=-0.58cm] (98, 100) --(98, 150) node [pos=0, rotate=90, anchor=east] {\(581\times\)};
        \addplot[bargreenline, fill=bargreen,  pattern color=bargreenline, pattern = north east lines] coordinates {(Mean, 23) (\ps, 9)(98, 66)};
        \addplot[bargreenline, fill=bargreen, fill=barorange,  pattern color=bargreenline, pattern = crosshatch] coordinates {(Mean, 17) (\ps, 9)(98, 37)}; 
        \end{axis}  
    \end{tikzpicture}
    \caption{Our GPU accelerated geometric planner is 101$\times$ faster on average compared to OMPL's implementation of RRTConnect available in \emph{Tesseract}. We observed an average speedup of 23$\times$ and 17$\times$ on the NVIDIA Jetson Orin at MAXN and 15W modes respectively. We see a much larger speedup on the \pn~percentile of the dataset across the compute platforms as our GPU accelerated graph builder is able to explore the workspace in parallel while a CPU based approach scales linearly with edge validation.}
    \label{fig:graph-time}
\end{figure}
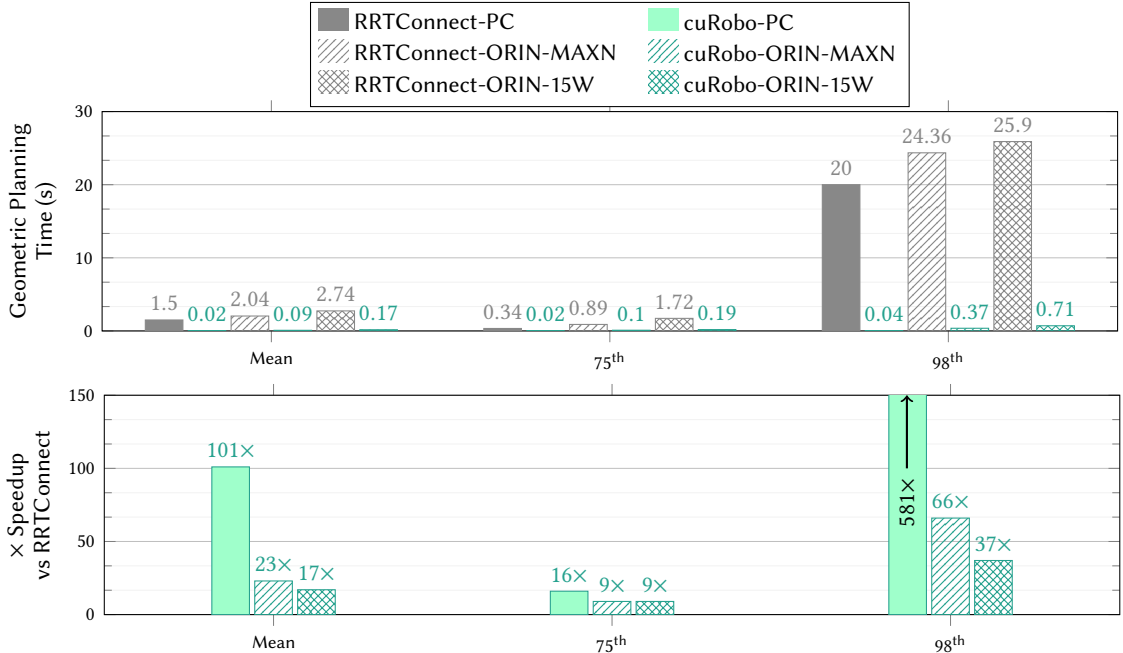

We compare the compute time in geometric planning between our GPU accelerated geometric planner introduced in Section~\ref{sec:graph} which we call~\emph{cuRobo-GP} to OMPL's RRTConnect implementation in Tesseract, which we call~\emph{Tesseract-GP} in Figure~\ref{fig:graph-time}. \emph{Tesseract-GP} takes 1.5 seconds on average while \emph{cuRobo-GP} takes 0.02 seconds leading to a 101$\times$ speedup in geometric planning with~\emph{cuRobo-GP}. Looking at the \pn~percentile planning time, \emph{Tesseract-GP} takes 20 seconds while~\emph{cuRobo-GP} takes 0.04 seconds, giving us a 581$\times$ speedup. A very recent work from Thomason~\etal~\cite{thomason2024vamp} that explores vectorized geometric planning leveraging SIMD on CPU. The results from their paper show that it takes 0.1ms (mean) to plan on the motion benchmaker dataset. However, their code is not available at the time of this publication and we leave comparing to it for a future work.

\subsubsection{Trajectory Optimization}
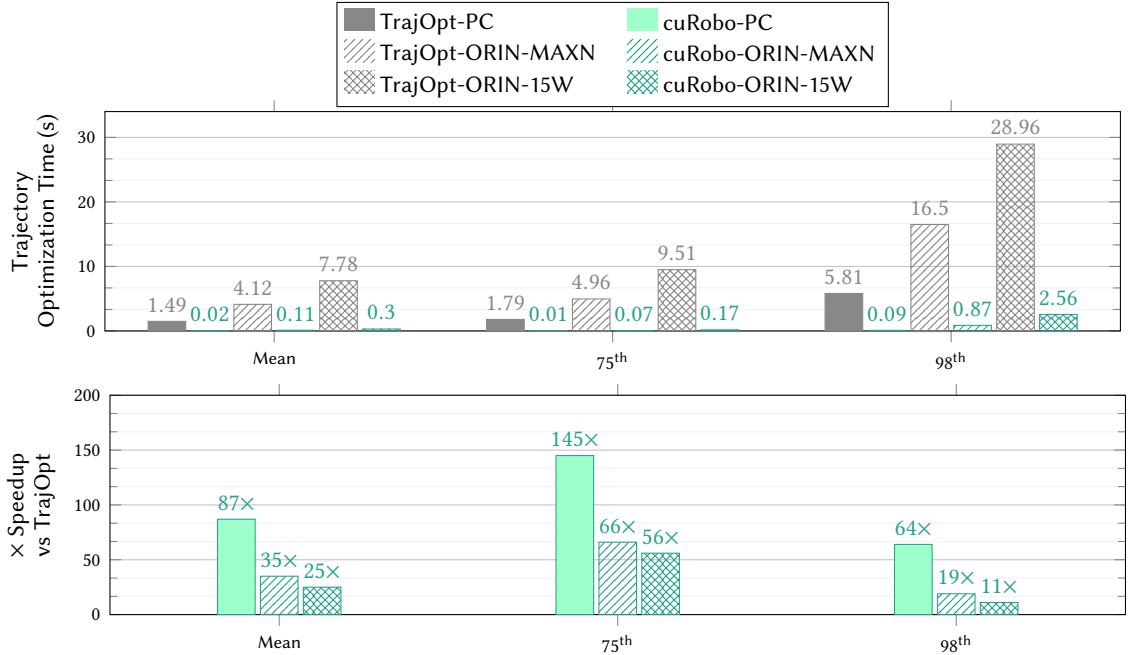
\begin{figure}
    \centering
     \begin{tikzpicture}  
     \tikzstyle{every node}=[font=\footnotesize]
        \begin{axis}  
        [
            ybar,  
            grid style={line width=.1pt, draw=gray!10},major grid style={line width=.2pt,draw=gray!50},
            ymajorgrids,
            yminorgrids,
            minor y tick num=2,
            bar width=0.5cm,
            x=4.5cm,
            enlarge x limits=0.25,
            every node near coord/.append style={/pgf/number format/fixed},
            ylabel style={align=center},
            ylabel={Trajectory \\Optimization Time (s)},
            symbolic x coords={Mean, \ps, \pn}, 
            xtick=data,  
             nodes near coords,   
            nodes near coords align={vertical},  
            ymin=0,
            ymax=34,
            height=4.5cm,
            width=0.9\textwidth,
            xticklabel style={align=center},
            legend columns=2,
            legend style={at={(0.5,1.5)},anchor=north,
            /tikz/column 2/.style={
                column sep=10pt,
            },
            },
            legend image code/.code={%
                    \draw[#1, draw=none] (0cm,-0.1cm) rectangle (0.4cm,0.2cm);
            }, 
            legend cell align={left},
            ]  
         \addplot[barorangeline, fill=barorange] coordinates {(Mean, 1.49) (\ps, 1.79) (\pn, 5.81)}; 
        \addplot[bargreenline, fill=bargreen] coordinates {(Mean, 0.017) (\ps, 0.012) (\pn, 0.09)}; 
        
         \addplot[barorangeline, fill=barorange, fill=barorange,  pattern color=barorange, pattern = north east lines] coordinates {(Mean, 4.12) (\ps, 4.96) (\pn, 16.5)}; 
        \addplot[bargreenline, fill=bargreenlight,  pattern color=bargreenline, pattern = north east lines] coordinates {(Mean, 0.11) (\ps, 0.07) (\pn, 0.87)}; 
         \addplot[barorangeline, fill=barorange, pattern color=barorange, pattern = crosshatch] coordinates {(Mean,7.78) (\ps, 9.51) (\pn, 28.96)}; 
         \addplot[bargreenline, fill=bargreen,  pattern color=bargreenline, pattern = crosshatch] coordinates {(Mean, 0.3) (\ps, 0.17) (\pn, 2.56)}; 
         \legend{TrajOpt-PC, cuRobo-PC, TrajOpt-ORIN-MAXN, cuRobo-ORIN-MAXN, 
                TrajOpt-ORIN-15W, cuRobo-ORIN-15W}
        \end{axis}  
    \end{tikzpicture}
    \\ 
    \begin{tikzpicture}  
     \tikzstyle{every node}=[font=\footnotesize]
        \begin{axis}  
        [
            ybar,  
            grid style={line width=.1pt, draw=gray!10},major grid style={line width=.2pt,draw=gray!50},
            ymajorgrids,
            yminorgrids,
            minor y tick num=2,
            bar width=0.5cm,
            x=4.5   cm,
            enlarge x limits=0.25,
            ylabel style={align=center},
            ylabel={$\times$ Speedup\\vs TrajOpt}, 
            nodes near coords={\pgfmathprintnumber\pgfplotspointmeta $\times$},
            symbolic x coords={Mean, \ps, \pn},
            xtick=data,  
            nodes near coords align={vertical},  
            ymin=0,
            ymax=200,
            height=4.5cm,
            width=0.9\textwidth,
            xticklabel style={align=center},
            legend columns=3,
            legend style={at={(0.5,1.2)},anchor=north,
            /tikz/column 2/.style={
                column sep=10pt,
            },
            /tikz/column 4/.style={
                column sep=10pt,
            },},
            legend image code/.code={%
                    \draw[#1, draw=none] (0cm,-0.1cm) rectangle (0.4cm,0.2cm);
            }, 
            ]  

       \addplot[bargreenline, fill=bargreen] coordinates {(Mean, 87) (\ps, 145)(\pn, 64)};
       
        \addplot[bargreenline, fill=bargreen,  pattern color=bargreenline, pattern = north east lines] coordinates {(Mean, 35) (\ps, 66)(\pn, 19)};
        \addplot[bargreenline, fill=bargreen, fill=barorange,  pattern color=bargreenline, pattern = crosshatch] coordinates {(Mean, 25) (\ps, 56)(\pn, 11)};
        \end{axis}  
    \end{tikzpicture}
    \caption{We compare our trajectory optimization to \emph{TrajOpt} which is integrated in \emph{Tesseract} for collision-free trajectory generation. On average, our approach is 87$\times$ faster than \emph{TrajOpt} on a desktop PC. On a NVIDIA Jetson ORIN, our appraoch is 35$\times$ and 25$\times$ faster at MAXN and 15W modes.}
    \label{fig:to-time}
\end{figure}

We compute the time it takes to perform trajectory optimization in Figure~\ref{fig:to-time} between~\emph{cuRobo}'s GPU accelerated approach and TrajOpt~\cite{schulman2014motion} implementation leveraged by~\emph{Tesseract}. We see that \emph{cuRobo} is 87$\times$ and 145$\times$ faster in optimization than TrajOpt on average and \ps~percentile respectively. \emph{cuRobo} takes a mere 10ms to perform trajectory optimization compared to TrajOpt taking 1.79 seconds on \ps~percentile of the evaluation set. We get speedups of 23$\times$ and 17$\times$ on Jetson device as well, taking 0.09 and 0.17 seconds on average on ORIN at MAXN and 15W respectively. These speedups are interesting because numerical optimization is predominantly iterative where we run many sequential iterations until convergence. These sequential iterations can make the entire computation graph memory access heavy. Our efficient parallelization of compute across the whole pipeline enables us to get these speedups. We not only run each seed of optimization in seperate threads but also split many of our workload heavy kernels across many threads on a per seed basis.

While these timings are based on optimizing collision-free trajectories, we hope that these speedups encourage roboticists to leverage \emph{cuRobo}'s trajectory optimization implementation for other robotics tasks.

\subsubsection{Inverse Kinematics}
\label{sec:inv-kin-benchmark}
We compare the performance of our inverse kinematics solver against TracIK~\cite{beeson2015trac}. We perform two sets of experiments: (1) without collision checking and (2) with self and environment collision checking. We chose an instance of the~\emph{bookshelf-small-panda} scene from motion benchmaker~\cite{chamzas2021motionbenchmaker} for the collision checking experiment. We sample feasible joint configs from a Halton Sequence and average the results across 5 trials for different batch sizes. Since TracIK does not account for collisions, we perform rejection sampling with PyBullet, allowing 10 reattempts. For all cuRobo IK queries, we run 30 seeds in parallel and return the best solution from these seeds. We evaluate IK with 5 different batch sizes -- 1, 10, 100, 500, and 1000. For a single query (batch size=1), cuRobo takes 2.7ms while TracIK only takes 0.9ms. However, as we increase the batch size of IK queries, we see a speedup starting from a batch size of 10.

\begin{figure}
    \centering
     \begin{tikzpicture}  
        \begin{axis}  
        [  
            ybar,  
            grid style={line width=.1pt, draw=gray!10},major grid style={line width=.2pt,draw=gray!50},
            ymajorgrids,
            ylabel={Solve Time (ms)}, %
            xlabel={Batch Size},  
            symbolic x coords={1, 10, 100, 1000}, %
            xtick=data,  
             nodes near coords, %
            nodes near coords align={vertical},  
            ymin=0, 
            ymax=200,
            height=4.5cm,
            width=0.99\textwidth,
            legend columns=4,
            legend style={at={(0.5,1.3)},anchor=north,
            /tikz/column 2/.style={
                column sep=10pt,
            },
            /tikz/column 4/.style={
                column sep=10pt,
            },
            /tikz/column 6/.style={
                column sep=10pt,
            },
            },
            legend image code/.code={%
                    \draw[#1, draw=none] (0cm,-0.2cm) rectangle (0.4cm,0.2cm);
            }, 
            legend cell align={left},
            bar width=0.62cm,
            x=3.25cm,
            enlarge x limits=0.2,
        ]  
        \addplot[barorangeline, fill=barorange, pattern color=barorangeline, pattern = north west lines] coordinates {(1, 0.9) (10, 6.3) (100, 70) (1000, 629)
        }; 
        \addplot[bargreenline, fill=bargreen, bargreenline,  pattern color=bargreenline, pattern = north east lines] coordinates {(1, 2.7) (10, 2.8) (100, 4.1)  (1000, 27)
        };  
        \addplot[barorangeline, fill=barorange] coordinates {(1, 1.61) (10, 117) (100, 638) (1000,10469)
        };
        \addplot[bargreenline, fill=bargreen] coordinates {(1, 3.9) (10,4.87) (100, 14.7)  (1000, 131)
        };  
        \draw [->, thick, white, xshift=0.32cm] (100, 125) -- (100, 175)
        node [pos=0, rotate=90, anchor=east] {\(638\)};
        \draw [->, thick, white, xshift=0.32cm] (1000, 125) -- (1000, 175)
        node [pos=0, rotate=90, anchor=east] {\(10469\)};
        \draw [->, thick, black, xshift=-1cm] (1000, 125) -- (1000, 175)
        node [pos=0, rotate=90, anchor=east] {\(629\)};
        
        \legend{TracIK, cuRobo,
        TracIK-Cfree, cuRobo-Cfree}
        \end{axis}  
        \end{tikzpicture}   
    \\
    
      \begin{tikzpicture}  
        \begin{axis}  
        [  
            ybar,  
            grid style={line width=.1pt, draw=gray!10},major grid style={line width=.2pt,draw=gray!50},
            ymajorgrids,
            ylabel={Queries Per Second (hz)},
            xlabel={Batch Size},  
            symbolic x coords={1, 10, 100, 1000}, %
            xtick=data,  
            nodes near coords, %
            every node near coord/.append style={/pgf/number format/.cd, fixed,1000 sep={}},
            nodes near coords align={vertical},  
            ymin=0, 
            ymax=45000,
            height=4.5cm,
            width=0.99\textwidth,
            bar width=0.62cm,
            x=3.25cm,
            enlarge x limits=0.2,
            ]  
        \addplot[barorangeline, fill=barorange,  pattern color=barorangeline, pattern = north west lines] coordinates {(1, 1112) (10,1592) (100, 1427) (1000, 1590)
        }; 
        \addplot[bargreenline, fill=bargreen,  pattern color=bargreenline, pattern = north east lines] coordinates {(1, 370) (10, 3525) (100, 24335)  (1000, 37134)
        };  
        
        \addplot[barorangeline, fill=barorange] coordinates {(1, 620) (10, 85) (100, 156) (1000,95)
        }; 
        
        \addplot[bargreenline, fill=bargreen] coordinates {
        (1, 254) (10,2052) (100, 6827)  (1000, 7611)
        };  
        
        \end{axis}  
        \end{tikzpicture}
   \\
    \begin{tikzpicture}  
        \begin{axis}  
        [  
            ybar,  
            grid style={line width=.1pt, draw=gray!10},major grid style={line width=.2pt,draw=gray!50},
            ymajorgrids,
            ylabel={$\times$ speedup}, %
            xlabel={Batch Size},  
            symbolic x coords={1, 10, 100, 1000}, %
            xtick=data,  
             nodes near coords, %
            nodes near coords align={vertical},  
            ymin=0, 
            ymax=100,
            height=4.5cm,
            width=0.99\textwidth,
            bar width=0.75cm,
            x=3.25cm,
            enlarge x limits=0.2,
            legend columns=2,
            nodes near coords={\pgfmathprintnumber\pgfplotspointmeta $\times$},
            legend style={at={(0.5,1.3)},anchor=north,
            /tikz/column 2/.style={
                column sep=10pt,
            },
            },
            legend image code/.code={%
                    \draw[#1, draw=none] (0cm,-0.2cm) rectangle (0.4cm,0.2cm);
            }, 
            legend cell align={left},
            ]  
        \addplot[bargreenline, fill=bargreen, pattern color=bargreenline, pattern = north east lines] coordinates {(1, 0.33) (10, 2.25) (100, 17)(1000, 23.4)
        }; 
         \addplot[bargreenline, fill=bargreen] coordinates {(1, 0.41) (10, 24) (100, 43.4)  (1000, 80)
        };  
        \legend{IK, Collision-Free IK}
        \end{axis}  
        \end{tikzpicture}  
        \\
\caption{We compare the compute time in solving inverse kinematics~(IK) between \emph{cuRobo} and TracIK~\cite{beeson2015trac} across different batch sizes. \emph{TracIK} is able to solve 1000 poses in 629ms while \emph{cuRobo} only takes 27ms, 23.4$\times$ faster than \emph{TracIK}. In solving collision-free IK, \emph{cuRobo} solves 1000 poses in 131ms, 80$\times$ faster than rejection sampled~\emph{TracIK}.} 
\label{fig:ablation}
\end{figure}
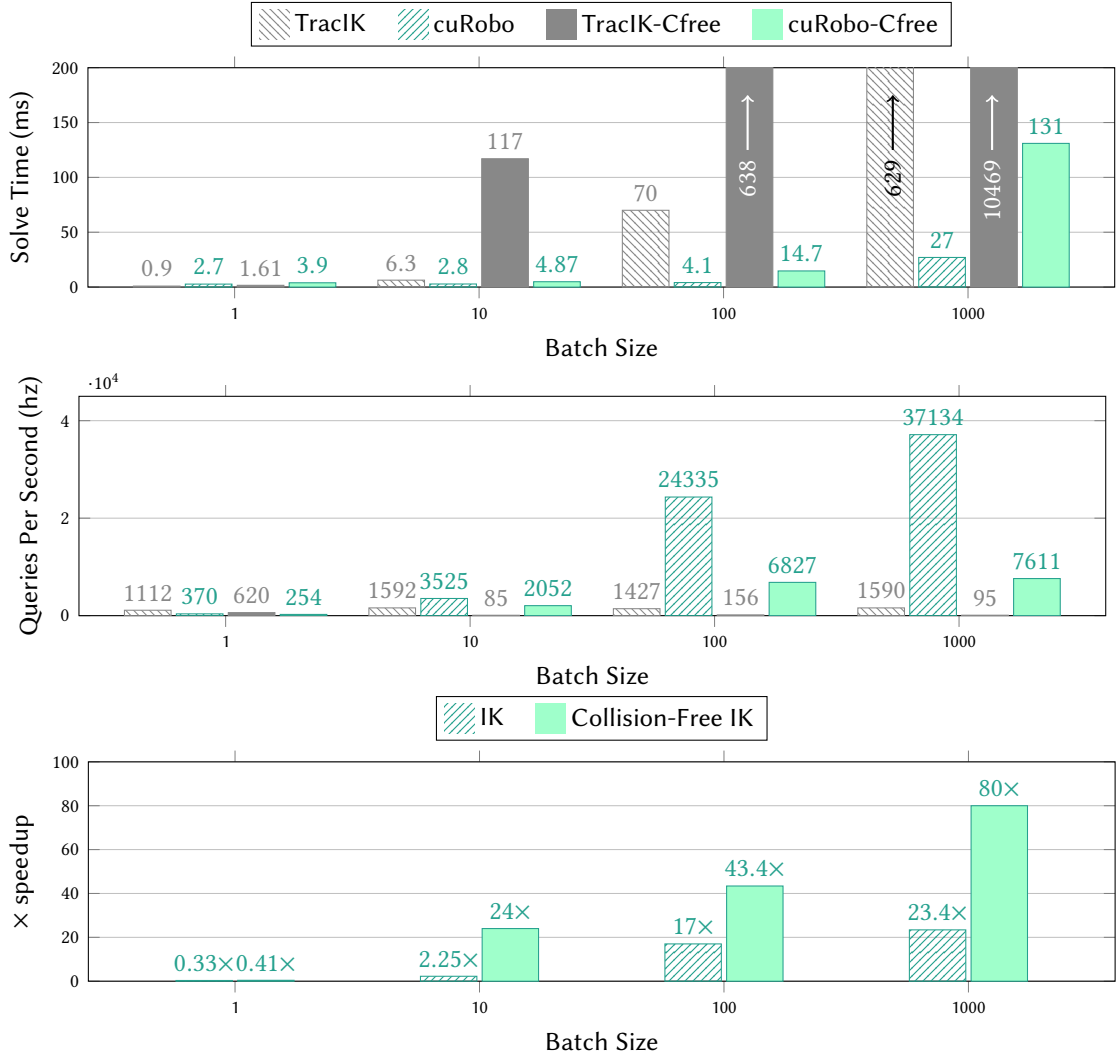

For the standard IK problem, we can generate 37134 solutions per second when we use a batch size of 1000 while TracIK can only generate 1590 solutions, 23.4$\times$ slower than our method. When we compare collision-free IK, our method~(\emph{cuRobo-Coll-Free}) can compute 7611 compared to rejection sampled Trac-IK (\emph{TracIK-Coll-Free}) which can only obtain 95 collision-free solutions per second in our experiments, 80x slower than our approach to collision-free IK as shown in Fig.~\ref{fig:motion_opt_plot}-B. We also found that rejection sampling approach to collision-free IK failed on 20\% of the problems tested. BioIK~\cite{starke2018memetic} reports that their approach can solve IK in 0.7ms (1428 solutions per second), it is not clear from their paper whether the runtime includes collision-free IK. Even if we consider their timing to be for collision-free IK, our method is still faster starting from a batch size of 10, taking 0.48ms per solution.

\begin{figure}
    \centering
    \includegraphics[width=\textwidth]{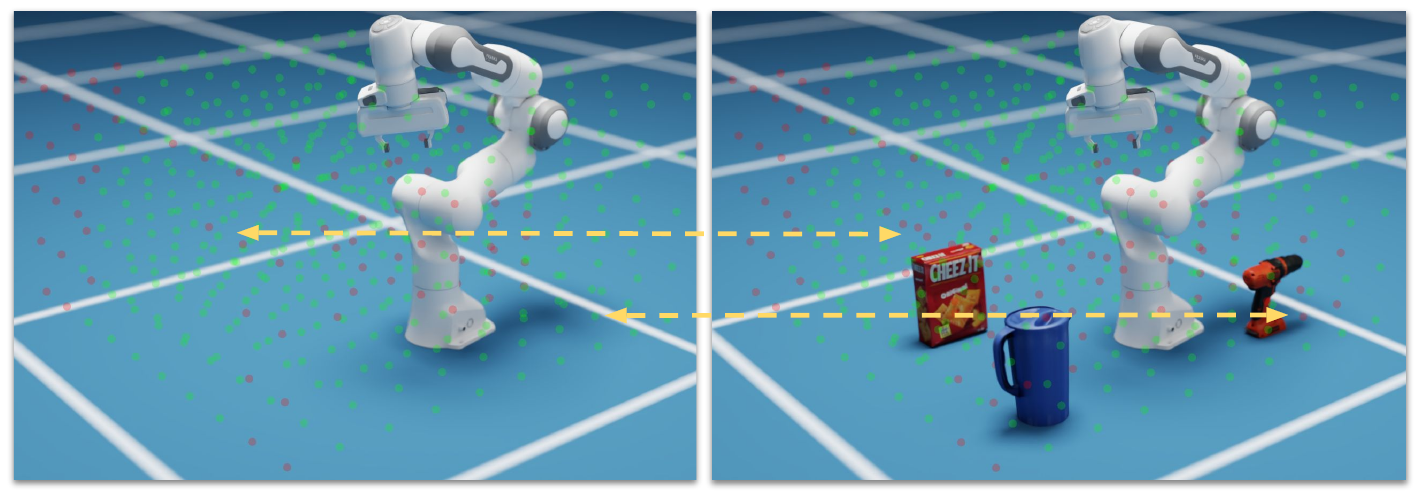}
    \caption{We show reachability analysis as an application of fast batched inverse kinematics in these images. We sample 500 poses in a grid shown by red and green spheres, and query \emph{cuRobo}'s batched inverse kinematics solver for joint configurations to match these poses. We color the spheres as red if IK was unsuccessful and green if successful. \emph{cuRobo} is able to run at 15Hz while also sharing the GPU resources with NVIDIA Isaac Sim. On the right we show our solver also reasoning about world collisions and marking poses near objects with red as they are not collision-free.}
    \label{fig:ik-reachability}
\end{figure}

\subsubsection{Kinematics and Distance Queries}
\label{sec:kin-benchmark}
\begin{figure}
    \centering
    \includegraphics[width=0.98\textwidth]{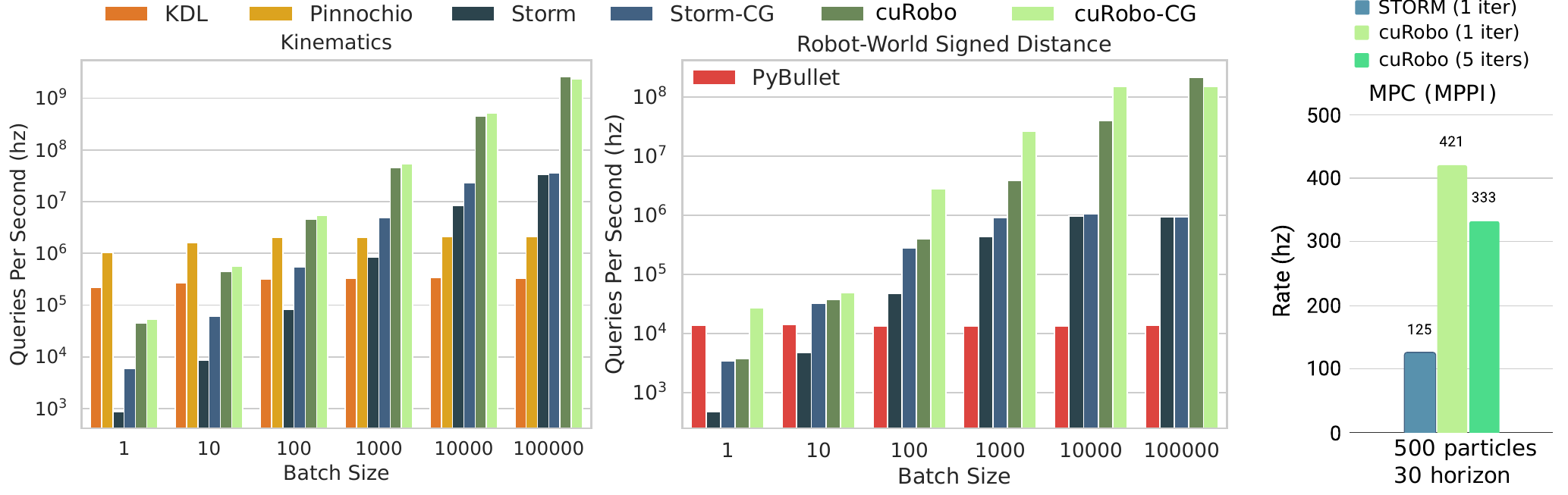}
    \caption{In the left plot, we see our forward kinematics match CPU methods at a batch size of 100 and becomes faster by upto 891x on 100k.
    In the middle plot, we see that our distance queries is upto 16,000x faster than \emph{PyBullet} as the batch size grows to 100k. Note that the y-axis is in log scale plots. Leveraging our faster kinematics and distance queries, we can run MPPI at upto 421hz, 3.36$\times$ faster than STORM~\cite{bhardwaj2021leveraging}.}
    \label{fig:kin_plot}
\end{figure}

One of the major breakthroughs in accelerating our motion generation approach was on developing a parallel compute friendly implementation of robot kinematics. Most common manipulators have many serially connected links, making computation of kinematics a largely serial operation. Existing SOTA methods for forward kinematics on CPUs such as pinnochio~\cite{carpentier2019pinocchio} take 1$\mu$s on average for 7-dof robots while GPU accelerated kinematics implemented in PyTorch such as STORM~\cite{storm2021} outmatch CPU methods only at a batch size of 1000. STORM improves upon implementation from Meier~\etal~\cite{meier2022differentiable} by keeping buffers in memory between calls without recreating them. This slowdown in GPU based kinematics is because existing implementations use many CUDA kernels to perform kinematics, e.g., STORM runs through 125 CUDA kernels to compute kinematics. In \emph{cuRobo}, we implement the entire kinematics in a single CUDA kernel, discussed in Appendix~\ref{app:cuda_kernels}. This enables our approach to outmatch pinnochio's performance at a batch size of 100 as shown in Figure~\ref{fig:kin_plot}. We also compare our kinematics implementation with KDL's implementation which is used by traciky. We additionally show the improvement with CUDA Graphs in calling GPU methods by adding a suffix ``-CG''. 

For signed distance queries, we compare with two prior methods -- PyBullet which uses Bullet to compute the signed distance~\cite{coumans2021} and STORM. We are faster beginning at a batch size of 1 as our approach uses many parallel threads on the GPU for a single query.

\subsubsection{Summary}
\begin{figure}
\centering
\begin{tikzpicture}
     \tikzstyle{every node}=[font=\footnotesize]

    \begin{axis}  
    [  
        ybar,  
        grid style={line width=.1pt, draw=gray!10},major grid style={line width=.2pt,draw=gray!50},
        ymajorgrids,
        yminorgrids,
        minor y tick num=2,
        ylabel={Median Time (ms)}, 
        xlabel={},  
        symbolic x coords={Kinematics, Collision\\Checking, Inverse\\Kinematics, Collision-Free\\Inverse\\Kinematics, Geometric\\Planning, Trajectory\\Optimization, 
        Motion\\Generation\\PC, Motion\\Generation\\ORIN-MAXN}, 
        xtick=data,  
         nodes near coords, %
        nodes near coords align={vertical},  
        nodes near coords={\pgfmathprintnumber\pgfplotspointmeta},
        ymin=0,
        ymax=120,
        height=4.5cm,
        width=\textwidth,
        xticklabel style   = {align=center},
        bar width=0.5cm,
        x=1.6cm,
        ]  
    \addplot[bargreenline, fill=bargreen] coordinates {
    (Kinematics, 1e-6)
    (Collision\\Checking, 1e-5)
    (Inverse\\Kinematics, 0.027) (Collision-Free\\Inverse\\Kinematics, 0.13) 
    (Geometric\\Planning, 16) (Trajectory\\Optimization, 10)     
    (Motion\\Generation\\PC, 30)
    (Motion\\Generation\\ORIN-MAXN, 100)};
    
\end{axis}  
\end{tikzpicture}    
\caption{We plot the median compute time across the many modules we have developed in \emph{cuRobo}. The compute time is within 100ms across all modules, with kinematics taking 1 nanosecond, and collision checking taking 10 nanoseconds when using a batch size of 100k. Inverse Kinematics takes 27 $\mu$seconds and collision-free inverse kinematics takes 130 $\mu$seconds with a batch size of 1000 on a NVIDIA RTX 4090. Trajectory optimization takes 10ms, followed by geometric planning taking 16ms, and motion generation taking 30ms across the benchmarking dataset.}
\label{fig:speedup}
\end{figure}
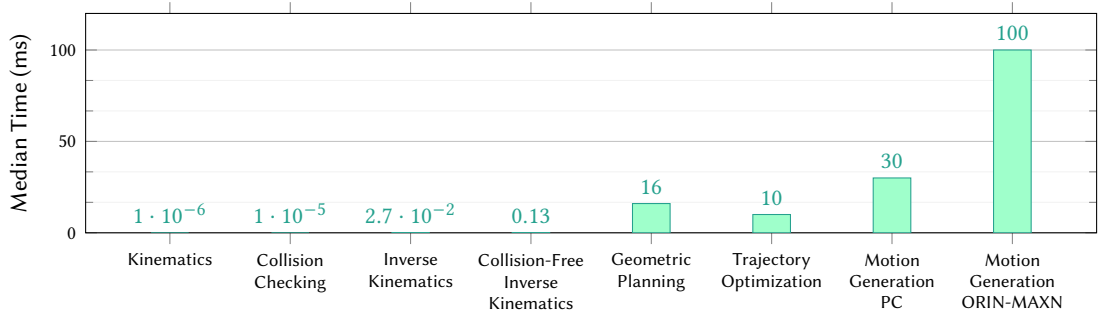

We summarize the median compute time across the different modules in \emph{cuRobo} in Figure~\ref{fig:speedup}. \emph{cuRobo}'s implementation of kinematics and collision checking can compute within 1 nanosecond and 10 nanoseconds respectively when using a batch size of 100k. For general inverse kinematics and collision-free inverse kinematics, \emph{cuRobo} can compute within 27 $\mu$seconds and 130 $\mu$seconds. This low computation time can accelerate existing robotics pipelines that use inverse kinematics such as reachability analysis~\cite{makhal2018reuleaux} and placement planning~\cite{murali2023cabinet}. Geometric planning has been used in verifying transition feasibility in hierarchical planning~\cite{zhu2021hierarchical} and task and motion planning~(TAMP)~\cite{garrett2020pddlstream}, where the quality of solutions is not critical and knowing if a path exists is sufficient. For these applications, leveraging~\emph{cuRobo}'s geometric planning can lead to a 101$\times$ speedup compared to using OMPL's RRTConnect algorithm. Our implementation of collision-free trajectory optimization takes 10ms, which could accelerate other robotics problems. The full motion generation pipeline already runs at 30ms on median on a modern PC. In addition, \emph{cuRobo}'s motion generation scales well to a NVIDIA AGX Orin running at 60W, taking only 100ms enabling deployment of motion generation on edge devices. \emph{cuRobo} obtains these low compute times while having most of it's stack in Python and with components implemented as separate modules. One could get even better compute times by fusing the modules at the CUDA kernel level, however that can make the library very rigid and inaccessible to robot practitioners. 

\subsection{Real Robot Tracking Performance}
\label{sec:real-robot-tracking}

\begin{figure}[b]
    \centering
    \includegraphics[width=\textwidth, trim= 0 0 0.25cm 0, clip]{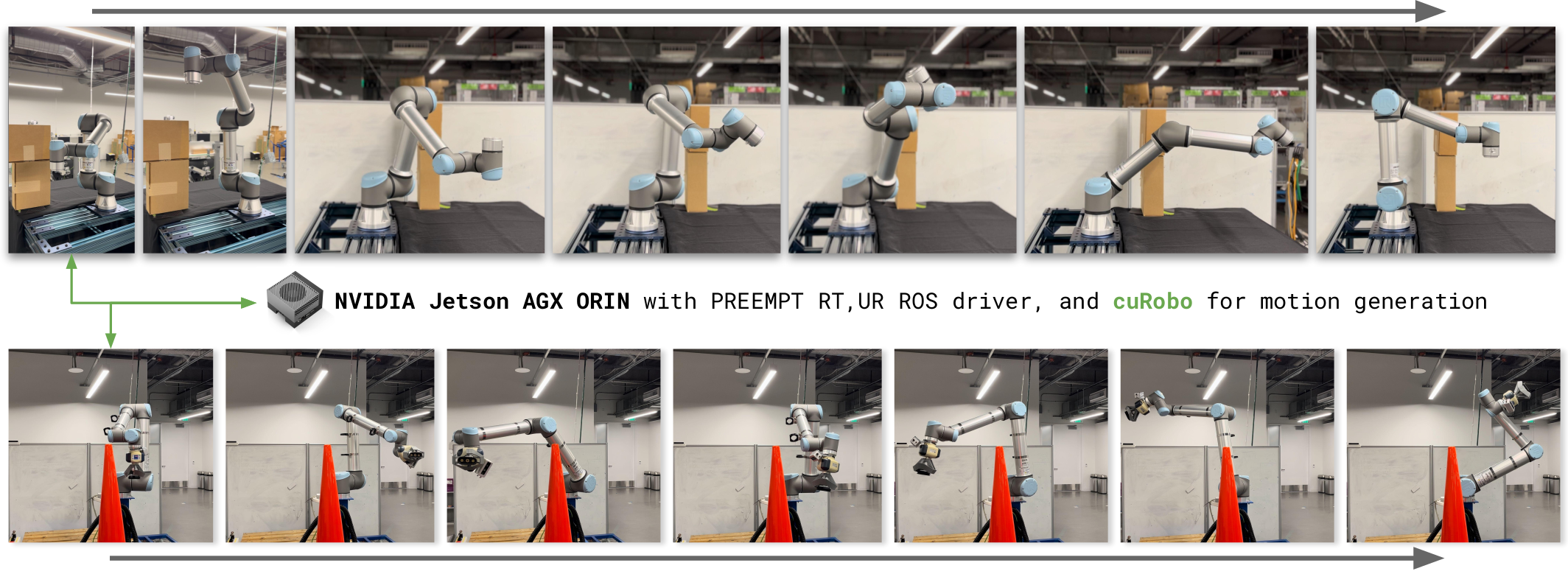}     
    \caption{We generate motions using~\emph{cuRobo} to reach the 7 poses shown here on the UR5e in the top and UR10 in the bottom. We use a NVIDIA Jetson AGX ORIN running at MAXN to generate motions using \emph{cuRobo} and send the trajectories to the robot.}
    \label{fig:ur5-poses}
\end{figure}

We study the tracking performance between the generated motions and executed motions by running \emph{cuRobo} on two Universal Robots, an UR5e robot and an UR10 robot. We connect the robots to a NVIDIA Jetson ORIN AGX which is running a PREEMPT RT kernel and uses the ros driver from universal robots for communication. We run~\emph{cuRobo} on the same Jetson device and send the generated trajectories to Universal Robot's trajectory tracking controller. For both robots, we setup an obstacle and selected seven random poses scattered around the robot's workspace, which are shown in Figure~\ref{fig:ur5-poses}. We then run motion generation to reach these seven poses in sequence five times, leading to a total of 35 reaching motion trials. The robots use an absolute magnetic encoder and an optical encoder together to measure the joint position. The accuracy of the magnetic encoder is +-0.0017 radians as obtained from~\cite{aksim}. We could not obtain the accuracy of the optical encoder and also the overall accuracy of the joint position measurement. We hence assume for all discussion below that the joint position is accurate up to 0.0017 radians.

We generate motions using~\emph{cuRobo} in two different modes, starting with~\emph{min-acc} where~\emph{cuRobo} performs trajectory optimization with acceleration minimization, without any jerk minimization, followed by~\emph{min-jerk} where we minimize jerk along with acceleration. In both these modes, the trajectory is optimized over 32 timesteps, interpolated to a 0.01 second resolution, and sent to the robot. We found no difference between sending an interpolated trajectory (at a 0.01 resolution) and the optimized coarse trajectory~(32 steps) to the UR10 and the UR5e. However, when executing trajectories with a low-level controller on many common robots, it might be necessary to interpolate the trajectory to a finer resolution before execution. We hence run all our experiments with interpolated trajectories.

\begin{figure}
    \centering
    \includegraphics[width=0.99\textwidth]{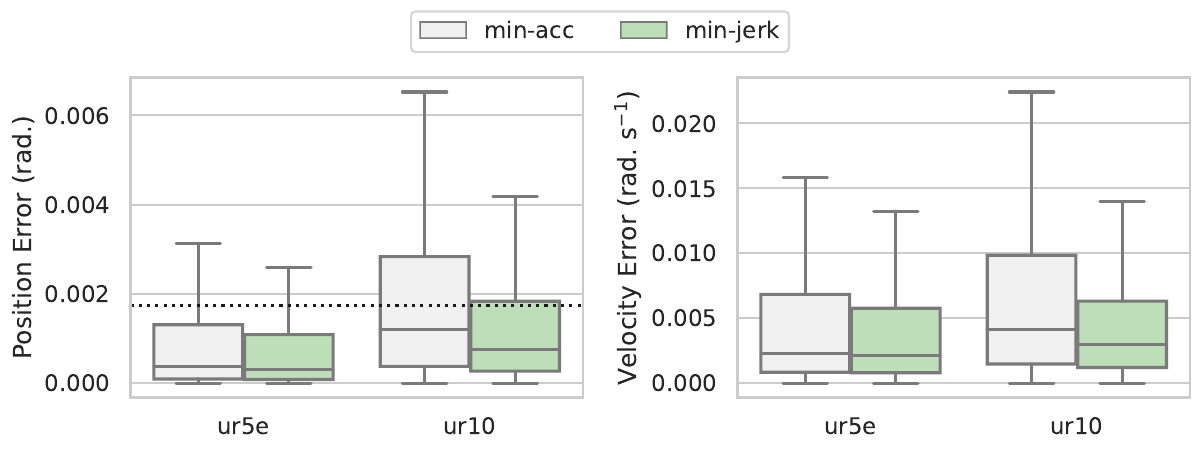}
    \caption{The tracking error in joint position space is plotted in the left and the error in velocity tracking is plotted on the right across the two real robots, UR5e and UR10. Here~\emph{min-acc} and \emph{min-jerk} refers to \emph{cuRobo}'s trajectory optimization with minimum jerk cost disabled and enabled respectively. The black dotted line in the left plot indicates the accuracy margin of the joint encoders on the robots.}
    \label{fig:tracking-error}
\end{figure}

\begin{figure}
    \centering
    \begin{tabular}{c c}
        \includegraphics[width=0.48\textwidth, trim=0 0 0 0cm, clip]{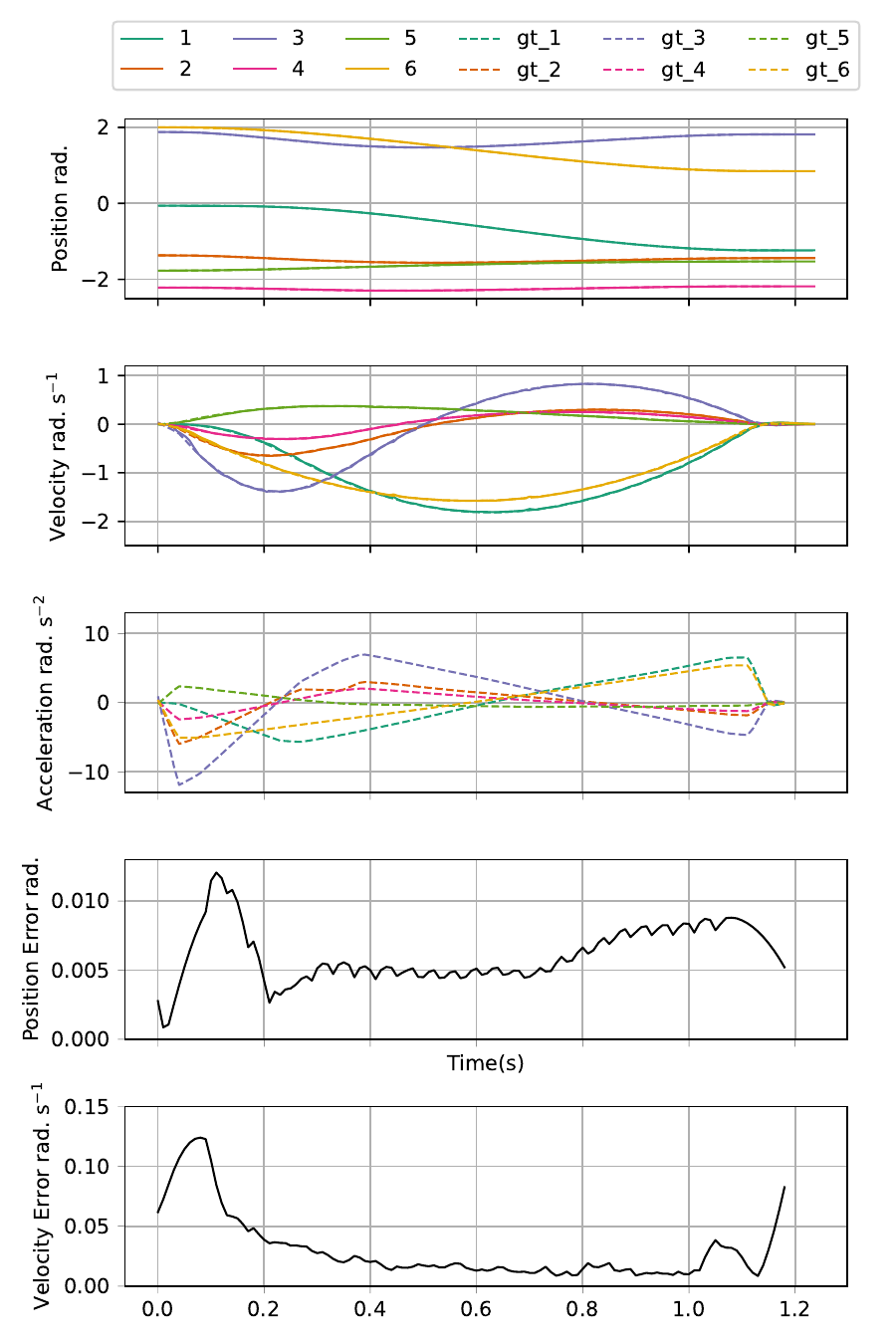}
        &
        \includegraphics[width=0.48\textwidth,trim=0 0 0 0cm, clip]{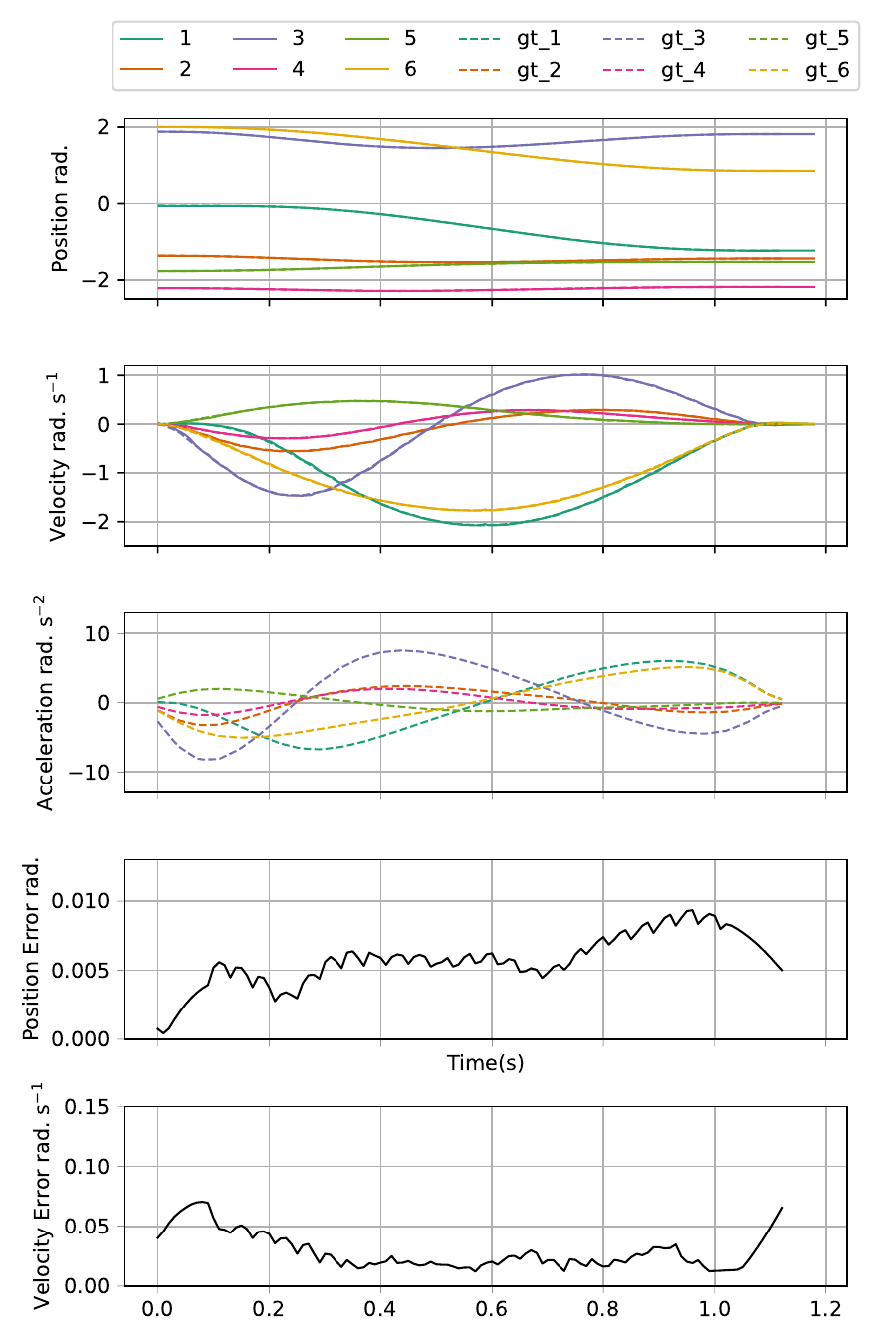}\\
        (a) Minimum Acceleration & (b)Minimum Jerk
    \end{tabular}
    \caption{The planned trajectory and the followed path is shown for the UR10 robot in position space and velocity space in the top two grid locations. We plot the planned trajectory's joint acceleration in the middle plot. We plot the sum of the absolute error in tracking position and velocity across all joints in the bottom two grid locations. We see the effect of large jerk at near the start time-steps of the velocity error and position error in (a) while (b) has a more flat error.}
    \label{fig:robot-execution}
\end{figure}

We first measure the error in tracking the position and velocity across the trajectories. Deviations from the planned path can lead to critical failure as the robot could hit obstacles in the world. Poor velocity tracking can lead to the robot taking more motion time than planned, creating uncertainty in cycle time for tasks. As reported in our results in Figure~\ref{fig:tracking-error}, \emph{min-jerk} has lower tracking errors in both position and velocity across both the robots. We observed a mean position error of 0.00117 radians and 0.00085 radians for \emph{min-acc} and \emph{min-jerk} on the UR5e robot, both within the 0.0017 radians accuracy margin of the joint encoders. On the UR10 robot, we found the mean position error to be 0.00250 radians and 0.00133 radians for \emph{min-acc} and \emph{min-jerk} respectively. We suspect the the larger position error on the UR10 to be because of the robot being physically larger, thereby requiring more dynamics compensation at high speeds compared to the UR5e. The position error for \emph{min-acc} is also larger than the accuracy margin of the encoder. 

To closely examine the difference in position error between \emph{min-acc} and \emph{min-jerk} on the UR10, we plot one executed trajectory from the trials in Figure~\ref{fig:robot-execution}. We observe that the robot with \emph{min-acc}, the robot has a large spike in position error at the start of the trajectory while in \emph{min-jerk} there is no steep increase in error at the start. We suspect this spike in \emph{min-acc} at the start to be because of the robot not being able to instantly accelerate to the maximum acceleration limit. With \emph{min-jerk}, we gradually increase the acceleration, thereby minimizing tracking error due to delay in robot's acceleration. While one could feed a feed forward torque to help the robot accelerate more quickly, sending torque commands is not possible in many industrial robots including the UR10 and UR5e.

\begin{figure}
    \centering
    \includegraphics[width=0.99\textwidth]{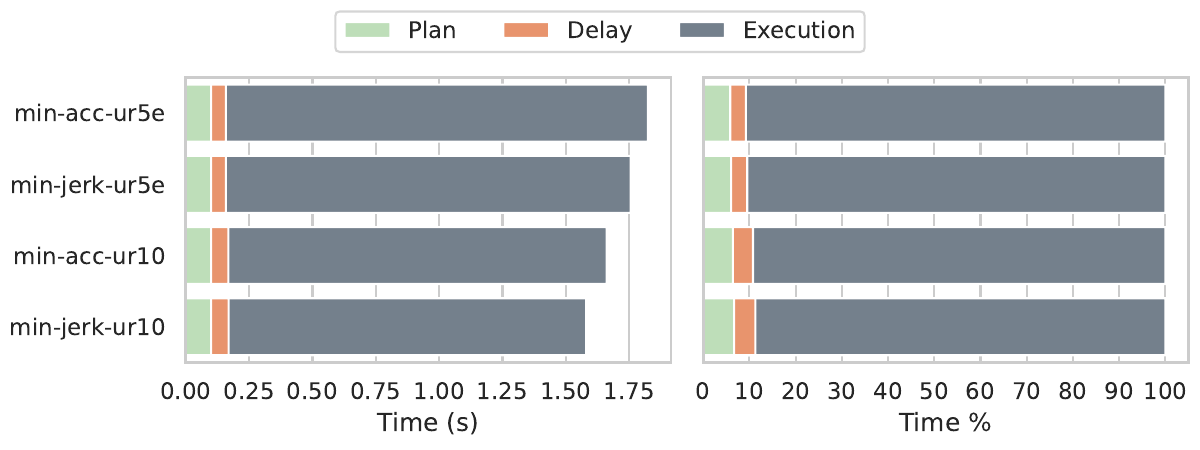}
    \caption{The time taken to plan and execute a trajectory across the two robots is plotted on the left. On the right, we plot the time split between planning~(\emph{Plan}), delay between sending a trajectory and the robot starting to move~(\emph{Delay}), followed by the time the robot is in motion~(\emph{Execution}).}
    \label{fig:real-robot-time}
\end{figure}

Our motion times for \emph{min-jerk} were [0.98, 1.57, 2.45] seconds and [0.87, 1.34, 2.16] seconds on the UR5e and UR10 respectively, where the numbers map to mean, \ps~percentile and \pn~percentile. Our motion times for \emph{min-acc} were [1.02, 1.59, 2.83] and [0.92, 1.52, 2.49] on the UR5e and UR10 respectively. 

As part of our real robot experiments, we also timed the whole pipeline, starting with when \emph{cuRobo} gets a planning query and completes computing a plan, followed by when the robot starts moving, and finally when the robot completes trajectory execution. We found \emph{cuRobo} to complete planning within 100ms on average for both UR5e and UR10 robots. We observed on average a delay of 58ms and 68ms between when a trajectory is sent to the UR ROS driver and when the robot starts moving on the UR5e and UR10 respectively. We plot the time it takes overall reach a target pose and the split between planning, delay, and execution in Figure~\ref{fig:real-robot-time}. We see that \emph{cuRobo} takes [6.02\%,  6.62\%, 9.12\%] and [6.67\%, 7.93\%, 9.13\%] of the time in the full pipeline on the UR5e and UR10 respectively in \emph{min-jerk} mode. The delay accounts for 5\% and 6\% of the time on \pn~percentile on UR5e and UR10 respectively. The robot is in motion [90\%, 92\%, 94\%] and [89\%, 90\%, 92\%] on UR5e and UR10 respectively. A common technique to reduce planning overhead in cycle time is to plan for the next sequence of targets while the robot is executing it's current trajectory. This has been leveraged with existing planners as they can take significantly longer planning times, in the range of 2.5 seconds. However, this can prevent the robot from reacting to any world or task changes between motions. With \emph{cuRobo} we can plan the next motion after executing the current trajectory as we only take 6\% of the cycle time on average. This also simplifies the robot programming pipeline, as computational tasks can be executed in serial.

\subsection{Deployment on different Robot Platforms}
\label{sec:other-robots}
\begin{figure}
    \centering
    \includegraphics[width=\textwidth]{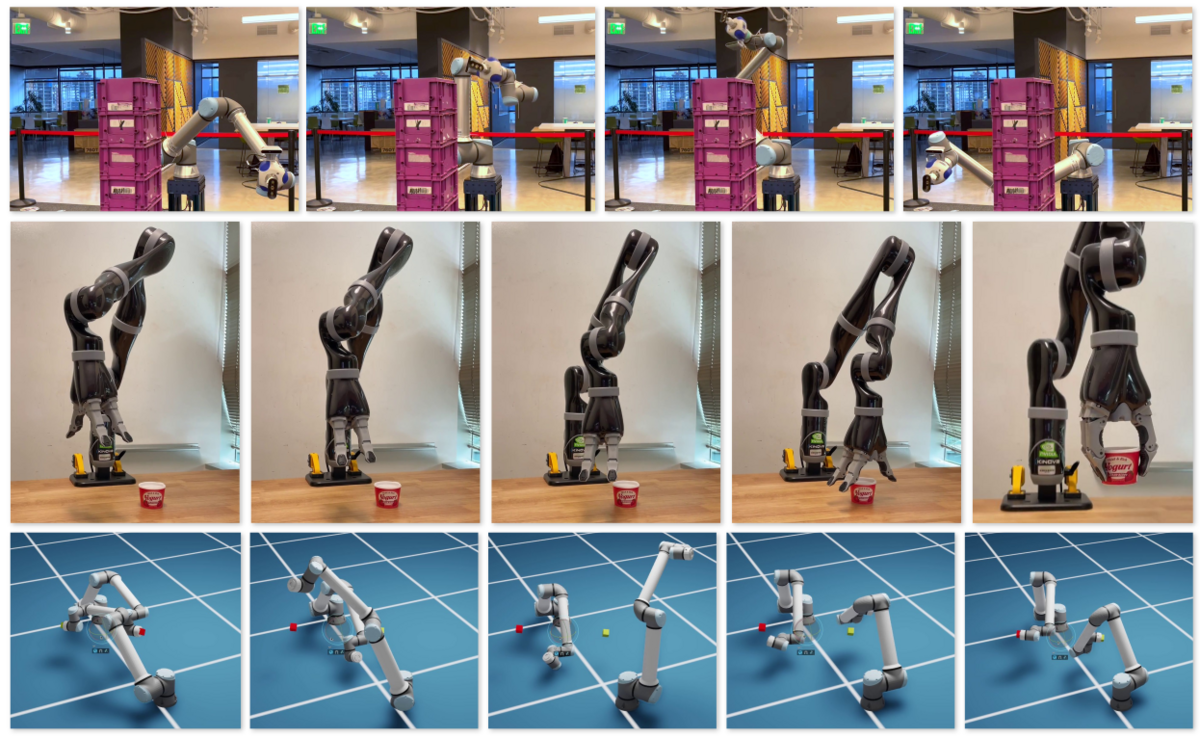}
    \caption{In the top row, we show the UR10e robot avoiding an obstacle by using cuRobo for motion generation in combination with an ESDF map that was built with {\it nvblox}. The second row shows a Kinova Jaco robot grasping an object by using \emph{cuRobo} to generate motions for moving to the grasp pose and lifting the object. The bottom row shows coordinated motion generation with a dual UR10e robot setup in NVIDIA Isaac Sim. The two UR10e robots start at a collision-free configuration and move around each other to reach their respective target poses given by red and yellow colored cubes.}
    \label{fig:robot_demos}
\end{figure}

We deployed \emph{cuRobo} on few different robot platforms as shown in Figure~\ref{fig:robot_demos}, with no changes to parameters in trajectory optimization. We created the robot spheres for these robots along with a collision-free rest configuration. We then called the inverse kinematics, geometric planning, and trajectory optimization methods. 

First, we deployed~\emph{cuRobo} on a UR10e with nvblox~\cite{nvblox, oleynikova2017voxblox, millane2023nvblox} to perform collision checking between the world and the robot as shown in Figure~\ref{fig:robot_demos}. We use nvblox to generate a euclidean signed distance field~(ESDF) map of the world, by scanning with a realsense D-415 camera attached to the end-effector. We then generate motions for the robot to go around obstacles. Next, we deployed on a Kinova Jaco arm as shown in Figure~\ref{fig:robot_demos}, where we implemented a PD controller in the velocity space to command the generated trajectory. We also tested coordinated motion generation for a dual arm UR10e robot setup in NVIDIA Isaac Sim and preliminary results are promising as \emph{cuRobo} finds collision-free paths to move both arms to their targets as shown in Figure~\ref{fig:robot_demos}.

\subsection{Summary}
We first compared the quality of motions generated from \emph{cuRobo} in Section~\ref{sec:motion-gen-benchmark} and showed that \emph{cuRobo} generates better solutions than existing techniques. We then showed in Section~\ref{sec:mb-time} that \emph{cuRobo} generates these high quality solutions in a fraction of the time taken by existing methods across different computing platforms including a 15W NVIDIA Jetson device. We then compared the compute time across sub-components, inverse kinematics, geometric planning, and trajectory optimization and showed double digit speedups compared to existing implementations. We validated our motion generation approach on two robots in Section~\ref{sec:real-robot-tracking}, a UR5e and UR10 robot, both tracking the minimum jerk high-speed trajectories from \emph{cuRobo} with position errors below the accuracy margin of the joint encoders. We also showed our approach working on different robot platforms in Section~\ref{sec:other-robots}.

\section{Component Analysis}
\label{sec:optimization-analysis}
We study the effect of different components in \emph{cuRobo}'s trajectory optimization, starting with collision cost formulation, followed by the effect of number of parallel seeds in trajectory optimization, and then the effect of different parameters in our numerical optimization solvers.

\subsection{Collision Cost Formulation}
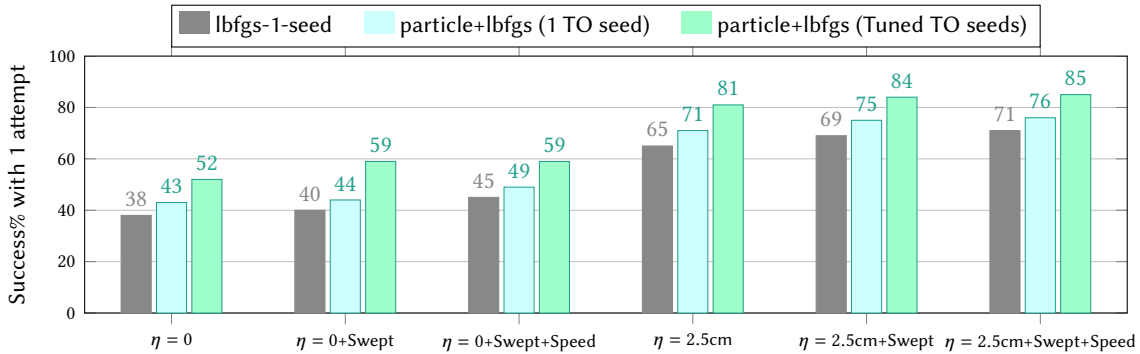
\begin{figure}[b]
    \centering
     \begin{tikzpicture}  
     \tikzstyle{every node}=[font=\footnotesize]

    \begin{axis}  
    [  
        ybar,  
        bar width=0.4cm,
        grid style={line width=.1pt, draw=gray!10},major grid style={line width=.2pt,draw=gray!50},
        ymajorgrids,
        ylabel={Success\% with 1 attempt},
        symbolic x coords={$\eta=0$,$\eta=0$+Swept,  
        $\eta=0$+Swept+Speed,
        $\eta=2.5$cm, $\eta=2.5$cm+Swept, $\eta=2.5$cm+Swept+Speed}, %
        xtick=data,  
         nodes near coords, %
        nodes near coords align={vertical},  
        ymin=0,
        ymax=100,
        height=5cm, 
        width=\textwidth,
        compat=1.5,
        ylabel style={align=center},
        legend columns=3,
        legend style={at={(0.5,1.2)},anchor=north,
        /tikz/column 2/.style={
            column sep=10pt,
        },
        /tikz/column 4/.style={
            column sep=10pt,
        }
        },
        legend image code/.code={%
                    \draw[#1, draw=none] (0cm,-0.1cm) rectangle (0.4cm,0.2cm);
        },
        x=2.31cm,
    ]
    \addplot[barorangeline, fill=barorange] coordinates {($\eta=0$, 38)
    ($\eta=0$+Swept,40) 
    ($\eta=0$+Swept+Speed,45) 
    ($\eta=2.5$cm, 65) 
    ($\eta=2.5$cm+Swept,69) 
    ($\eta=2.5$cm+Swept+Speed, 71)}; %
    
    \addplot[bargreenline, fill=bargreenlight] coordinates {
    ($\eta=0$, 43)
    ($\eta=0$+Swept,44) 
    ($\eta=0$+Swept+Speed,49) 
    ($\eta=2.5$cm, 71) ($\eta=2.5$cm+Swept, 75) ($\eta=2.5$cm+Swept+Speed, 76)}; %
    
    \addplot[bargreenline, fill=bargreen] coordinates {
    ($\eta=0$, 52)
    ($\eta=0$+Swept,59)
    ($\eta=0$+Swept+Speed,59) 
    ($\eta=2.5$cm, 81) ($\eta=2.5$cm+Swept, 84) ($\eta=2.5$cm+Swept+Speed, 85)}; %
    \legend{lbfgs-1-seed, particle+lbfgs (1 TO seed), particle+lbfgs (Tuned  TO seeds)} 
    \end{axis}  
    \end{tikzpicture}  
     
    \caption{We compare the effect of different metrics in computing world collision cost. We start with a traditional formulation of collision cost where we only add a cost when robot is in collision with the world~$\eta=0$, followed by addition of our continuous collision checking implementation~$Swept$, and then the speed metric introduced in~\cite{ratliff2009chomp}. We then add an activation distance of 2.5 cm to these metrics.}
    \label{fig:coll-speed-metric}
\end{figure}

We first analyze the impact of different collision cost formulations on success of the optimization problem in Figure~\ref{fig:coll-speed-metric}. We ran our motion generation pipeline without geometric planning and 500 IK seeds. We ran experiments without particle-based optimization, with particle-based optimization, and with 1 and many seeds. From the results of this experiment, we make the following observations:
\begin{itemize}
    \item Increasing activation distance from 0cm to 2.5cm improves success rate by 27\%, having the largest impact in success rate. 
    \item Using continuous collision checking (swept) improves success rate further by 4\% when compared to only using an activation distance of 2.5cm. 
    \item Speed metric improves success rate further by 2\% across the dataset.
\end{itemize}

We found activation distance is critical in improving success rate as we perform trajectory optimization at a coarse scale of a few timesteps(32-50) and then interpolate the trajectory to a fixed dt of 0.025 to validate success. After this interpolation, a collision-free trajectory can move into regions of collision and lead to failure. In addition, having an activation distance adds smoothness to the cost term, making it easier for an optimization solver to minimize collisions. Our continuous collision checker checks collisions between timesteps by linearly interpolating in the task space, approximating linear interpolation in the joint space. This improves the success rate by 7\% with tuned TO seeds and zero activation distance. The improvement diminishes with activation distance where it only improves by 3\%.

\begin{figure}
    \centering
    \includegraphics[width=0.99\textwidth, trim={0 0 0.5cm 0}, clip]{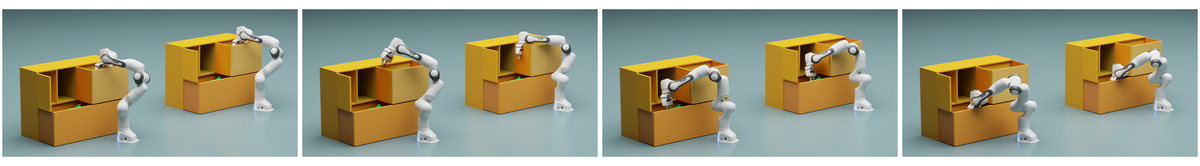}
    \caption{Effect of Speed Metric during Trajectory Optimization is visualized in the dresser environment. The left robot in each frame uses the speed metric during optimization, enabling the solver to move out of collision (in the middle image). The right robot in each frame does not use the speed metric and as a result speeds through the high cost collision region.}
    \label{fig:speed-metric-opt}
\end{figure}

The effect of speed metric is observable at an activation distance of 0.0 cm on trajectory optimization with 1 seed, where it provides a 5\% improvement over LBFGS and also over particle+LBFGS. This effect diminishes when we use many parallel seeds as a collision-free path is obtained through other seeds. We visualize the effect of speed metric in Figure~\ref{fig:speed-metric-opt}, where the use of speed metric enables the optimization solver to find a path that is collision-free while without the metric, the solver ends up speeding through the high cost region.

\begin{figure}
    \centering
    \begin{tikzpicture}  
         \tikzstyle{every node}=[font=\footnotesize]
        \begin{axis}
        [  
            ybar,  
            grid style={line width=.1pt, draw=gray!10},major grid style={line width=.2pt,draw=gray!50},
            ymajorgrids,
            bar width=2ex,
            ylabel style={align=center},
            ylabel=Success \%\\(1 attempt),
            xlabel={Number of TO seeds},  
            symbolic x coords={1, 4, 12, 24, 48, 128, GP1, GP4,
            GP12}, 
            xtick=data,  
             nodes near coords,
            nodes near coords align={vertical},  
            ymin=0, 
            ymax=115,
            height=4cm,
            width=0.98\textwidth,
            legend columns=2,
            legend style={at={(0.5,1.3)},anchor=north,
            /tikz/column 2/.style={
                column sep=10pt,
            },},
            legend image code/.code={%
                    \draw[#1, draw=none] (0cm,-0.1cm) rectangle (0.4cm,0.2cm);
                },  
            ]  

        \addplot[barorangeline, fill=barorange] coordinates {(1, 73) (4, 79) (12, 80) (24, 82) (48, 85) (128, 86) (GP1, 90) (GP4, 94) (GP12, 96)
        };  
         \addplot[bargreenline, fill=bargreen] coordinates {(1, 77) (4, 83) (12, 84) (24, 86) (48, 87) (128, 88) (GP1, 91) (GP4, 94) (GP12, 96)
        };        
        \legend{lbfgs, particle+lbfgs};
        \end{axis}  
        \end{tikzpicture}
    \begin{tabular}{c c}
    \begin{tikzpicture}
         \tikzstyle{every node}=[font=\footnotesize]
        \begin{axis}
        [  
            ybar,  
            grid style={line width=.1pt, draw=gray!10},major grid style={line width=.2pt,draw=gray!50},
            ymajorgrids,
            bar width=0.25cm,
            ylabel={Planning Time (ms)},
            xlabel={Number of TO seeds},  
            symbolic x coords={1, 4, 12, 24, 48, 128, GP1, GP4, GP12},   
            xtick=data,  
             nodes near coords,  
            nodes near coords align={vertical},  
            ymin=0, 
            ymax=140,
            height=4cm,
            width=0.48\textwidth,
            ]  

         \addplot[bargreenline, fill=bargreen] coordinates {(1, 25) (4, 26) (12, 27) (24, 30) (48, 36) (128, 66) (GP1, 60) (GP4, 74) (GP12, 116)
        };        
        \end{axis}  
        \end{tikzpicture}
             &  
    \begin{tikzpicture}  
         \tikzstyle{every node}=[font=\footnotesize]
        \begin{axis}  
        [  
            ybar,  
            grid style={line width=.1pt, draw=gray!10},major grid style={line width=.2pt,draw=gray!50},
            ymajorgrids,
            bar width=0.25cm,
            ylabel={\ps Motion Time (s)}, 
            xlabel={Number of TO seeds}, 
            symbolic x coords={1, 4, 12, 24, 48, 128, GP1, GP4,
            GP12},
            xtick=data,  
             nodes near coords,
            nodes near coords align={vertical},  
            ymin=0, 
            ymax=2.5,
            height=4cm,
            width=0.48\textwidth,
            ]  
        \addplot[bargreenline, fill=bargreen] coordinates {(1, 1.69) (4, 1.77) (12, 1.77) (24, 1.77) (48, 1.59) (128, 1.58) (GP1, 1.86) (GP4, 1.94) (GP12, 2.02)
        };  
        \end{axis}  
        \end{tikzpicture} 
    \end{tabular}
    
    \caption{The improvement in success rate across the 2600 problems in our dataset as we increase the number of trajectory optimization seeds is shown in the top plot. We also see that the use of geometric planning to seed trajectory optimization~(GP1, GP4, GP12) increases the success rate to 96\% with an increase in planning time. We plot the \ps~percentile \emph{Motion Time} on the bottom right plot across the number of seeds.}
    \label{fig:to-seeds-success}
   
\end{figure}
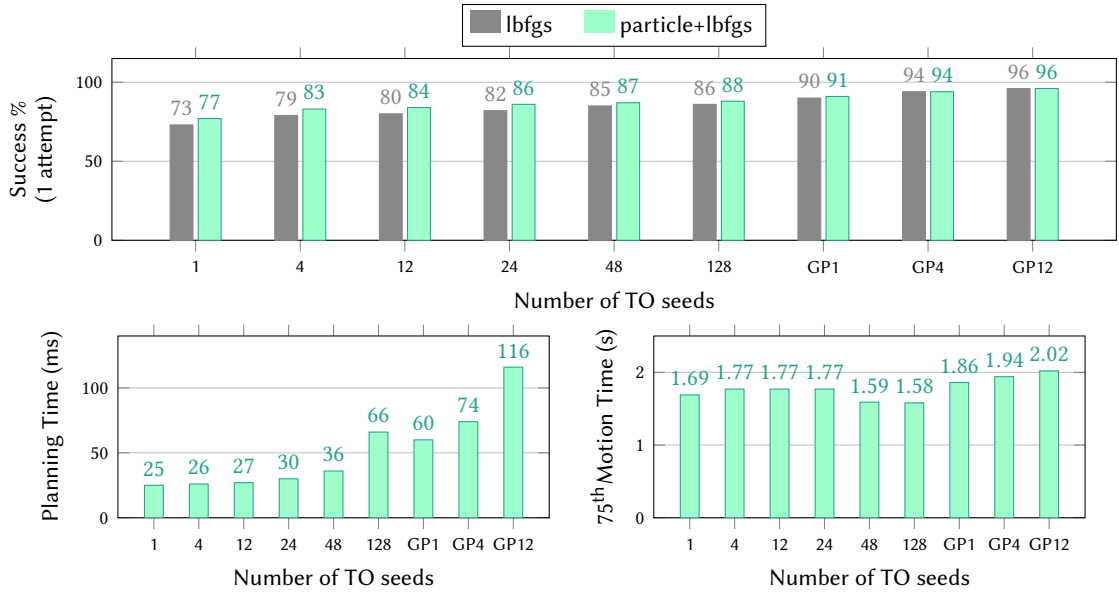

From the results in Figure~\ref{fig:coll-speed-metric}, we see that a naive implementation of collision cost with a gradient based optimizer and 1 seed achieves a success rate of 38\%. Adding an activation distance, as introduced by~\cite{ratliff2009chomp} would bring this to 65\%, followed by addition of a continuous collision checking would bring this to 69\%. Addition of the speed metric from~\cite{ratliff2009chomp} would increase this further to 71\%. Adding a particle-based optimizer to initialize gradient-based solver would increase the success to 76\%. Finally, running trajectory optimization across many seeds would increase the success rate to 85\%. 

Overall, our application of existing techniques from~\cite{ratliff2009chomp} in combination with our continuous collision checking module and many parallel seeds for trajectory optimization improves the success rate from 38\% to 85\%, enabling \emph{cuRobo} to solve 85\% of the motion generation problems within 1 attempt.

\subsection{Effect of Parallel Seeds}

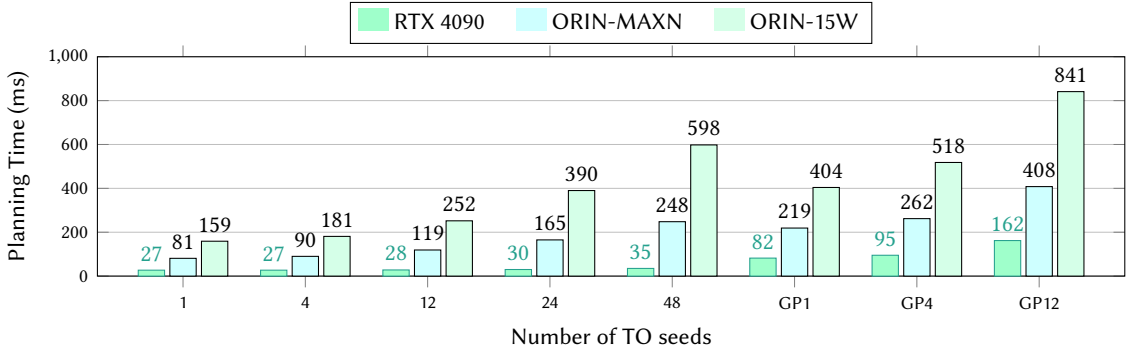
\begin{figure}
    \centering
    \begin{tikzpicture}  
         \tikzstyle{every node}=[font=\footnotesize]

        \begin{axis}
        [  
            ybar,  
            grid style={line width=.1pt, draw=gray!10},major grid style={line width=.2pt,draw=gray!50},
            ymajorgrids,
            bar width=0.35cm,
            ylabel={Planning Time (ms)}, %
            xlabel={Number of TO seeds},  
            symbolic x coords={1, 4, 12, 24, 48, GP1, GP4, GP12},   
            xtick=data,  
            nodes near coords, %
            nodes near coords align={vertical},  
            ymin=0, 
            ymax=1000,
            height=4.5cm,
            width=\textwidth,
            legend columns=3,
            legend style={at={(0.5,1.25)},anchor=north,
            /tikz/column 2/.style={
                column sep=10pt,
            },
            /tikz/column 4/.style={
                column sep=10pt,
            },},
            legend image code/.code={%
                    \draw[#1, draw=none] (0cm,-0.1cm) rectangle (0.4cm,0.2cm);
                },  
            ]  

         \addplot[bargreenline, fill=bargreen] coordinates {(1, 27) (4, 27) (12, 28) (24, 30) (48, 35) (GP1,82) (GP4,95) (GP12,162)
        };        
         \addplot[black, fill=bargreenlight] coordinates {(1, 81) (4, 90) (12, 119) (24, 165) (48, 248)
         (GP1, 219) (GP4,262) (GP12,408)
        };        
         \addplot[black, fill=bargreenlight2] coordinates {
         (1, 159) (4, 181) (12, 252) (24, 390) (48, 598) (GP1, 404) (GP4, 518) (GP12, 841)
        };        
        \legend{RTX 4090, ORIN-MAXN, ORIN-15W}
        \end{axis}  
        \end{tikzpicture}
    \caption{The slowdown in planning time as we increase number of seeds used in trajectory optimization is visualized across the 100 problems on \emph{cage-panda} environment. The slowdown is more significant on the NVIDIA Jetson Orin as it has lower number of SMs compared to a RTX 4090.}
    \label{fig:to-seeds-time}
\end{figure}

In an ideal setting, we would want to run as many parallel seeds of optimization as possible and pick the best solution. However, as we reduce the compute available for motion generation, the number of parallel seeds used can have a significant impact on compute time. We study the interaction between number of seeds and compute time we ran motion generation on the 100 problems from the \emph{cage-panda} environment with varying number of trajectory optimization seeds across the three compute platforms. We used 500 IK optimization seeds and also run particle based optimization to initialize L-BFGS in these experiments. We observed that on a RTX 4090, the compute time only changes by 8ms from using 4 seeds to 48 seeds while it changes by 158ms and 417ms on the ORIN at MAXN and 15W respectively. We plot these results in Figure~\ref{fig:to-seeds-time}.

Next, we study the impact of trajectory optimization seeds on success and motion time in Figure~\ref{fig:to-seeds-success}. We observed that increasing the number of parallel seeds for trajectory optimization increases the success rate and also decreases the motion time starting at 48 seeds. Initializing trajectory optimization with collision-free paths from our geometric planner also increases the success rate. However, using geometric planner to initialize trajectory optimization doubles the planning time on most problems. In addition, we found that the motion time was higher when trajectory optimization is initialized with a geometric planner. Based on these results, we use 12 trajectory optimization seeds for most environments in our evaluation dataset and increase this up to 28 for harder environments. We discuss the parameters used in Appendix~\ref{app:curobo-parameters}.

\subsection{Line Search}
The role of line search in gradient-based numerical optimization solvers is to find the best magnitude to scale the step direction. A common strategy to find the best magnitude is by backtracking search, where we start with a magnitude of 1 and reduce this value until some conditions are met. Two common conditions used in many modern numerical solver libraries are \emph{weak-wolfe} and \emph{strong-wolfe} which we term~\emph{wolfe} and \emph{st-wolfe} in our comparisons. We also compare against no scaling of the step direction which we term~\emph{no-ls}. We compare these options with our noisy line search technique introduced in Section~\ref{sec:opt_solver}. We use [0.01,0.3,0.7,1.0] as the values for the noisy line search which we term~\emph{noisy-ls}. We also compare with only 1 value for noisy line search~[0.01,1.0] which we term~\emph{noisy-ls-1}. For wolfe and strong wolfe, we use the values~[0.0001,0.001,0.1,0.2,0.3,0.4,0.5,0.6,0.7,0.8,0.9,1.0] in the line search. We evaluate these line search techniques across the full dataset with 1 trajectory optimization seed initialized by 2 iterations of particle-based optimization. 

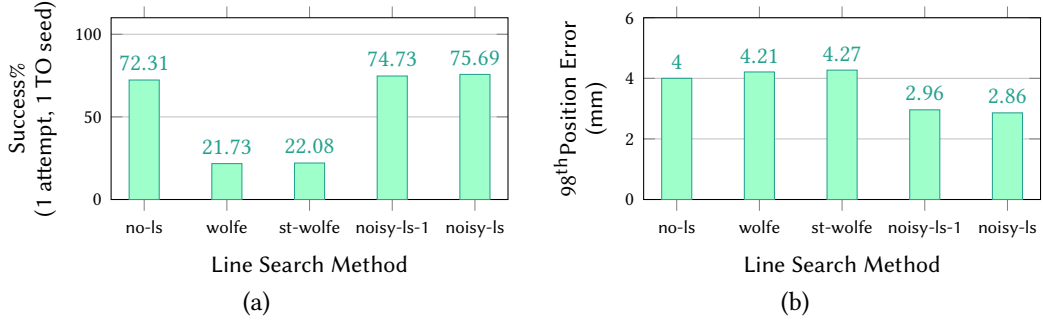
\begin{figure}
\centering
\begin{tabular}{c c}
    \begin{tikzpicture}  
    \tikzstyle{every node}=[font=\footnotesize]
    \begin{axis}  
    [   ybar,  
        grid style={line width=.1pt, draw=gray!10},major grid style={line width=.2pt,draw=gray!50},
        ymajorgrids,
        bar width=0.4cm,
        height=4cm,
        width=0.45\textwidth,
        ylabel={Success\% \\(1 attempt, 1 TO seed)}, %
        xlabel={Line Search Method},  
        symbolic x coords={no-ls, wolfe, st-wolfe, noisy-ls-1,
        noisy-ls}, %
        xtick=data,  
         nodes near coords, %
        nodes near coords align={vertical},  
        ymin=0,
        ymax=110,
        ylabel style={align=center},
        ]  
    \addplot[bargreenline, fill=bargreen] coordinates {(no-ls, 72.31) (wolfe, 21.73) (st-wolfe, 22.08) (noisy-ls-1, 74.73) (noisy-ls, 75.69)};

    \end{axis}  
    \end{tikzpicture}
    &
\begin{tikzpicture}  
    \tikzstyle{every node}=[font=\footnotesize]
    \begin{axis}  
    [   ybar,  
        grid style={line width=.1pt, draw=gray!10},major grid style={line width=.2pt,draw=gray!50},
        ymajorgrids,
        bar width=0.4cm,
        height=4cm,
        width=0.45\textwidth,
        ylabel={\pn  Position Error\\ (mm)}, %
        xlabel={Line Search Method},  
        symbolic x coords={no-ls, wolfe, st-wolfe, noisy-ls-1,
        noisy-ls}, %
        xtick=data,  
         nodes near coords, %
        nodes near coords align={vertical},  
        ymin=0,
        ymax=6,
        ylabel style={align=center},
        ]  
    \addplot[bargreenline, fill=bargreen] coordinates {(no-ls, 4) (wolfe, 4.21) (st-wolfe, 4.27) (noisy-ls-1, 2.96) (noisy-ls, 2.86)};  
    \end{axis}  
    \end{tikzpicture}
   \\
(a) & (b)
\end{tabular}
\caption{We compare the different line search techniques and their impact on trajectory optimization, starting with no line search~\emph{no-ls}, followed by common line search methods~weak wolfe~\emph{wolfe} and strong wolfe~\emph{st-wolfe}. We compare these methods to our proposed noisy line search scheme~\emph{noisy-ls}. We also compare to noisy line search with only two values~[0.01,1] which we term~\emph{noisy-ls-1}.}
\label{fig:line-search-plot}
\end{figure}
From Figure~\ref{fig:line-search-plot}, we see that \emph{no-ls}, \emph{noisy-ls-1}, and \emph{noisy-ls} all have a success rate higher than 70\%, with \emph{noisy-ls} having the highest success rate of 75.69\%. While most of these techniques lead to good success rate, the impact of line search is more observable in the position error at the final timestep of the trajectory across the dataset as plotted in Figure~\ref{fig:line-search-plot}-(b). We found noisy line search to have the lowest error of 2.86 mm while no line search ends up with an error of 4 mm. We observed wolfe and strong wolfe perform worse in both success and position error across the dataset as for these line search techniques, we use a magnitude scale of 0.0 if none of the chosen line search values satisfy the conditions. 

\subsection{Gradient Descent and Effect of history length in L-BFGS}

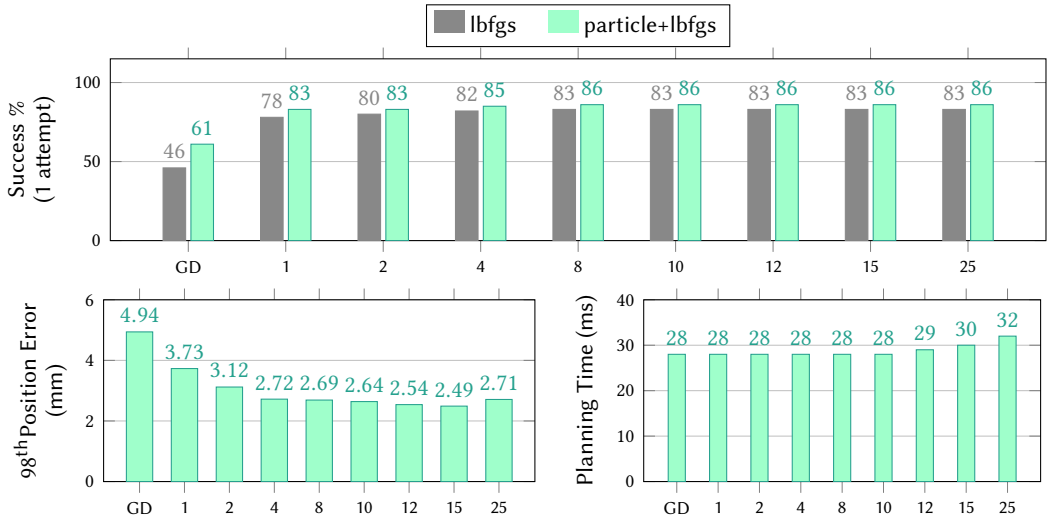
\begin{figure}
    \centering
     \begin{tikzpicture}  
         \tikzstyle{every node}=[font=\footnotesize]

        \begin{axis}
        [  
            ybar,  
            grid style={line width=.1pt, draw=gray!10},major grid style={line width=.2pt,draw=gray!50},
            ymajorgrids,
            bar width=0.3cm,
            ylabel style={align=center},
            ylabel= Success \%\\(1 attempt),
            symbolic x coords={GD, 1, 2, 4, 8,10, 12, 15, 25},
            xtick=data,  
             nodes near coords,
            nodes near coords align={vertical},  
            ymin=0, 
            ymax=115,
            height=4cm,
            width=0.92\textwidth,
            legend columns=2,
            legend style={at={(0.5,1.3)},anchor=north,
            /tikz/column 2/.style={
                column sep=10pt,
            },
            },
            legend image code/.code={%
                    \draw[#1, draw=none] (0cm,-0.1cm) rectangle (0.4cm,0.2cm);
            }, 
            ]  
        \addplot[barorangeline, fill=barorange] coordinates {
        (GD, 46)
        (1, 78) (2,80) (4,82)
        (8,83) (10, 83) (12,83) (15,83) (25,83)
        };  
        \addplot[bargreenline, fill=bargreen] coordinates {
        (GD, 61)
        (1, 83) (2,83) (4,85)
        (8,86) (10, 86) (12,86) (15,86) (25,86)
        };  
        \legend{lbfgs, particle+lbfgs};
    \end{axis}  
    \end{tikzpicture} 
    \begin{tabular}{c c}
    \begin{tikzpicture}  
         \tikzstyle{every node}=[font=\footnotesize]

        \begin{axis}
        [  
            ybar,  
            grid style={line width=.1pt, draw=gray!10},major grid style={line width=.2pt,draw=gray!50},
            ymajorgrids,
            bar width=0.35cm,
            ylabel style={align=center},
            ylabel=\pn Position Error\\(mm),
            symbolic x coords={GD, 1, 2, 4, 8,10, 12, 15, 25},
            xtick=data,  
         nodes near coords,   
            nodes near coords align={vertical},  
            ymin=0, 
            ymax=6,
            height=4cm,
            width=0.48\textwidth,
            ]  

        \addplot[bargreenline, fill=bargreen] coordinates {
        (GD, 4.94)
        (1, 3.73) (2,3.12) (4,2.72)
        (8,2.69) (10, 2.64) (12,2.54) (15,2.49) (25,2.71)
        };  
        \end{axis}  
        \end{tikzpicture} 
        &
    \begin{tikzpicture}  
         \tikzstyle{every node}=[font=\footnotesize]
        \begin{axis}
        [  
            ybar,  
            grid style={line width=.1pt, draw=gray!10},major grid style={line width=.2pt,draw=gray!50},
            ymajorgrids,
            bar width=1.5ex,
            ylabel={Planning Time (ms)},
            symbolic x coords={GD, 1, 2, 4, 8,10, 12, 15, 25},
            xtick=data,  
             nodes near coords, %
            nodes near coords align={vertical},  
            ymin=0, 
            ymax=40,
            height=4cm,
            width=0.45\textwidth,
            ]  
        
        \addplot[bargreenline, fill=bargreen] coordinates {
        (GD, 28)
        (1, 28) (2,28) (4,28)
        (8,28) (10, 28) (12,29) (15,30) (25,32)
        };  
        \end{axis}  
        \end{tikzpicture} 
    \\
    \end{tabular}
    \caption{We compare Gradient Descent~(GD) to L-BFGS with different history lengths, starting with the success rate in the top plot. We also plot the \pn~percentile position error in bottom left and the average planning time on the bottom right.}
    \label{fig:lbfgs-history}
  
\end{figure}

\begin{figure}
    \centering
    \begin{tabular}{c c}
        \includegraphics[width=0.46\textwidth]{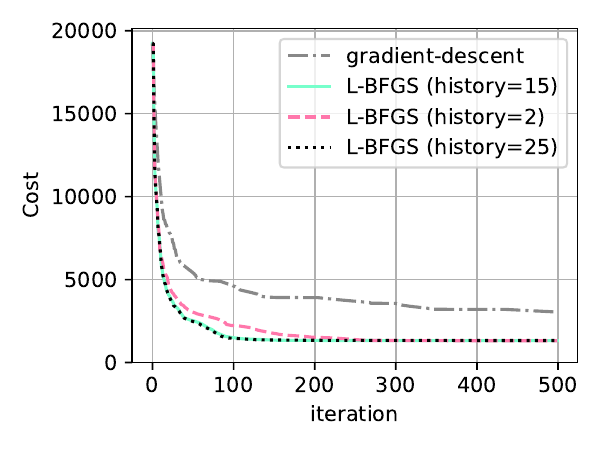} & 
        \includegraphics[width=0.46\textwidth]{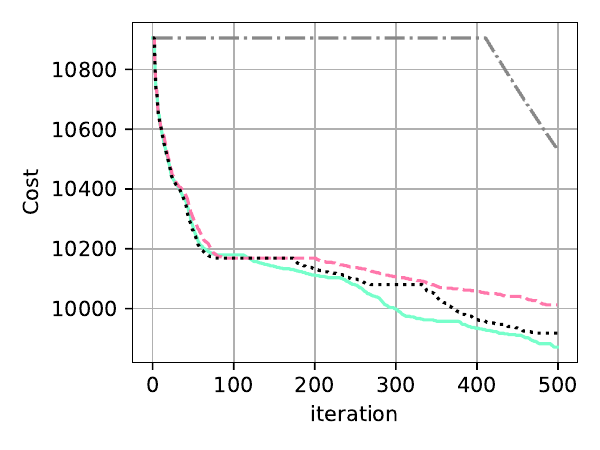} \\
        (a) Trajectory Optimization & (b) Timestep Optimization 
    \end{tabular}
    \caption{The cost at every iteration on one trajectory optimization problem from our evaluation set is plotted in (a) and the optimization after finding the optimal timestep is plotted in (b). We see that L-BFGS converges within 100 iterations at a history of 15 and 25, while a history of 2 leads to requiring 200 iterations to converge in (a). Gradient descent does not converge to a similar cost as L-BFGS even after 500 iterations. We also show the reduction in cost during timestep optimization in (b), where gradient descent doesn't start improving until 400 iterations.}
    \label{fig:finetuning}
\end{figure}

We ran trajectory optimization with gradient descent as the optimizer, swapping out L-BFGS, but keeping our noisy line search technique and found that it succeeded in 46\% of the problems. Using particle-based optimization to initialize gradient descent increased the success to 61\%. However, the \pn~percentile position error was 4.94cm compared to L-BFGS's 2.72cm. In addition, our efficient implementation of L-BFGS enables us to compute step direction using L-BFGS within similar compute times to using Gradient Descent upto a history length of 12. The gap in performance between gradient descent and L-BFGS is also visualized on one trajectory optimization problem in Figure~\ref{fig:finetuning} where we see that L-BFGS with any history length converges to the minimum within 100-200 iterations while gradient descent has not converged even after 500 iterations. 

We also ran experiments to study the impact of history length in L-BFGS. L-BFGS approximates the hessian by using the recent gradients and costs, which are stored in a history buffer. The more history L-BFGS uses to approximate the hessian, the closer the hessian gets to the true hessian. However, with increasing history, the compute required to compute the step direction also increases as shown in Figure~\ref{fig:lbfgs-history}-(b). To allow for the best chance between methods, we run our method with tuned number of parallel seeds across the dataset. We also compare the improvement we get with particle-based solver for initialization. From Figure~\ref{fig:lbfgs-history}, we found that increasing history indeed improves convergence as seen by the improvement in success rate and reduction in position error. And increasing history starts having an impact on planning time at a value of 12. We chose a history of 4 for all our evaluations as we observed that the improvement in success rate and position error was not significant.

\subsection{Effect of Particle Optimization}
Across analysis of many of the components, we compare the improvement we obtain with particle optimization. We observed that since we only run 2 iterations of particle-based optimization, it only adds 2 ms to the planning time on average while improving success. We observed that using two iterations of particle-based optimization improved the success by 5\% from 71\% to 76\% when running trajectory optimization with a single seed. With multiple seeds, the success improved by 3\%, from 82\% to 85\%.

\section{Concluding Remarks}
\label{sec:discussion} 

We formulated the global motion generation problem as an optimization problem and introduced techniques from global optimization literature to efficiently solve them with accelerated parallel compute. We empirically validated our approach on a difficult set of motion generation problems for manipulators~\cite{chamzas2021motionbenchmaker, fishman2022mpinets} and showed a 60$\times$ speedup over SOTA motion generation methods on a modern PC and a speedup of 28$\times$ on a NVIDIA Jetson AGX ORIN at 60W. We release our implementations as a high performance CUDA accelerated library~\emph{cuRobo} with pytorch and python wrappers for easy integration in robotic pipelines. We will discuss some limitations of our work and potential extensions in the next section. 

\subsection{Limitations \& Open Research Problems}
There are several open research problems in motion generation that our approach does not solve in it's current form. We hope that our results and framework can be leveraged to solve these problems. We list some key problems below, 
\\ \noindent \textbf{Global Reactive Motion Generation} Our approach is currently limited to planning full motions, where the robot starts from a static state. While we can extend our trajectory optimization to start from a non-static state, reactive motion generation requires motions at a fixed solve rate and our approach can take longer time on hard problems as shown by our longer \pn~percentile time.
\\ \noindent \textbf{Constrained Motion Generation} Our preliminary evaluation on pose constraints~\cite{berenson2011task} during motion such as maintaining orientation or height by adding a large weight to a running cost showed promise. However, we did not evaluate rigorously and also did not implement constrained graph planning which could be important for getting global guarantees. 
\\ \noindent \textbf{Task Sequencing and Route Planning} Finding the optimal sequence of poses to reach given an unordered list is an important problem in the industry, often called Task Sequencing \cite{ alatartsev2015robotic} and route planning~\cite{gentilini2013travelling}. We do not explore this task in this work but do see point to point motion generation to be a critical component of finding the optimal sequence. Task and motion planning~\cite{garrett2021integrated, garrett2020pddlstream, task_constructor_moveit} could also leverage our point to point motion generation for faster feasibility checks and finding more optimal sequence of motions.
\\ \noindent \textbf{Contact-Rich and Full-Dynamics Trajectory Optimization} Our current cost terms are focused on kinematic trajectory optimization with limits on position derivatives. Interacting with objects in the world would require optimizing over contacts, e.g. with contact implicit formulations~\cite{manchester2020variational} of trajectory optimization. Trajectories that minimize torques or other parts of dynamics would require optimizing over the robot's dynamics. Integrating existing GPU accelerated dynamics implementations~\cite{plancher2022grid,plancher2020performance,plancher2019realtime} into \emph{cuRobo} could potentially solve for trajectories over dynamics.  
\\ \noindent \textbf{Robotics Solvers} We implemented MPPI, gradient descent and L-BFGS in our framework to solve motion generation. Robotics focused numerical solvers are showing promise in quicker and better convergence compared to standard numerical solvers~\cite{Bambade-RSS-22,howell2022calipso} on robotics tasks. Formulating parallel compute friendly versions~\cite{schubiger2020gpu, pineda2022theseus} of these solvers could reduce the compute time even further and also enable new features in \emph{cuRobo} such as solving with hard constraints.
\\ \noindent \textbf{Collision Avoidance from partial world sensing} Our approach does not tackle partial perception and instead assumes that we can obtain a complete representation of the world. Incorporating learned representations of the world~\cite{murali2023cabinet,yamada2023leveraging,vandermerwe-icra2020-reconstruction-grasping} into our collision cost term could extend our motion generation method to work in unknown environments. 
\\ \noindent \textbf{Accelerating Computational Blocks} Our approach encapsulates the key components in motion generation into modular components and enables researchers to develop improved algorithms for these components without requiring full expertise on the stack. We hope that this broadens our library's audience to experts in non-robotics fields. We think that this is important as researchers in computer architecture are starting to accelerate manipulator algorithms~\cite{shahisca2023, plancher2022grid} and providing a reference SOTA implementation along with benchmarks can greatly reduce the entry barrier for researchers to accelerate robotics. As a step in this direction, \emph{cuRobo} has been used by computer architecture researchers to reduce memory bottlenecks in motion generation with reduced precision techniques~\cite{vapriros2023}.

\subsection{Acknowledgements}
We acknowledge several people who have helped in this work. Adithyavairavan Murali for helpful discussions on motion planning for grasping objects. Julen Urain for feedback on code design and early testing of collision checking functions. Rowland O'Flaherty for building the python template, CI, UR jetson docker, and parts of the USD API. Sandeep Desai for setting up the UR robots. Jonathan Tremblay for rendering nvblox meshes and testing cuRobo on a kinova Jaco. Buck Babich, Bryan Peele, and Avi Rudich for building the robot sphere generation utility in NVIDIA Isaac Sim and helpful discussions on minimum jerk trajectories. Miles Macklin and Lukasz Wawrzyniak for adding pytorch CUDA Graph integration into warp and also for helping build the mesh signed distance kernels. Alexander Zook and Hammad Mazhar for debugging CUDA contexts between pyTorch and NVIDIA Omniverse. Felix Wang, Claudia D'Arpino, and Xuning Yang for creating cuRobo's Isaac sim examples. YuShun Hsiao for testing an early prototype of the library and providing feedback on CUDA kernels. 

\bibliographystyle{IEEEtranN}
\clearpage
\appendix
\startcontents[appendices]

\section{Trajectory Optimization: Cost Terms \& Solvers}
\label{app:trajopt}
As mentioned in Section~\ref{sec:mopt}, we setup our trajectory optimization problem to optimize over the joint configurations~$\theta_{t\in[1,T]}$ across~$t$ timesteps. As our implementation of L-BFGS can only handle box constraints on optimization variables, we write all our constraints as cost terms with large penalties. We discuss in detail the cost terms in our trajectory optimization, followed by our technique to optimize for the resolution of time discretization. We then discuss different trajectory profiles that can be obtained using cuRobo in Section~\ref{sec:trajectory_profiles} and details on our solvers in Section~\ref{sec:solver-details}.

\subsection{Pose Reaching Cost}
The main goal of our trajectory optimization problem is to reach a target pose~$X_g$ for the robot's end-effector at the final timestep~$T$. We implement this cost leveraging our differentiable kinematics function~$K_e(\theta_T)$ which gives us the end-effector's position~$p_T\in \mathbb{R}^3$ and the unit quaternion~$\hat{q}_T$. We use a L-2 norm for computing the linear distance~(i.e.,$||p_g - p_T||$) and use the quaternion distance for orientation~$q_g^\top q_T$. We scale these distances~$d$ by~$\alpha_0 \log \cosh(\alpha_2 d)$ to compute the cost. This scaling enables our cost to have both high accuracy near the goal and stable gradients when far away from the goal as shown in Figure~\ref{fig:bound_cost}-(a). Our pose reaching cost term is written as,
\begin{align*}
    C_{\text{goal}}(X_g, \theta_T) &=\alpha_0 \log \cosh(\alpha_2||p_g - p_T||_2)  + \alpha_1 \log \cosh(\alpha_3(q_g^\top q_T)) \numberthis{eq:pose_cost_term}
\end{align*}

While we can also add this cost with a smaller weight across timesteps to encourage the robot to reach the target pose as quickly as possible~(i.e. a running cost), we found this running weight to be sensitive across environments and instead add a weighted joint space path length minimization cost which is discussed next. We use $\alpha_0=2000, \alpha_1=350, \alpha_2=100,$ and  $\alpha_3=100$ in our experiments. 

For tasks that require reaching a joint configuration, we use a similar cost term in the joint position space,
\begin{align*}
    C_{\text{cspace}}(\theta_g, \theta_T) &= \alpha_4 \log \cosh(\alpha_5 ||\theta_g - \theta_T||^2_2) \numberthis{eq:cspace-cost}
\end{align*}
where~$\alpha_4=5000, \alpha_5=50$.
\subsection{Path-Length Minimization \& Smoothness Costs}
\label{sec:smooth}
Our path length minimization cost term is a squared L-2 norm on the joint acceleration across timesteps~$\sum_{t\in[0,T]}||\ddot{q}_t||_2^2$. This term encourages the robot to find a trajectory that has the least acceleration, thereby resulting in ramp up of velocity in one direction for the joints when feasible to reach the goal as that velocity profile will result in the least acceleration. We also have a term that minimizes jerk~$\sum_{t\in[0,T]}||\dddot{q}_t||_2^2$, penalizing large instantaneous changes in acceleration which can cause poor trajectory tracking as discussed in Section~\ref{sec:real-robot-tracking}.

We also want the robot to stop at the final timestep~$T$ which can be encouraged by adding a zero velocity cost term for the final timestep~$T$. However, if we only penalize velocity to be zero at the final timestep, the trajectory obtained could have very large acceleration and jerk.  We hence also want to have zero acceleration and zero jerk at the final timestep~$T$. We can achieve this by penalizing velocity at the last three timesteps~$t\in[T-3,T]$. Similarly, to encourage smooth acceleration at the start of the trajectory, we can add velocity constraints for the first three timesteps~$t\in[0,2]$. Empirically, we found adding these velocity constraints made our optimization become sensitive to the weights we used for these constraints between optimization problems. An alternative way to formulate this constant state criteria is to implicitly make the first and last three states be the same, similar to \cite{toussaint2014komo}. We hence use this implicit formulation to ensure smooth acceleration throughout the trajectory. The cost terms for smoothness are,
\begin{align*}
    C_{\text{smooth}} &= \sum_{t\in Q}\alpha_6 \log \cosh (\alpha_7  \dot{q}_t) + \sum_{t\in[0,T]}(\alpha_8 ||\ddot{q}_t||_2^2 + \alpha_9 ||\dddot{q}_t||_2^2 ) 
    \numberthis{eq:smooth_cost}
\end{align*}
where we use $\alpha_6=5000, \alpha_7=50, \alpha_8=5000$ and $\alpha_9 = 1.0$.

In addition to the above cost terms, we have three constraints for trajectory optimization: (1) enforcing joint limits, (2) avoiding self collisions between links of the robot, and (3) avoiding collisions between the robot and the world. We discussed the collision terms in Section~\ref{sec:collision-avoidance}. We will discuss enforcing joint limits next in Section~\ref{sec:jl}.

\subsection{Enforcing Joint Limits}
\label{sec:jl}
\begin{figure}
    \centering
    \begin{tabular}{c c}
    \includegraphics[width=0.48\textwidth]{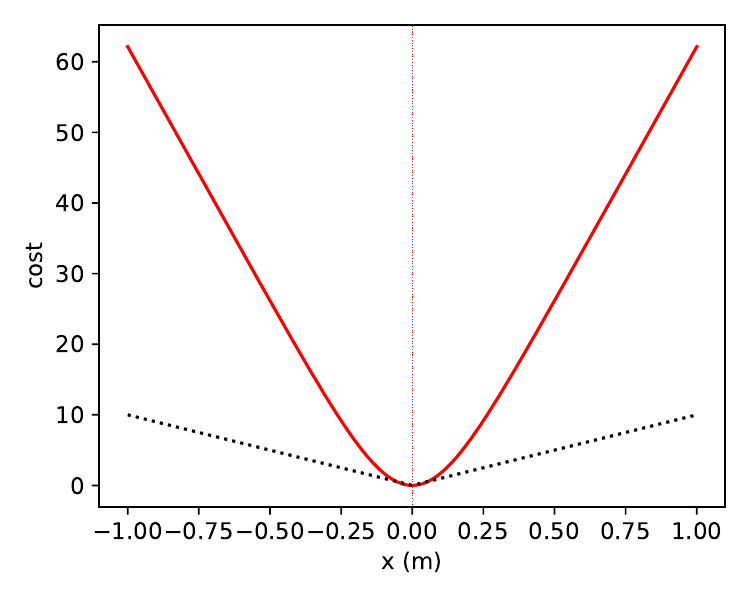} &
    \includegraphics[width=0.48\textwidth]{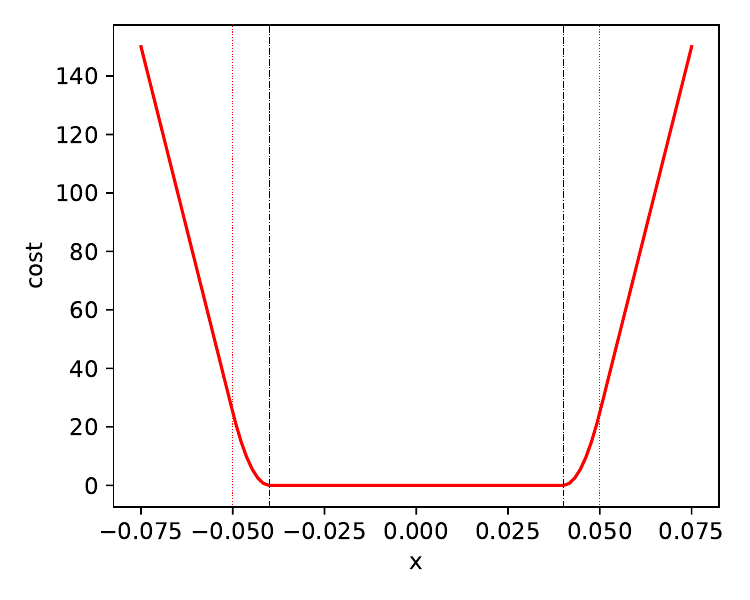} \\
    (a) Pose/CSpace Attraction cost & (b) Bound cost
    \end{tabular}
    \caption{Our attraction cost profile has a tighter curve~shown by the red line, compared to a standard l-2 norm which shown by dotted black line in (a). Our joint limit cost profile, shown in (b) transitions smoothly from a quadratic increase to a linear increase avoiding sudden linear jump that can happen right at a limit where the limit is shown by the red vertical dotted line.}
    \label{fig:bound_cost}
\end{figure}

To minimize the optimization from hitting joint position, velocity, acceleration, and jerk limits we add a smooth cost motivated by the potential function introduced by Ratliff~\etal~\cite{ratliff2009chomp}. For a joint value~$q_t$, we define the cost term as
\begin{align*}
    c_p &= \begin{cases}
        q_l - q_t + 0.5  \eta_2  & \text{if } q_t < q_l \\
        \frac{0.5}{\eta_2}  (q_l - q_t + \eta_2)^2  & \text{if } q_l + \eta_2 > q_t \geq q_l \\
        q_t - q_u + 0.5  \eta_2  & \text{if } q_t > q_u \\
        \frac{0.5}{\eta_2}  (q_t - q_u + \eta_2)^2  & \text{if } q_u - \eta_2 < q_t \leq q_u \\
        0 & \text{otherwise}
    \end{cases} \numberthis{eq:bound_cost}
\end{align*}
where~$q_u, q_l$ are the upper and lower joint limits respectively. The parameter $\eta_2$ defines the threshold before a joint limit from where we start penalizing the joint value. The profile of this cost is shown in Figure~\ref{fig:bound_cost}-(b). We use a similar cost term for velocity, acceleration, and jerk limits, all with an activation distance of $\eta_2=0.1$.

\subsection{Time-step Descretization and Temporal Weight Scaling}
\label{app:timeopt}
To perform trajectory optimization across the full workspace of the robot, we need to be able to solve for both very short and very large motions. One way to do this is to attempt to solve the trajectory optimization problem, first with a very large number of timesteps~$t$, and then sequentially reduce the timesteps until the optimization fails to converge as done by Ichnowski~\etal~\cite{ichnowski2019motion}. However, this requires us to run sequential calls to trajectory optimization which can be slow on the GPU, especially when we have to reinitialize all buffers that are affected by the number of timesteps~$t$. Another option is to treat~$dt$ also as part of the optimization variables, which we do not explore in this work. 

In our current implementation, we fix the number of timesteps~$t$ and instead change the time discretization~$dt$ between timesteps to be able to solve motions of all ranges as shown in Algorithm~\ref{alg:timestep}. To run optimization with different~$dt$, we scale all our cost terms that relate to velocity, acceleration, and jerk to account for the change in magnitude of~$dt$. We first run trajectory optimization with a very large~$dt_i$ of 0.25 seconds and then scale the output trajectory to find a~$dt_o$ that pushes the velocity, acceleration or jerk to the robot's limits. We then scale our weights by~$dt_o$
and rerun trajectory optimization with~$dt_o$ to get the final trajectory. We then re-time this final trajectory to find the best~$dt_{f}$. We empirically found that not penalizing jerk during the first trajectory optimization led to better convergence.

    \begin{algorithm}
        \DontPrintSemicolon
        \algokw
        \caption{Time Discretization}\label{alg:timestep}
        \KwInit{$\Theta_{i}$, $dt_{i}$}
        \KwOut{$\Theta_f$, $dt_f$}
        $\Theta_{o} \gets$ traj\_opt($\Theta_{init}$, $dt_{i}$) \tcp{run trajectory optimization with initial dt}
        $dt_{opt} \gets$ retime($\Theta_{o}$) \tcp{find dt that pushes trajectory to robot limits (velocity, acceleration or jerk)}
        scale\_traj\_opt($dt_{opt}$) \tcp{Scale weights by new dt}
        enable\_jerk\_cost() \tcp{enable jerk minimization cost term}
        $\Theta_f\gets$ traj\_opt($\Theta_{o}$, $dt_{opt}$) \tcp{run trajectory optimization with new dt}
        $dt_f\gets$retime($\Theta_{o}$)\tcp{get final dt by retiming the final trajectory}
    \end{algorithm}

\subsection{Finite Difference for State Derivatives}
\label{sec:finite_difference}
Our optimization variables are in the joint position space~$q_{t\in[0,T]}$ while we have cost terms in the velocity, acceleration, and jerk spaces. A standard technique to compute time derivatives is by backward finite difference, where we finite difference position to get velocity, then finite difference velocity to get acceleration, and finally finite difference acceleration to get jerk. However, we observed oscillations on the real robot when executing high-speed trajectories due to the low accuracy of backward finite difference in computing velocities. We then tried central difference and that got rid of the oscillations on real robot execution. The optimization also required more iterations to converge compared to backward difference. 

We investigated finite difference approaches from Pose graph optimization literature and found the five point stencil method to be a good fit for computing time derivatives in our trajectory optimization problem. With a five point stencil method, we read five adjacent joint positions to compute velocity, acceleration, and jerk values with much higher precision than central or backward difference. An added benefit of five point stencil method for computing time derivatives is that any gradients from time derivatives have a blending effect on the position space~\cite{wedel2009improved}. The improved accuracy and the blending effect reduced the number of iterations required to converge by 50\%. The trajectories obtained with the stencil approach also ran successfully on the real robot as shown by our tracking results in Section~\ref{sec:real-robot-tracking}. Across all these methods for computing derivatives, we observed aliasing artifacts in the acceleration space when our jerk minimization cost was enabled. The aliasing was larger in backward difference, reduced in the central difference approach and even smaller in the five point stencil difference. This artifact could be reduced further by using even larger number of points in the stencil. However, we did not observe any issues on the real robot when executing the five point stencil trajectories. In our current implementation, we smooth out the acceleration artifacts with an average sliding window filter after convergence.

\subsection{Trajectory Profiles}
\label{sec:trajectory_profiles}
\begin{figure}
    \centering
    \begin{tabular}{c c c}
        \includegraphics[width=0.3\textwidth]{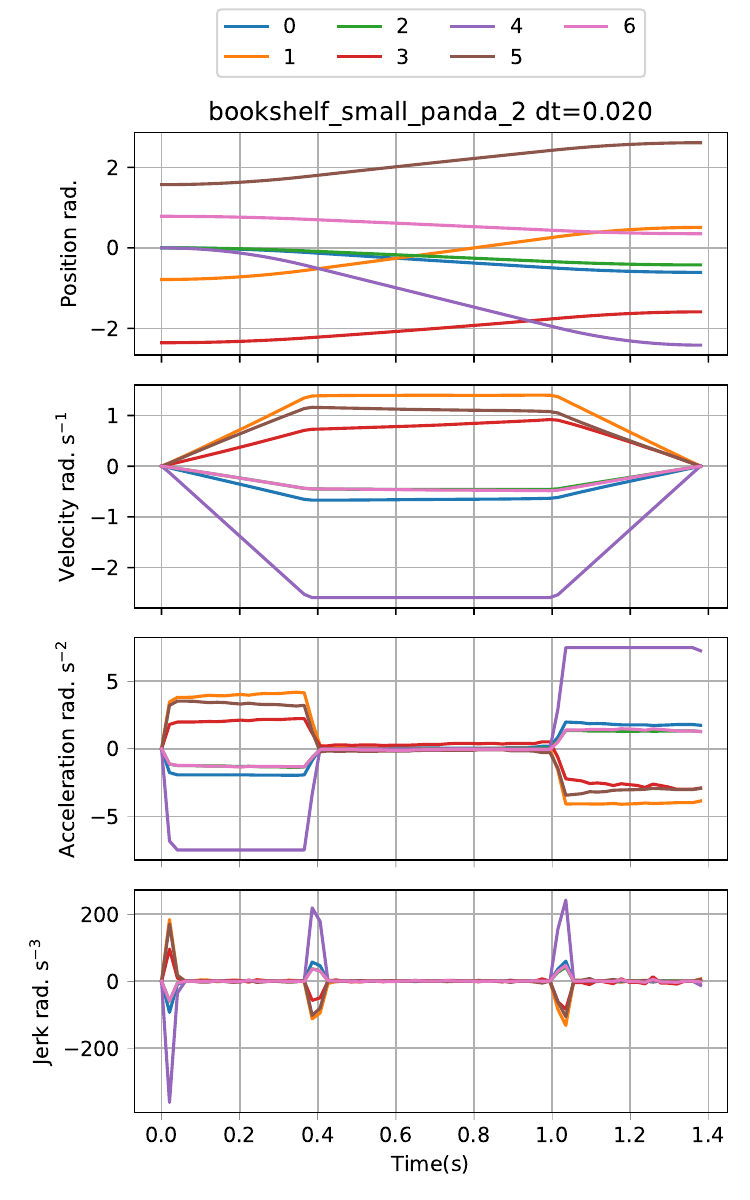}
        &
        \includegraphics[width=0.3\textwidth]{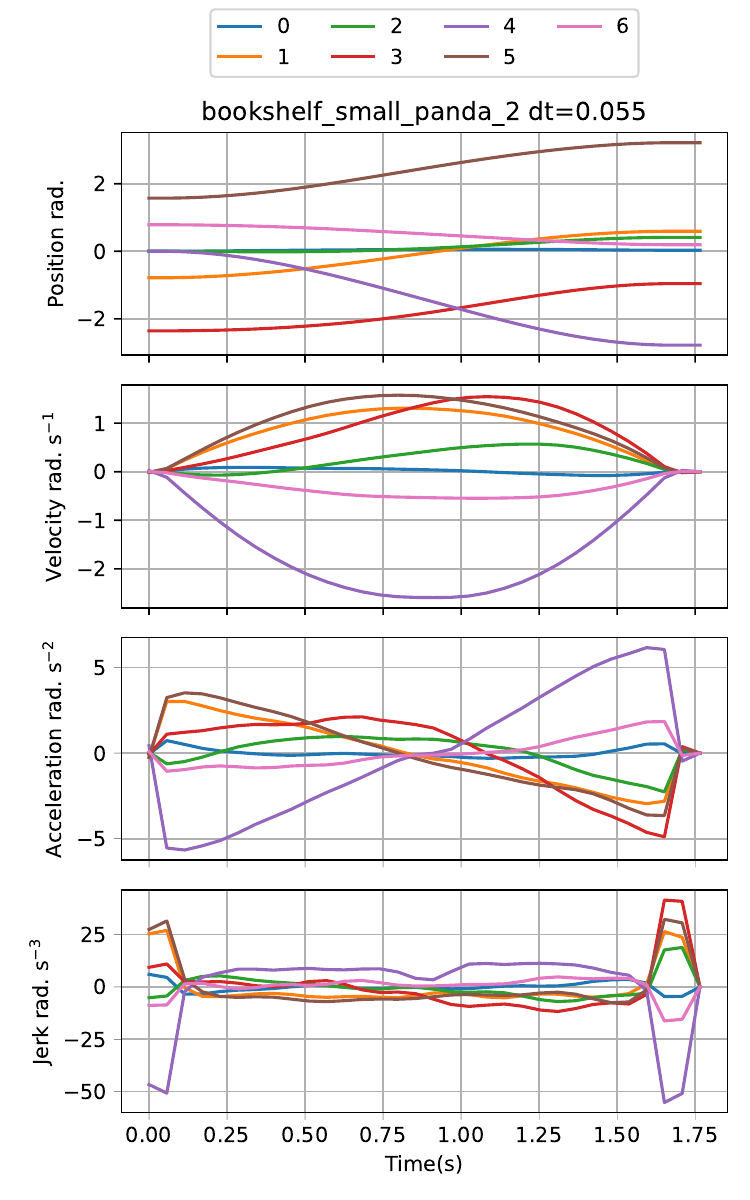}
        &
        \includegraphics[width=0.3\textwidth]{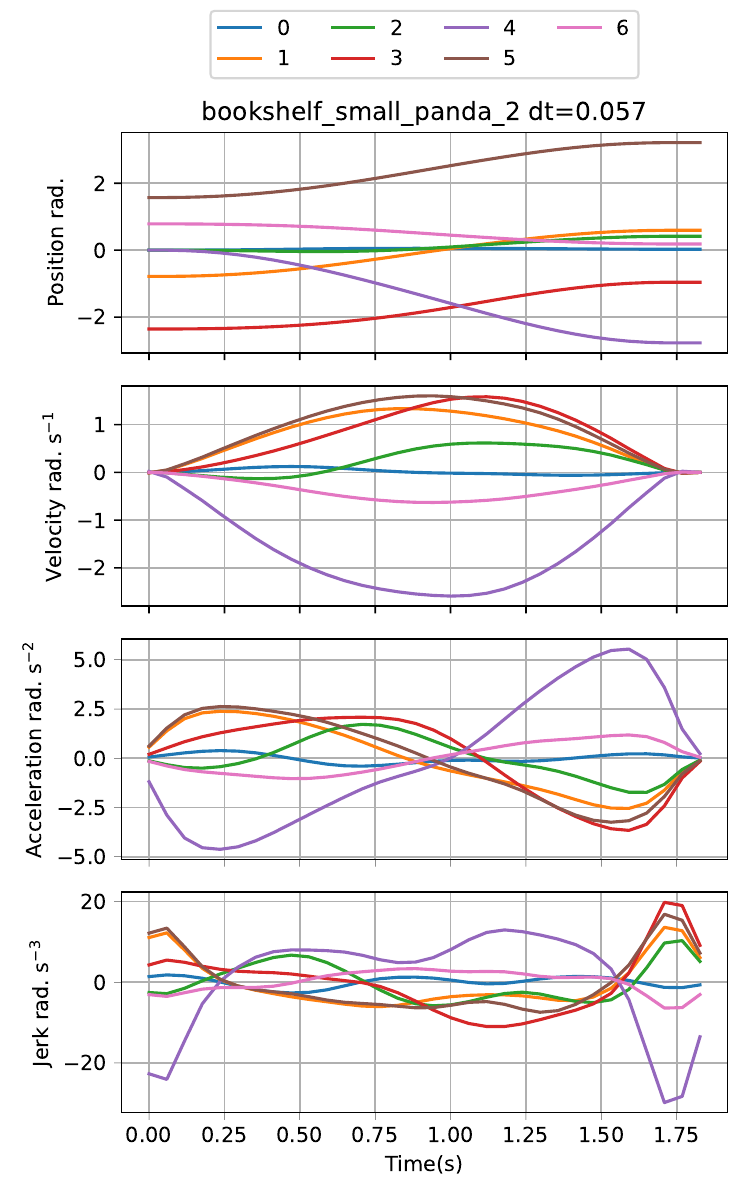}\\
        (a) Kunz and Stillman~\cite{kunz2012time}  & (b) Minimum acceleration & (c) Minimum jerk
    \end{tabular}
    \caption{We show the different trajectory profiles possible leveraging cuRobo's trajectory optimization. The first way to get a bounded acceleration and velocity profile is by using Kunz and Stilman's trajectory post processing~\cite{kunz2012time}, shown in (a). This can have very high jerks due to instantaneous accelerations. We can obtain a better trajectory profile in cuRobo using a minimum acceleration cost as shown in (b), which can still have instantaneous jerks. By adding a minimum jerk cost term, we can get an even smoother trajectory profile as shown in (c).}
    \label{fig:trajectory_profile_plot}
\end{figure}

Trajectory profiles have been studied~\cite{kunz2012time,berscheid2021jerk} as a post processing step to geometric path planning, where methods try to obtain a time-optimal solution with bounded limits on velocity, acceleration, and possibly jerk. Kunz and Stilman~\cite{kunz2012time} explored a technique to traverse waypoints without stopping at them by relaxing the accuracy in reaching the waypoints. Their approach was able to obtain trajectories with bounded velocity and acceleration. However, their trajectories have very large jerks as shown in Figure~\ref{fig:trajectory_profile_plot}-(a). Ruckig~\cite{berscheid2021jerk} explored introducing jerk constraints, which constrained the solution to have zero velocity, acceleration, and jerk at the starting timestep and the final timestep. However, their solution was only optimal when the path had no waypoints between the start and goal configuration. They provide a solution to also solve for intermediate waypoints but do mention that it's not the optimal solution~(\href{https://docs.ruckig.com/md_pages__intermediate_waypoints.html}{ruckig documentation}). In addition, we were unable to evaluate their waypoint solver as it is not accessible without a license. 

Given that there is no technique that can output bounded jerk, acceleration, and velocity trajectories with zero velocity, acceleration, and jerk at start and final timesteps, we use our cost terms from Sec~\ref{sec:smooth} to encourage this trajectory profile. Our framework also enables getting different trajectory profiles by scaling the relative weights between cost terms as shown in Table~\ref{tab:trajectory_profile_cost} and Figure~\ref{fig:trajectory_profile_plot}. We also integrate our library with Kunz and Stilman's method~\cite{kunz2012time} through a python wrapper~(\href{https://github.com/balakumar-s/trajectory_smoothing}{github\_link}) to output bounded velocity and acceleration trajectories for applications that allow for large jerks.

\begin{table}
    \centering
    \begin{tabular}{l l l l}
    \toprule
    Constraints  & Acceleration-l2 & Jerk-l2 & Effect\\  \toprule 
      None & None & None & Starts with non-zero velocity.\\
      None & All timesteps & None & Starts with non-zero velocity.\\
      $\dot{x}_{t\in[0,T]} = 0$ &  None & None & Not smooth.\\ \midrule
      $\dot{x}_{t\in[0,T]} = 0$ &  All timesteps & None & Large Jerk at first, last $t$.\\
      $\dot{x}_{t\in[0,T]} = 0 $&  All timesteps &  All timesteps & Low jerk.\\ %
       $\dot{x}_{t\in[0,3], [T-3,T]} = 0$&  All timesteps &  All timesteps & Low jerk, hard to optimize.\\ \midrule
      $x_{t\in[1,3]},x_{t\in[T-3,T-1]} = x_0,x_T$ &  All timesteps &  All timesteps & Low jerk, easier to opimize.\\ \bottomrule
      
    \end{tabular}
    \caption{Trajectory profiles by penalizing position derivatives.}
    \label{tab:trajectory_profile_cost}
\end{table}

\subsection{Numerical Optimization Solvers}
\label{sec:solver-details}
\begin{figure}
    \centering
    \includegraphics[width=0.98\textwidth]{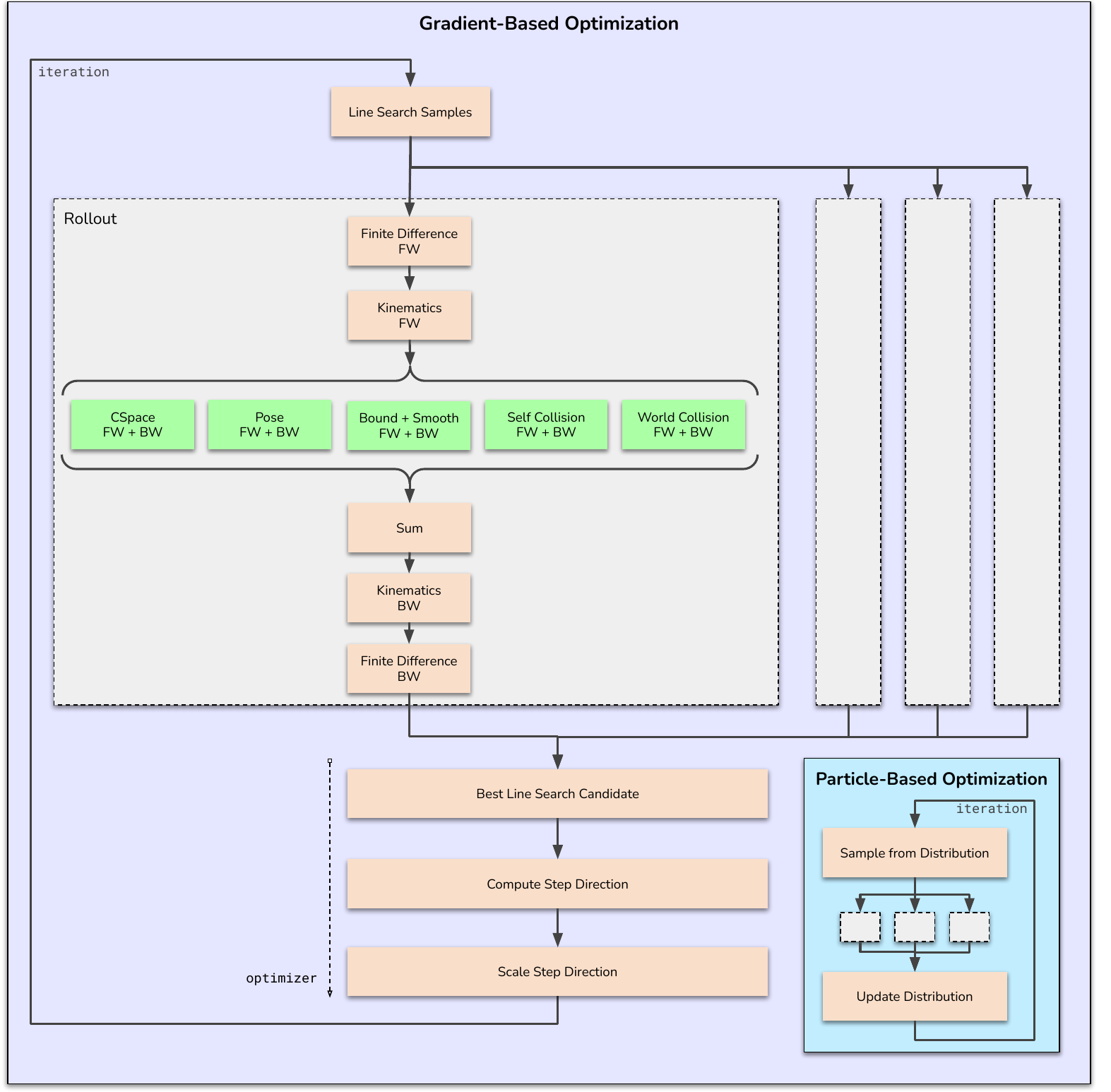}
    \caption{The compute graph in a single iteration of optimization is shown for both Gradient-based and Particle-based optimization in this figure (best viewed in color). The grey-block is the~\emph{Rollout} class that computes the cost given a batch of action trajectories. We leverage parallel execution of~\emph{Rollout} class to run parallel line search in gradient-based optimization and to compute cost across many samples in particle-based optimization. The green blocks highlight the different cost terms used for motion generation.}
    \label{fig:solver-design}
\end{figure}

We design our solvers to take in a rollout class instance that provides a differentiable map from optimization parameters to the total cost. Our rollout class also takes in a batch of queries and outputs a batch of total costs. This batch query-able rollout class design allows us to use it for computing the cost for all samples in a particle based solver in parallel. This also enables us to compute the differentiable maps for all line search magnitudes in parallel. We illustrate thus design of our solvers and the rollout class in Figure~\ref{fig:solver-design}. We detail the L-BFGS step update in Algorithm~\ref{alg:lbfgs-step} which follows the algorithm from Nocedal and Wright~\cite{nocedal1999numerical}. Our algorithm for particle-based optimization is in Algorithm~\ref{alg:particle}.

Once we compute the step direction, we have the option to clamp the step direction by~$\gamma (q_{H}-q_{L})$ to avoid the step direction from exploding~(i.e., out of numerical precision) due to large violations of soft constraints~(e.g., collision term). Scaling step direction is a common technique to add robustness to a numerical solver and has also been done in trajectory optimization previously by Ratliff~\etal~\cite{ratliff2009chomp}. In the environments and problems we tested, we did not see significant improvement with scaling the step direction and hence did not use it. 
\begin{algorithm}
\DontPrintSemicolon
\algokw
\caption{Particle-Based Optimization}\label{alg:particle}
\KwProb{$\Theta_\text{init}$, $C(\cdot)$}
\KwIn{$\sigma_0, c_\text{min}$}
\KwResult{$\mu$}
\KwInit{$\mu \gets \Theta_\text{init} $, $\sigma \gets \sigma_0$}

\For (\tcp*[f]{Run optimization}){$n\gets0$ \KwTo $N$}{ 
$\Theta_l \gets$SAMPLE($\mu$, $\sigma$)\;
$c_l \gets C(\Theta_l)$\;
$\mu, \sigma \gets $ UPDATE($c_l, \Theta_l, \mu, \sigma$) \;
}
\end{algorithm}

    \begin{algorithm}
        \DontPrintSemicolon
        \algokw
        \caption{L-BFGS Step Update}
        \label{alg:lbfgs-step}
        \KwInit{$q\gets \delta \Theta$}
        $y_{[0,k]}, s_{[0,k]}, \rho_{[0,k]}  \gets \text{sh}(y_{[0,k]},-1)$, $\text{sh}(s_{[0,k]},-1)$, $\text{sh}(\rho_{[0,k]},-1)$ \tcp*{Shift Buffers}
        $y_k \gets \delta \theta - \delta \Theta_{-1}$ \tcp*{Update Buffers}
        
        $s_k \gets \Theta - \Theta_{-1}$ \;
        
        $\rho_k \gets 1/{y_k^\top s_k}$\;
        $\Theta_{-1}, \delta \Theta_{-1} \gets \Theta, \delta \Theta$ \tcp*{Store for next iteration}        
        \tcc{Serial compute}
        \For(){$i=k-1$ \KwTo $k-m$}{
            $\alpha_i \gets \rho_i s_i^\top q$

            $q \gets q - \alpha_i y_i$
        }
        
        $r \gets H^0_K q$ \;
        \tcc{Serial compute}
        \For(){$i=k-m$ \KwTo $k-1$}{
        $\beta \gets \rho_i y_i^\top r$\;
        $r \gets r + s_i (\alpha_i - \beta)$
        }
        $\Delta \Theta \gets r$\;
    \end{algorithm}

\subsection{Tuning Weights \& Parameters}
\label{app:traj-tune}
Trajectory optimization with numerical solvers is often very sensitive to the weights used between cost terms, especially when the solvers don't solve for hard constraints directly. We found the following procedure very helpful in finding a good set of weights for the different cost terms,
\begin{enumerate}
    \item Start the tuning process with a large number of seeds for IK~(e.g., 500) and trajectory optimization~(100). Also use a large number of timesteps~(40) for trajectory optimization.
    \item Tune pose cost term and collision cost term to solve IK. 
    \item Then, tune Pose cost and smoothness cost for trajectory optimization with collision costs disabled. 
    \item Finally, tune collision cost weights for trajectory optimization.
\end{enumerate}
Once you have a good set of weights, start reducing the number of seeds until the weights do not work anymore. At this step in the process, we found increasing the weights for the pose cost improved success rate. Once we tuned our weights following these steps, the weights transferred across different environments and also across different robot platforms.  

\subsection{Evaluating Optimized Trajectories}
Our inverse kinematics solver and trajectory optimization solver optimizes over many parallel seeds. After completing iterations, we often found more than one valid solution from these seeds. For inverse kinematics, we return the solution with a lowest weighted sum of pose error and the distance of the solution to the current joint configuration. For choosing between the valid trajectory optimization seeds, we use a blended sum of the pose error, maximum jerk, and motion time. 

\section{Benchmarking Details}
\label{app:baseline}
\subsection{Dataset \& Evaluator}
We benchmarked motion planning with the 800 problems from the motion benchmaker dataset~\cite{chamzas2021motionbenchmaker} and the 1600 problems from the mpinets dataset~\cite{fishman2022mpinets}(Version 1.0.2). The motion planning problems available through the motion benchmaker dataset~\cite{chamzas2021motionbenchmaker} had the Franka Panda's gripper at a 16cm opening which is outside of the real Franka Panda's 8cm limit. We reduce the radius of the obstacle~(cylinder) that was the between the gripper by 4cm to make it fit within the Franka Panda's real gripper limits. The mpinets dataset~\cite{fishman2022mpinets} uses a gripper width of 2.5cm, which we set in our robot configuration during benchmarking. We retooled the evaluator that was written by~\cite{fishman2022mpinets}, which is available at \href{https://github.com/fishbotics/robometrics}{github.com/fishbotics/robometrics} along with the dataset.

\subsection{Tesseract Baseline}
We used Tesseract planning with commit~\texttt{1627231f3d}, compiled with~\texttt{Release} CMAKE flag for our experiments. As discussed in Section~\ref{sec:trajectory_profiles}, we add costs to Trajopt that have the first three and last three timestep velocity to be at zero. We also set~\texttt{smooth\_acceleration=True} and set~\texttt{smooth\_velocity=False} to get the solution as mentioned by Tesseract developers (\href{https://github.com/tesseract-robotics/tesseract_planning/discussions/190}{github-discussion}). We tried setting~\texttt{smooth\_jerk=True}, however this caused the planner to fail on many planning problems and the trajectories also exhibited aliasing in the acceleration.

In addition, we couldn't run the Cartesian planning from Tesseract. We compute 20 IK solutions using cuRobo's collision-free IK solver for each problem and use one of these solutions per attempt of planning in Tesseract. We found that Tesseract failed to find many trajectories without offsetting the collision margin by -1.5cm. Hence, for all our comparisons we use a collision margin of -1.5cm. In addition, the OMPL planning phase failed to find solutions with an offset of -1.5cm, hence we further added an offset of -2cm only to the OMPL planning phase. We enable bullet continuous collision checking for the whole pipeline in Tesseract to leverage the fastest implementation of collision checking in their pipeline. We provide the Tesseract workspace with all changes at \href{https://github.com/balakumar-s/tesseract_ws}{tesseract\_ws} and also develop a python wrapper for benchmarking which is available at \href{https://github.com/balakumar-s/tesseract_wrapper}{github.com/balakumar-s/tesseract\_wrapper}.

\subsection{cuRobo Parameters}
\label{app:curobo-parameters}
We use an activation distance~$\eta$ of 2.5cm and use a scalar weight of 5000 for all cost terms that are soft constraints. We tuned the number of trajectory optimization~(TO) seeds on a per scene level depending on average number of attempts needed to succeed when we ran with only 4 seeds. We used 30 seeds for Inverse Kinematics for most problems. For motion planning problems in~\emph{Table under Pick Panda}, we generate seeds for trajectory optimization from our geometric planner~(Section~\ref{sec:graph}) as we found that we need more than 100 linearly interpolated seeds to succeed. For trajectories that require many switching points, we use a larger number of timesteps. The values used across the scenes are shown in Tab.~\ref{tab:seed-set}. We run 100 fixed iterations followed by 300 variable iterations for the trajectory optimization. We run 2 iterations of particle-based optimization before running L-BFGS for IK and Trajectory optimization. For our evaluations, we have a 2 second warm-up phase where the tensors need to be initialized, followed by CUDA Graph creation on the GPU. After this warm-up phase is complete, we can change the environment obstacles, the start configuration, and the goal before querying a solution for a motion generation problem.
\begin{table}
    \centering
    \begin{tabular}{l r r r r}
    \toprule
        \textbf{Planning Environment} & \textbf{IK seeds} & \textbf{TO seeds}  & \textbf{Time-steps} & \textbf{Force Graph} \\ \toprule
         Bookshelf Small Panda &  30 & 12 & 32 & False \\
         Bookshelf Tall Panda &  30 & 12 & 32 & False \\
         Bookshelf Thin Panda &  30  & 12 & 32 &False \\ \midrule
         Box Panda &  30 & 12 & 32 & False \\
         Box Panda Flipped &  30 & 12 & 32 & False \\ \midrule
         Cage Panda & \cellcolor{bargreenlight} 100 & \cellcolor{bargreenlight} 16 & 32 & False \\
         Table Pick Panda & 30 & 12 & 32 & False \\
         Table under Pick Panda & \cellcolor{bargreenlight} 112 & \cellcolor{bargreenlight} 28 & \cellcolor{bargreenlight} 44 & \cellcolor{bargreenlight} True \\ \midrule
          Table Top & 30 & 12 & 32 & False \\
          Merged Cubby & 30 & 12 & 32 &False \\
          Cubby &  30 & 12 & 32 & False \\
          Dresser &  30 & 12 & 32 & False \\ \midrule
          Cubby-task-oriented & \cellcolor{bargreenlight} 100 & \cellcolor{bargreenlight} 16 & 32 & False \\
    Dresser-task-oriented & \cellcolor{bargreenlight} 100 & \cellcolor{bargreenlight} 20 & 32 & False \\ 
        \bottomrule
    \end{tabular}
    \caption{Planning Parameters used in \emph{cuRobo} across the benchmark.}
    \label{tab:seed-set}
\end{table}

\clearpage
\section{Additional Results}
\subsection{Comparison to pyBullet and RRTStar}
\label{app:rrtstar}
\begin{figure}[h]
    \centering
    \includegraphics[width=0.8\textwidth]{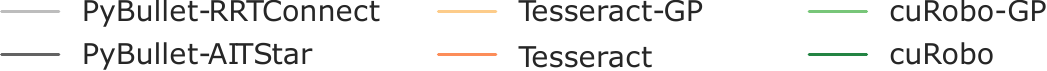}
    \includegraphics[width=0.98\textwidth, trim=0.5cm 0 6cm 0, clip]{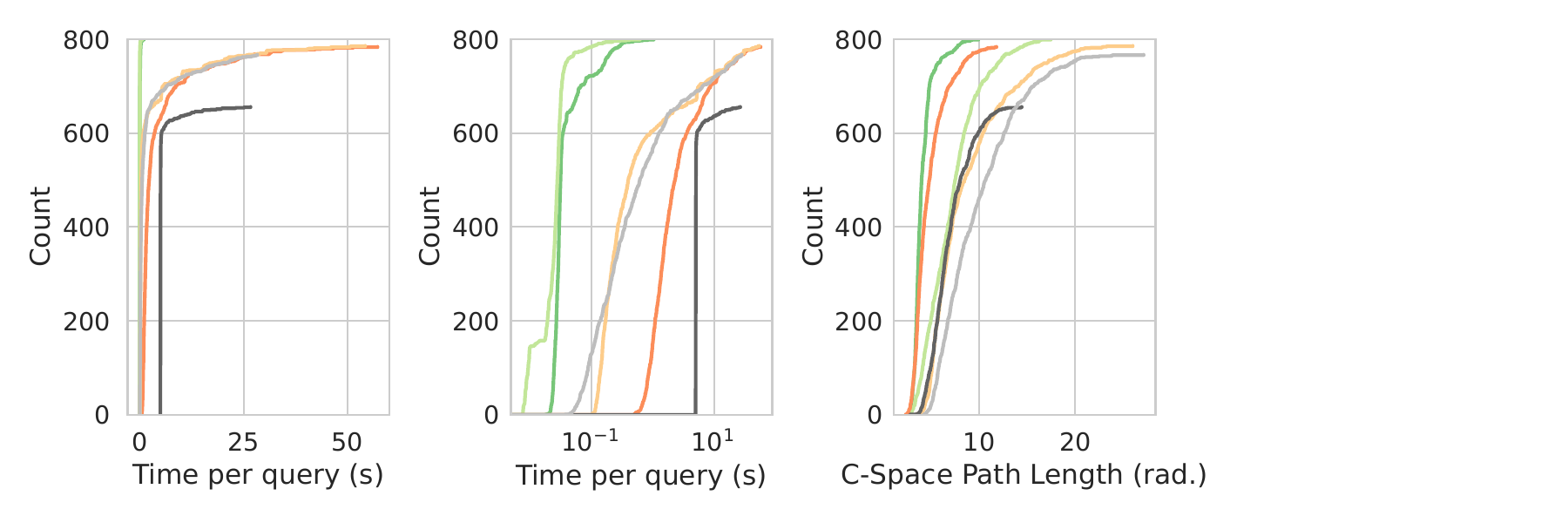}
    \caption{We compare the time taken by different motion generation methods in linear time scale on the left and log time scale in the middle. We see that~\emph{cuRobo}'s curve is significantly faster than \emph{PyBullet}. On the right, we plot the C-Space path length across methods where we see again that \emph{cuRobo} produces shorter paths than \emph{PyBullet-AITStar}.
    }
    \label{fig:motion_opt_mb_plot}
\end{figure}

In addition to comparing to \emph{Tesseract}, we compare our method with pybullet wrapped OMPL implementations which are commonly used in many research efforts as they are accessible from Python~\cite{garrett2020pddlstream}.  We specifically compare to geometric planning methods,  RRTConnect~\cite{kuffner2000rrt} which is a bidirectional feasible planner that is shown to be the fastest in finding a path (while the path may not be optimal) and AITStar~\cite{strub2020adaptively} which is an asymptotically optimal planner that is proven to converge to the shortest path given infinite time.  We use their implementations from OMPL~\cite{sucan2012open} with  PyBullet~\cite{coumans2021} for collision checking. We call them~\emph{PyBullet-RRTConnect} and~\emph{PyBullet-AITStar}, respectively. We only compare on the motionbenchmaker dataset~\cite{chamzas2021motionbenchmaker} as we found it. To be significantly slower compared to trajectory optimization methods while also producing longer c-space path lengths as shown in Figure~\ref{fig:motion_opt_mb_plot}. We found \emph{PyBullet-RRTStar} to be faster than \emph{Tesseract-GP} which uses RRTConnect from ompl wrapped with Bullet's continuous collision checker.   

\subsection{Compute Time Coverage Plot}

\begin{figure}
    \centering
    \includegraphics[width=0.8\textwidth]{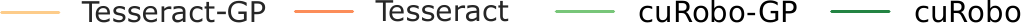}
    \begin{tabular}{c c}
    \includegraphics[width=0.48\textwidth, trim={0 0 0 0}, clip]{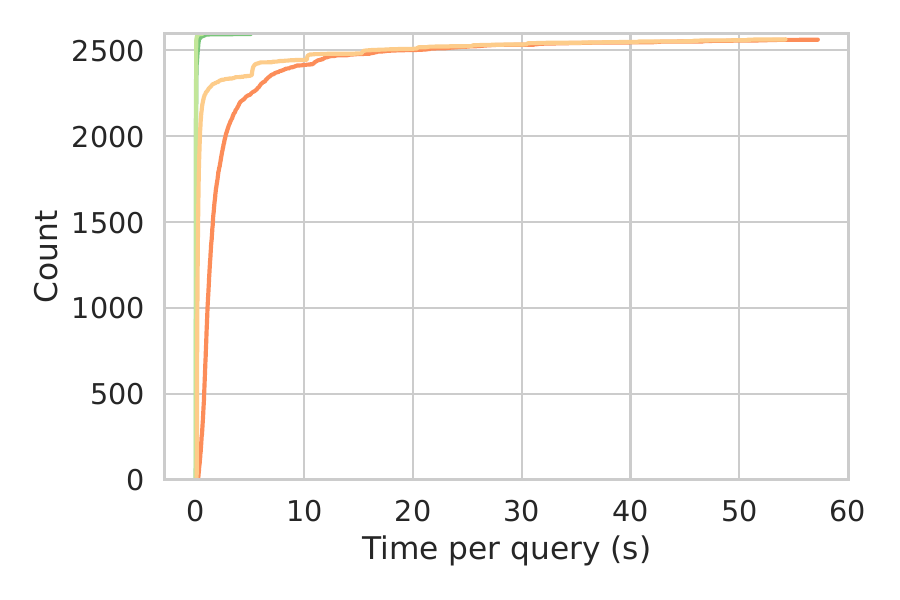} & 
    \includegraphics[width=0.48\textwidth,  trim={0 0 0 0}, clip]{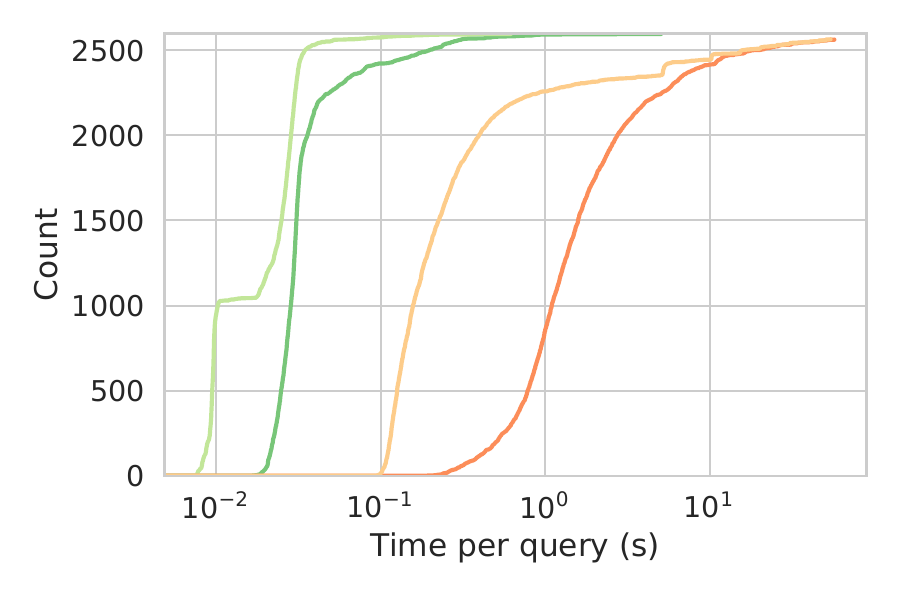}\\
    (a) PC  & (b) PC log scale \\
    \includegraphics[width=0.48\textwidth]{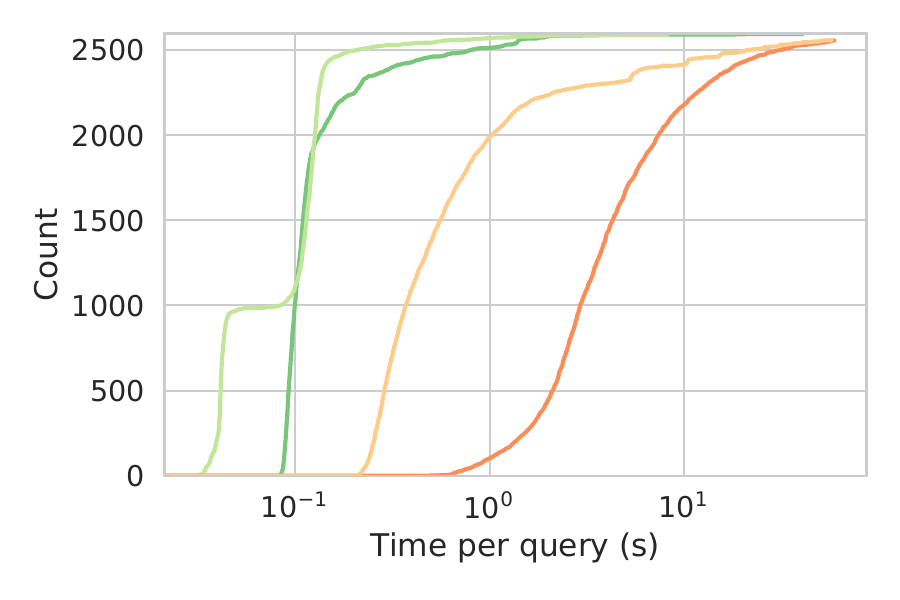}
    &
    \includegraphics[width=0.48\textwidth]{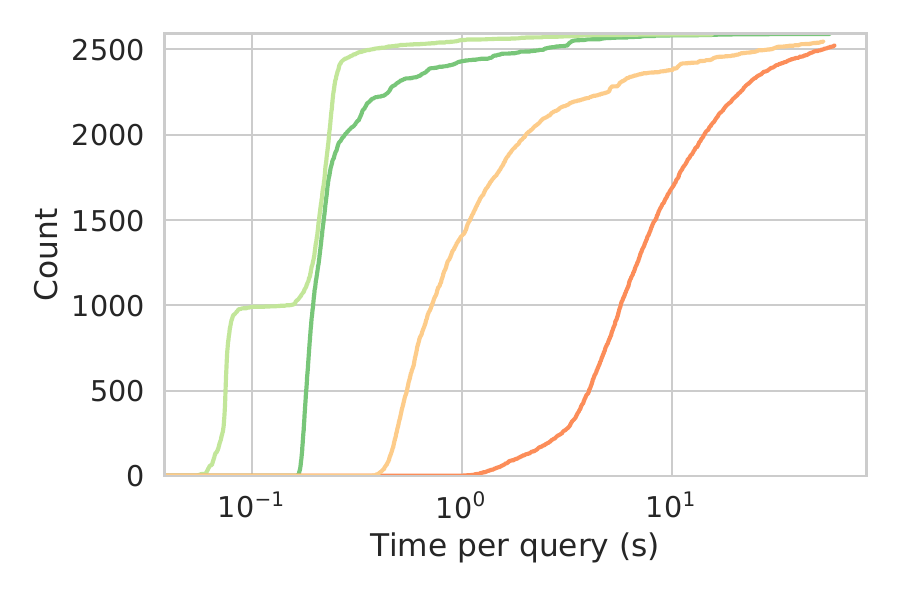}\\
    (c) Orin MAXN & (d) Orin 15W
    \end{tabular}
    \caption{We compare the planning time obtained by our approach~\emph{cuRobo} to~\emph{Tesseract}.}
\label{fig:motion_opt_plot}
\end{figure}
We plot the distribution of compute time across the dataset for methods in \emph{Tesseract} and \emph{cuRobo} in Figure~\ref{fig:motion_opt_plot}. From the distribution across different compute devices, we can observe that \emph{cuRobo} is faster than both \emph{Tesseract} and \emph{Tesseract-GP}. Our geometric planner~\emph{cuRobo-GP} is faster than our trajectory optimization approach~\emph{cuRobo} across different compute platforms. 
\subsection{Real Robot Quirks}
\begin{figure}
    \centering
    \begin{tabular}{c c}
        \includegraphics[width=0.45\textwidth]{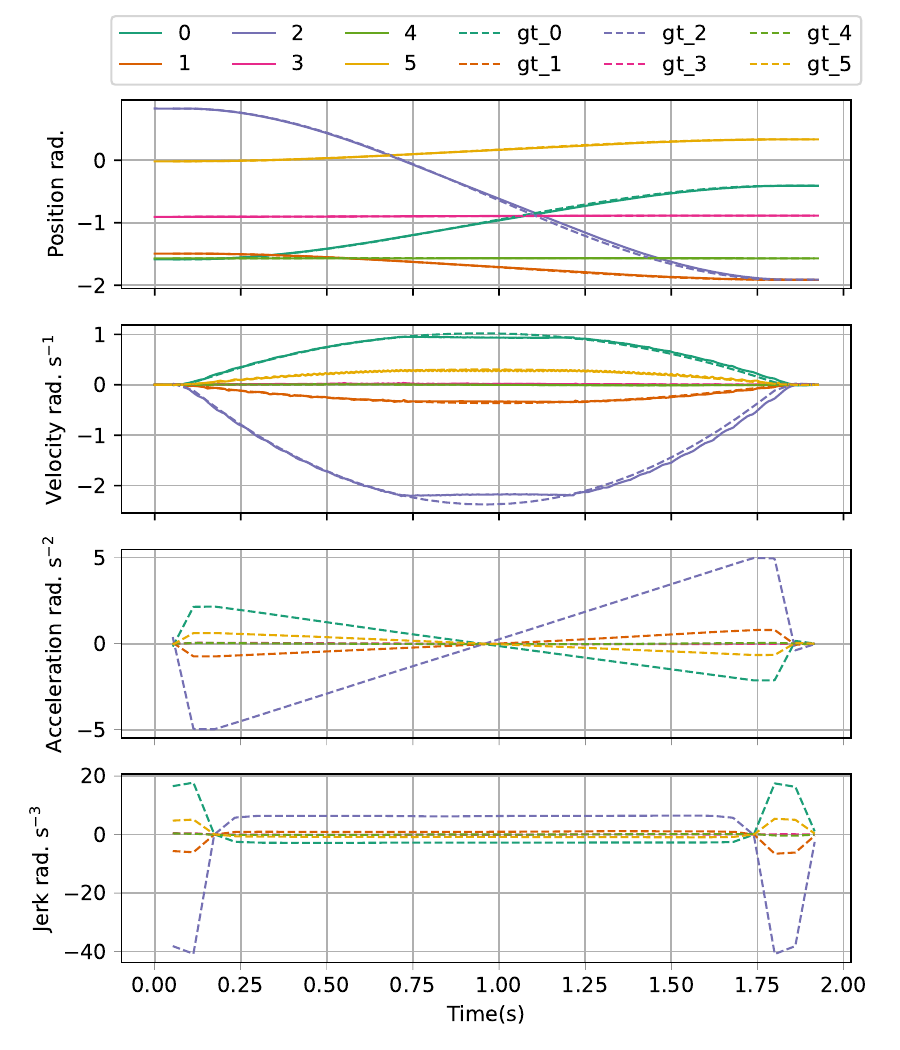} & \includegraphics[width=0.45\textwidth]{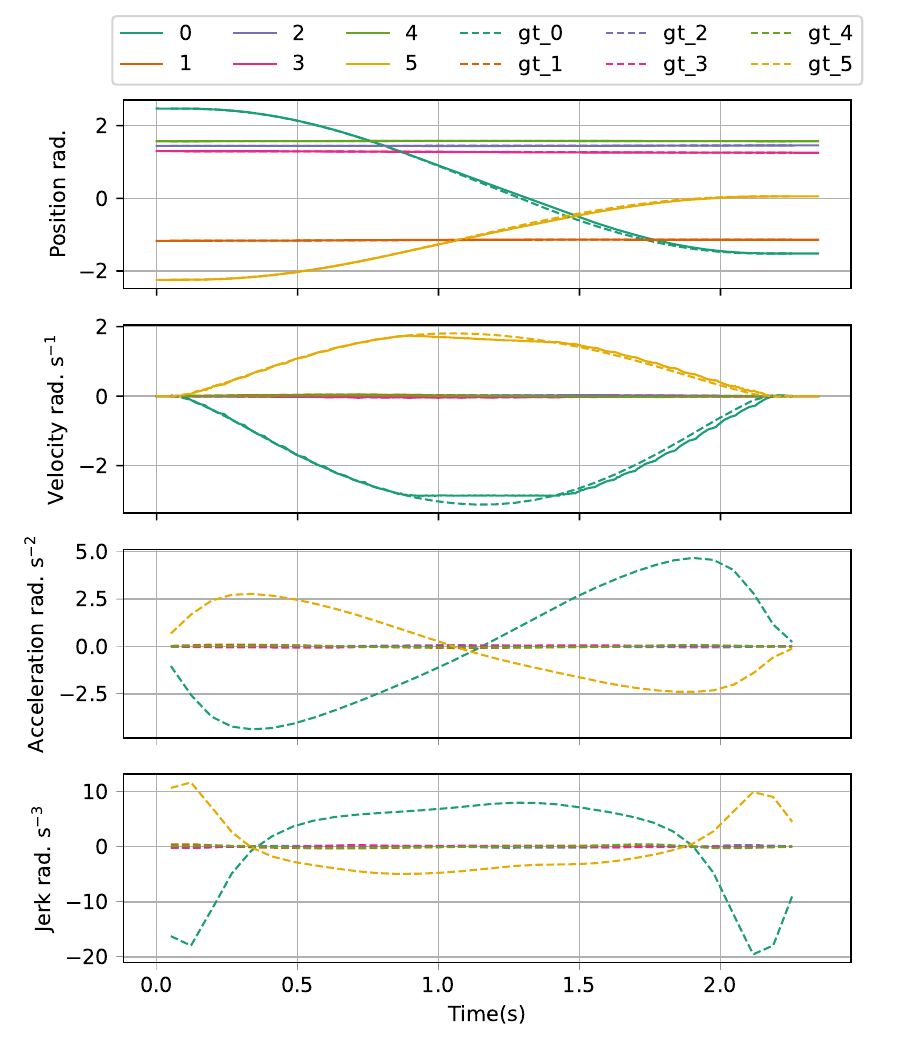} \\
         (a) Minimum Acceleration & (b) Minimum Jerk \\ 
    \end{tabular}
    \caption{UR5e robot clipped peak velocities when executing some motions as show in (a) and (b). This happens when the robot's safety mode is enabled. Once we disabled the safety mode, the robot was able to execute trajectories with accelerations up to 12 radians$s^{-2}$, reaching maximum joint velocities.}
    \label{fig:fail-execution}
\end{figure}

We record some real robot quirks we observed when deploying our trajectories on the Universal Robots. First, we observed poor tracking as we increased the maximum allowed acceleration during trajectory optimization, and also occasionally at peak velocities of the robot as shown in Figure~\ref{fig:fail-execution}. The UR robots come with safety features enabled by default, which restricts the speed of the robot. We overcame this issue by selecting the least conservative safety configuration which set the maximum power limit of the robot to 1000W (from 300W). 

A second issue we observed with the Universal Robots was that trajectories with velocities computed through backward difference caused oscillations at high speeds. Switching to higher accuracy finite difference approaches such as central difference or five point stencil difference fixed this issue.

\section{cuRobo Library}
\label{app:curobo-library}
Implementing the algorithms discussed in this work requires a robotics toolkit that works seamlessly with CUDA code, interfaces with robot configuration files (urdf, usd), and also enables users to implement their own modules and cost terms without adding large overheads to the pipeline. Additionally, most of our algorithms run native on a GPU, requiring the framework to handle GPU tensors. We couldn't find a robotics toolkit that had these features so we built our own robotics framework~\emph{cuRobo}. We build our framework with PyTorch as the front-end enabling us to leverage the vast range of tools built by the PyTorch community. We specifically leverage the following from PyTorch:
\begin{enumerate}
    \item Differentiable mapping between operations on tensors, enabling us to build a compute graph that contains the forward and backward passes for use with our numerical optimization solvers.
    \item Profiling tools to analyze compute graphs and optimize bottlenecks. 
    \item CUDA code compilation and execution with pyTorch tensors enables us to write sophisticated high-performance algorithms as CUDA kernels and access them from python.
    \item CUDA Graph creation and execution reduces kernel launch overheads, we found this to reduce our compute time significantly~(10$\times$ faster) as we run 25 iterations of our optimization as a single CUDA graph call.
\end{enumerate}

\begin{figure}[b]
    \centering
    \includegraphics[width=0.98\textwidth]{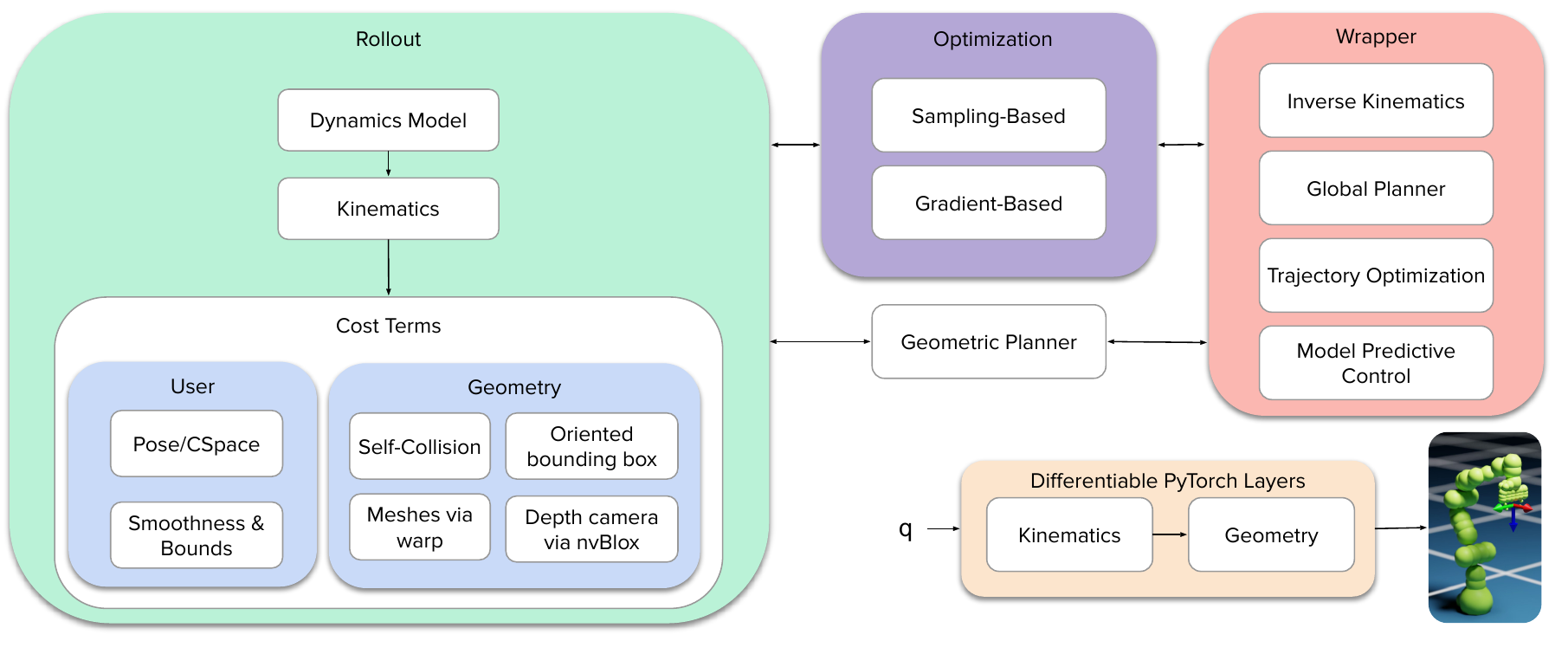}
    \caption{We illustrate the design of modules in our GPU accelerated library~\emph{cuRobo}. We have two core modules, a~\emph{Rollout} module that computes the cost given actions, and an~\emph{Optimization} module that has solvers to run numerical optimization. In addition, we have a~\emph{Wrapper} module that provides an high-level API to motion generation. We also provide differentiable PyTorch layers for kinematics and collision checking for use in neural networks.}
    \label{fig:curobo_library_design}
\end{figure}
In addition to the above tools available in PyTorch, many existing GPU libraries also expose interface to PyTorch such as Warp~\cite{warp}, Kaolin~\cite{KaolinLibrary}, and PyPose~\cite{wang2023pypose}, making it easier for users to use these libraries with our library. We leverage NVIDIA Warp~\cite{warp} in our library to write simple user-defined cost terms such as squared l2-distance~Eq.~\ref{eq:cspace-cost}, our smoothness cost term~Eq.~\ref{eq:smooth_cost}, and our joint limit cost term~Eq.~\ref{eq:bound_cost}. An example warp kernel and it's interface with PyTorch is shown in Fig.~\ref{fig:warp_pytorch}. We also leverage Warp's geometry kernels to compute the collision cost term for obstacles represented as meshes. We also developed a PyTorch API to nvblox~\cite{nvblox} which we use to compute collision cost when obstacles come from a depth stream. We are releasing the PyTorch API to nvblox as a separate library and provide interfacing code for use in cuRobo.

Our library is designed as shown in Fig.~\ref{fig:curobo_library_design} with a \emph{Rollout} module for rolling out the dynamics and computing the cost terms, an \emph{Optimization} module which contains numerical solvers, a \emph{Geometry} module that contains signed distance functions, a \emph{Geometric Planner} module for graph-based planning, and a wrapper module that provides a high-level api to motion generation tasks. We also provide differentiable PyTorch layers for kinematics and collision costs, for potential applications inside a neural network and also as part of a loss function motivated by recent successes in SDF-based rewards for contact-based manipulation~\cite{tang2023industreal}.

\begin{figure}
\centering
\begin{tabular}{p{6.2cm}  p{5.7cm}}
\begin{lstlisting}[language=Python, linewidth=0.45\textwidth]
@wp.kernel
def forward_l2_warp(
    pos: wp.array(dtype=wp.float32),
    target: wp.array(dtype=wp.float32),
    target_idx: wp.array(dtype=wp.int32),
    weight: wp.array(dtype=wp.float32),
    out_cost: wp.array(dtype=wp.float32),
    out_grad_p: wp.array(dtype=wp.float32),
    batch_size: wp.int32,
    horizon: wp.int32,
    dof: wp.int32,
):
    tid = wp.tid()
    # initialize variables:
    b_id = wp.int32(0)
    h_id = wp.int32(0)
    d_id = wp.int32(0)
    b_addrs = wp.int32(0)
    target_id = wp.int32(0)
    w = wp.float32(0.0)
    c_p = wp.float32(0.0)
    target_p = wp.float32(0.0)
    g_p = wp.float32(0.0)
    c_total = wp.float32(0.0)
    # we launch batch * horizon * dof kernels
    b_id = tid / (horizon * dof)
    h_id = (tid - (b_id * horizon * dof)) / dof
    d_id = tid - (b_id * horizon * dof + h_id * dof)
    if b_id >= batch_size or h_id >= horizon or d_id >= dof:
        return
    # read weight
    w = weight[0]
    b_addrs = b_id * horizon * dof + h_id * dof + d_id
    # read buffers
    c_p = pos[b_addrs]
    target_id = target_idx[b_id]
    target_id = target_id * dof + d_id
    target_p = target[target_id]
    # calculate error
    c_p = c_p - target_p
    c_total = w * c_p
    g_p = 2.0 * w  
    # write cost
    out_cost[b_addrs] = c_total
    # write gradient
    out_grad_p[b_addrs] = g_p
\end{lstlisting}
 & 
\begin{lstlisting}[language=Python, linewidth=0.45\textwidth]
class SquaredL2DistFunction(torch.autograd.Function):
    @staticmethod
    def forward(
        ctx,
        pos,
        target,
        target_idx,
        weight,
        out_cost,
        out_gp,
    ):
        wp_device = wp.device_from_torch(pos.device)
        b, h, dof = pos.shape
        wp.launch(
            kernel=forward_l2_warp,
            dim=b * h * dof,
            inputs=[
                wp.from_torch(pos.view(-1), dtype=wp.float32),
                wp.from_torch(target.view(-1), dtype=wp.float32),
                wp.from_torch(target_idx.view(-1), dtype=wp.int32),
                wp.from_torch(weight, dtype=wp.float32),
                wp.from_torch(out_cost.view(-1), dtype=wp.float32),
                wp.from_torch(out_gp.view(-1), dtype=wp.float32),
                b,
                h,
                dof,
            ],
            device=wp_device,
            stream=wp.stream_from_torch(pos.device),
          )
          cost = torch.sum(out_cost, dim=-1)
        ctx.save_for_backward(out_gp)
        return cost
    @staticmethod
    def backward(ctx, grad_out_cost):
        (p_grad,) = ctx.saved_tensors
        p_g = None
        if ctx.needs_input_grad[0]:
            p_g = p_grad * grad_out_cost 
        return p_g, None, None, None, None, None, None
\end{lstlisting} \\
 (a) Warp kernel in Python
& (b) Launching kernel from PyTorch \\
\end{tabular}
\caption{An example of using NVIDIA's Warp language for writing CUDA kernel is shown in (a), followed by it's interface with PyTorch in (b). As shown in line 29 of (b), we execute the warp kernel in the same CUDA stream as PyTorch's current stream to enable capture of kernel operations across pyTorch, Warp, and also custom CUDA kernels. This code was tested on PyTorch 1.13, along with Warp 0.9.0.}
\label{fig:warp_pytorch}
\end{figure}

\begin{figure}
    \centering
    \includegraphics[trim=0 2cm 0 2cm,clip,width=0.99\textwidth]{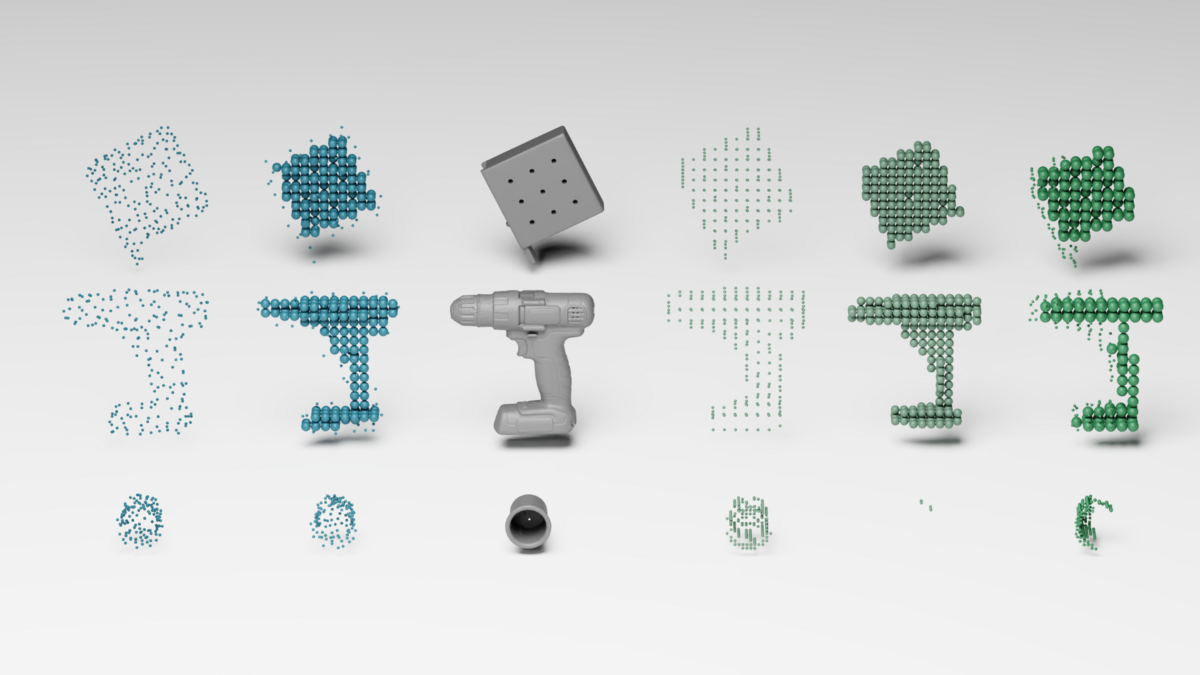}
    \caption{We illustrate some techniques to fit spheres to a mesh. We show three objects from the YCB dataset~\cite{calli2017yale} in the third column. The last three columns show different ways to get spheres from a voxelization of the mesh. The fourth column projects all occupied voxels to the surface and then takes these points as surface points, the fifth column only takes voxels that are within the surface of the mesh and uses the voxel pitch to compute the radius of the spheres. The last column combines fourth and fifth column spheres to get spheres that are in the volume and also the surface. However this representation does not cover the full geometry of the object, missing some key details on the surface as seen by \emph{cup} object in the last row. We hence only use the voxelization for computing spheres internal to the mesh and combine it with evenly sampled surface points (surface samples shown in the first column) to get a more accurate sphere representation in column 2. All approximations shown use 200 spheres per mesh.}
    \label{fig:sphere-fit}
\end{figure}

We have also developed interfacing code to read kinematic structures~(e.g., robots) from URDF and USD. To make environment configuration easier, we have developed a USD parser that can read obstacles and the robot directly from a USD stage. This USD parser enables our stack to work inside simulation engines such as NVIDIA Isaac sim without requiring extensive API integration as we can directly read the robot and the obstacles from the current USD stage. We provide an example in our code that shows how cuRobo can update it's obstacles based on an Isaac Sim world, with dynamic loading of new obstacles and obstacle pose changes. 

To obtain a sphere representation of the robot for collision cost terms, we leverage an existing utility from NVIDIA's Isaac sim (\href{https://docs.omniverse.nvidia.com/prod_digital-twins/app_isaacsim/advanced_tutorials/tutorial_motion_generation_robot_description_editor.html#adding-collision-spheres}{sphere-generator}). The sphere-generator gives a UI for manually adjusting the generated spheres to more accurately match the robot's geometry. In addition, we also develop a sphere approximation algorithm that can fit spheres to a mesh leveraging volume approximation techniques as shown in Fig.~\ref{fig:sphere-fit}. The first technique samples the surface of the mesh evenly. We also develop a second method that voxelizes the mesh and treats each occupied voxel as a sphere and combines this with surface sampled spheres. We currently use this technique only for approximating the geometry of objects that are attached to a gripper during manipulation to enable collision avoidance between a grasped object and the world. One limitation of this technique is that it works well only when the number of spheres is greater than 100. We do not use this automatic technique to approximate the robot's geometry as this can lead to requiring thousands of spheres, potentially slowing down our pipeline on low-power compute devices such as the NVIDIA Jetson ORIN. In addition, our current implementation of the self collision kernel is limited to 1024 spheres as we leverage warp-wide primitives to find the largest penetration distance.
\section{Parallelized Compute Kernels}
\label{app:cuda_kernels}
GPUs are specialized for highly parallel computations as they have more transistors devoted to data processing compared to caching and flow control. This enables GPUs to hide memory access latencies in compute bottle-necked scenarios by running more parallel compute. A GPU has significantly more instruction throughput and memory bandwidth that a CPU within a similar power envelope and cost. To efficiently leverage GPU compute for motion generation, we found the following to be important:
\begin{enumerate}
    \item Reducing reads and writes to global memory to avoid hitting memory access latency. 
    \item Skip writing zeros for gradients by keeping track of sparsity. As the optimization converges, most gradients will go to zero and skipping rewrites of zeros greatly reduces the memory access bottlenecks in our cost terms. 
    \item Reducing number of kernels called in an optimization iteration by combining small kernels into a single larger kernel that does all the work from the small kernels.
    \item Using shared memory to share data across a thread block, enabling for-loops to be run in parallel.
    \item Leveraging warp-wide operations for computing values across small thread groups such as reductions and finding the maximum.
\end{enumerate}
We discuss the implementation of some of the key kernels in our framework in the following sections, starting with our kinematics kernels in Section~\ref{app:kinematics-cuda}, followed by our self-collision kernel in Section~\ref{app:self-collision-cuda}, our continuous signed distance kernel in Section~\ref{app:signed_distance_cuda}, and then our L-BFGS kernel in Section~\ref{sec:lbfgskernel}. We then report compute times for these kernels across compute devices in Section~\ref{app:kernel-timing}. There are many more kernels in our library and urge the readers to look at our code for the full set.
\subsection{Kinematics}
\label{app:kinematics-cuda}
The role of the forward kinematics function is to map a robot's joint configuration to the pose of the robot's geometry in world coordinates. In cuRobo, we represent the robot's geometry by spheres and also provide task space poses for any links thats required in computing costs (e.g , the pose of the end-effector for motion generation) as shown in Figure~\ref{fig:app_robot_sphere}. To perform gradient-based optimization, we require the backward mapping to project gradients from the world coordinates for poses and spheres, to the joint configuration which we call backward kinematics (i.e., this is also called as the Kinematic Jacobian). There are many ways to represent a robot's kinematics~\cite{dual_quaternion}. We wanted a representation that not only has less number of operations but also allows for running many operations in parallel. We found representing the transformations as 4$\times$4 homogeneous transformation matrices to be the best representation as it enables us to run 4 parallel threads to compute matrix multiplications using shared memory.

The work for forward kinematics consists of computing a 4$\times$4 matrix (which we call $cumul$) per link, which is then used to compute robot sphere locations for each of the links. We distribute the work per batch across a number of threads such that the compute resources are well utilized for the matrix multiplications and sphere transformations. We use four threads per batch in our implementation as shown in Algorithm~\ref{alg:fk}. 

For backward kinematics, the $cumul$ matrix is used to compute the gradients for each of the spheres. Similar to the above implementation, we distribute the work to multiple threads (16 threads/batch) as shown in Algorithm~\ref{alg:backward-kinematics}. The $cumul$ matrix can either be saved to memory in forward and reused by the backward kernel, or regenerated by the backward. Reusing the matrix may be slower than regenerating it for large batch sizes on systems with low memory bandwidth. We hence use a flag to choose between reusing from memory and recomputing for this kernel. We also write out the transformation matrices that map a joint value to a matrix in Table~\ref{tab:joint-transform} and their gradients in Table~\ref{tab:gradient-transform}.

\subsection{Self-Collision}
\label{app:self-collision-cuda}
Our CUDA kernel checks for self-collision for a batch of robot configurations by reading the location of the spheres in the joint configuration along with the radii. Per batch, we compute distances between all pairs in the set of spheres~$S$ that represent the robot and find the pair with the minimum distance. 

We map the work per batch to one thread-block so that the parallel reduction can be performed relatively quickly using intra-warp sharing primitives and shared memory, avoiding atomic accesses altogether. For distance computations we evaluated two versions. First version maps 32$\times$32 distance computations to a warp such that the memory accesses are coalesced and we reuse the first loaded sphere for 32 distance computations. To avoid extra work, this version checks that the first sphere's index is greater than second sphere's index. In doing so, some of the fetches get wasted. The second version, assigns a set of distance computations (set size being 8, 16, 32, etc) per thread. The assignment consists of sphere index pairs and is pre-computed. Two spheres are fetched by index per computation. To reduce the fetch overhead, we store the spheres in shared memory. We use this version in our evaluations and is written in Algorithm~\ref{alg:self-collision}.
\subsection{Signed Distance}
\label{app:signed_distance_cuda}
\begin{figure}
    \centering
    \begin{tabular}{c | c}
             \includegraphics[height=15cm]{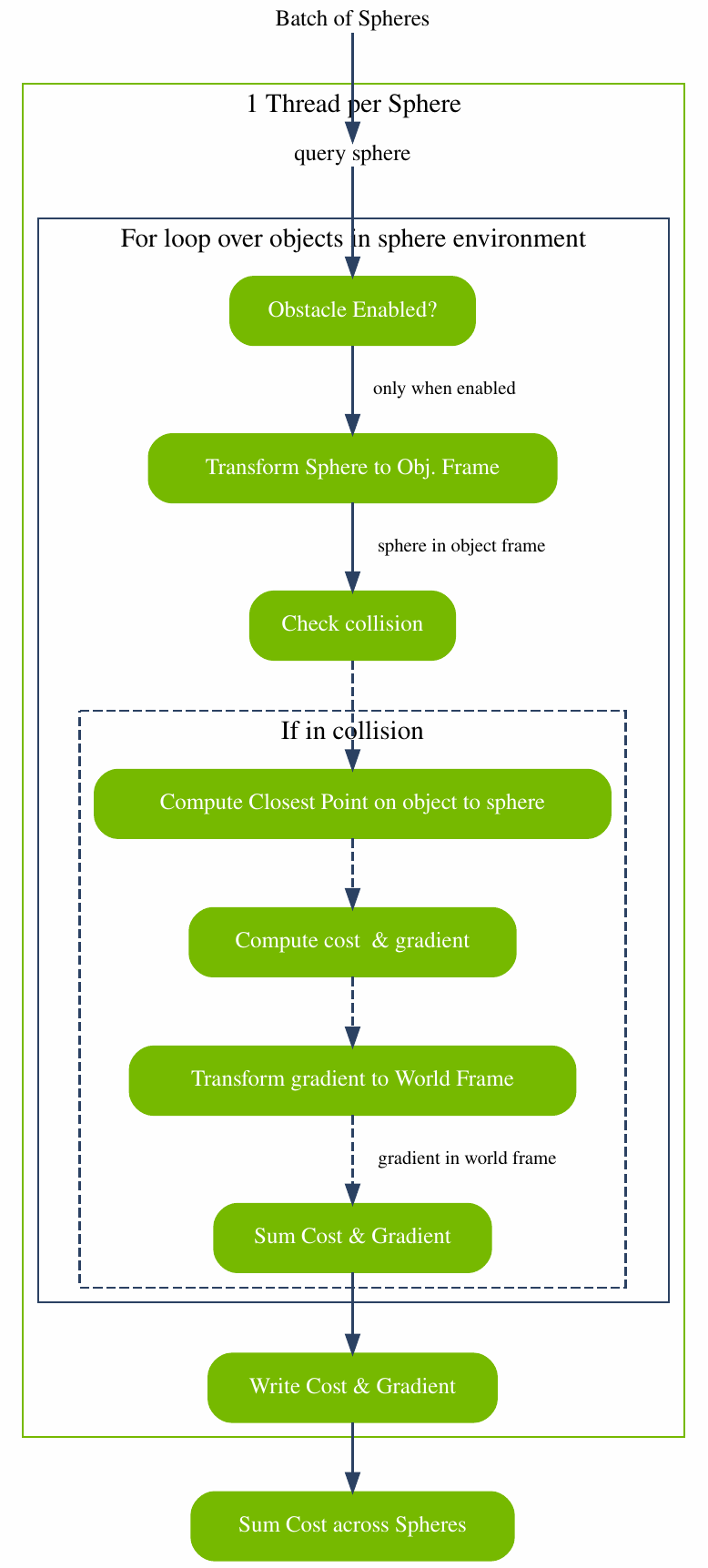}
&      \includegraphics[height=15cm]{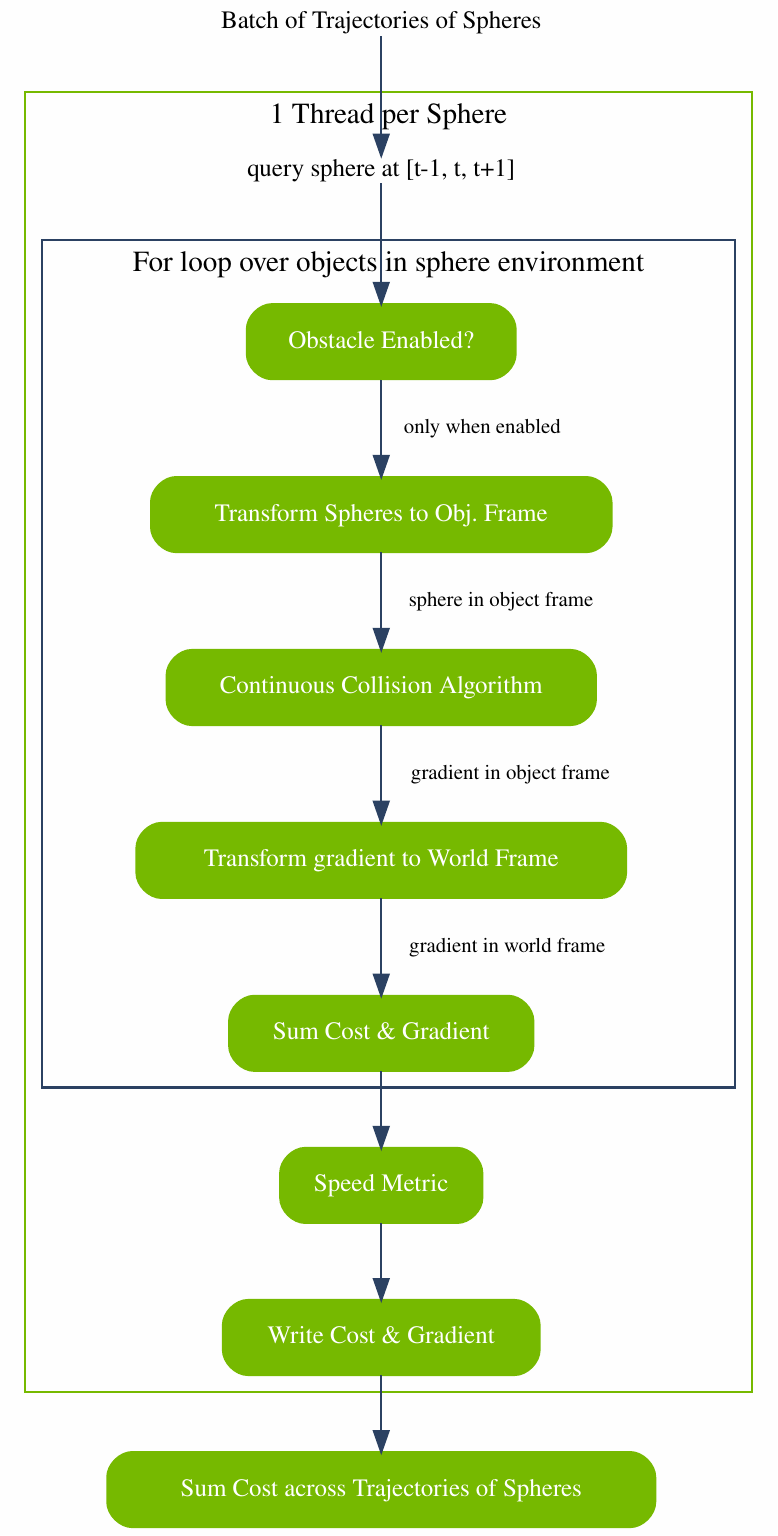}
\\
(a) Sphere-World Signed Distance     & (b) Sphere-World Continuous Signed Distance
    \end{tabular}
    \caption{The functions to compute signed distance and gradient between a query sphere and objects in the world is shown in (a). In (b), a continuous signed distance algorithm is shown for computing signed distance across a trajectory taken by a query sphere.}
    \label{fig:collision_algorithm_flowchart}
\end{figure}

We distribute the collision checking work per batch across a number of threads equal to the number of spheres and horizons. Each thread loads a sphere along with two others from adjacent horizons. For each of the OBBs, the distance between the sphere and OBB is computed by first transforming the sphere to OBB coordinate frame and then checking if the sphere is within bounds as it becomes an axis-aligned bounding box~(AABB). If there's a potential collision, gradient is computed. If not, we check whether a potential collision is possible between two adjacent time horizons to compute gradients based on the continuous collision checking algorithm described in Section~\ref{sec:collision-avoidance}. For inverse kinematics and graph planning, we only check collisions at discrete times which is written in Algorithm~\ref{alg:collision-kernel}. The continuous collision kernel is written in Algorithm~\ref{alg:continuous-collision} followed by the steps in computing continuous collision distance in Algorithm~\ref{alg:continuous-collision-distance}. The flow of the algorithms are also illustrated in Figure~\ref{fig:collision_algorithm_flowchart}.

\subsection{L-BFGS}
\label{sec:lbfgskernel}
L-BFGS consists of a series of dot products to compute the step direction from the history of gradients and optimization parameters. We use many threads in one thread-block to perform the product and reduce (sum-up) the products in parallel. This will work as long as the history is less than 16. The increase in run-time with batch size will follow a step function as GPUs include multiple SMs that can execute thread-blocks simultaneously (e.g., Jetson Orin AGX includes 16 SMs in 60W mode). Another compute component in L-BFGS is shifting buffer elements, which when done in parallel and asynchronously requires storing the values in a temporary buffer and then updating the buffer to avoid data corruption. In our implementation, the buffer size was $\le$32, typical size of a warp in a GPU. Based on this fact and the GPU's SIMT execution model, we leverage warp shuffle operation to shift values down, avoiding explicit memory loads/stores for temporary values~\cite{CUDAProgrammingGuide}. This logic can also be extended to shift larger buffers by taking care of the elements at the warp boundaries.

\subsection{Kernel Timing Benchmark}
\label{app:kernel-timing}
We profile trajectory optimization for one problem in our evaluation and write the timings for all kernels in an iteration of the L-BFGS solver in Table~\ref{tab:kernel-parameters-10}. Our trajectory optimization first runs 100 iterations with 12 parallel seeds, selects a good time-step discretization value and optimizes 1 seed for upto 300 seeds. We plot the time it takes for running 1 seed in Table~\ref{tab:kernel-parameters}.

We compare the kernel time for computing the world collision cost between three different world representations. First, we represent the world as oriented bounding boxes (OBB) and use our custom CUDA kernel to compute the collision cost. Second, we represent the world with meshes and use Warp's BVH based signed distance function to compute the collision cost. Third, we render depth images of the same world and build an ESDF map using nvblox and compute collision cost with this map using our custom nvblox CUDA kernel. We compute times for both 12 TO seeds and 1 TO seed in Table~\ref{tab:coll-kernel-time}. We did not implement the continuous algorithm for the nvblox kernel and only compute collision cost at the sphere waypoint. We compute the speed metric for all kernels.

From the compute times, we observe that cuboid collision cost~\emph{OBB-Collision} kernel is much faster~(8$\times$) than using meshes for collision cost. Using a mesh representation also uses more registers~(207 vs 77). The slowdown with the use of meshes was even more significant on the Jetson which has lower number of SMs and lower memory bandwidth. Our nvblox kernel takes less time than even the cuboid collision kernel on a RTX 4090. We do not implement the continuous collision algorithm for nvblox, which makes the comparison weak as the cuboid and mesh kernels are doing more work than nvblox. It's still promising see that collision cost computation with nvblox does not add significant overhead compared to other world representations. We leave implementation of the continuous collision algorithm in nvblox for a future work.

\subsection{Profiling CUDA Flags}
\label{sec:cuda-profile}
PyTorch exposes optional CUDA optimizations, some require recompilation of the CUDA kernels such as \texttt{fast-math} while others only require enabling a flag at runtime such as using \texttt{tf32} for arithmetic operations. In addition, pyTorch uses it's own memory allocator by default. We compared the solve time on our 2600 problems and found no significant difference between \texttt{cudaMallocAsync} memory allocator and pyTorch's default allocator. We found that not using \texttt{tf32} made our approach 15\% slower and not using \texttt{fast-math} made our approach 18\% slower. In addition to flags, PyTorch also exposes CUDAGraphs which enables capturing a sequence of kernel launches on a CUDA device and replaying this captured graph for repeated calls. This fits very well to numerical optimization, where we call a sequence of functions to map the optimization variables to cost, followed by computation of step direction and a better set of optimization variables at every iteration. We found that capturing 25 iterations of our solver in a CUDAGraph and replaying this capture gave us a 10$\times$ speedup compared to calling these kernels native.

\begin{table} %
    \centering
    \begin{tabular}{l r r r r >{\columncolor{bargreenlight}}r >{\columncolor{bargreenlight2}}r  >{\columncolor{bargreenlight3}}r }
    \toprule
    \multirow{2}{*}{\textbf{ Kernel }} & \multirow{2}{*}{\textbf{grid}} & \multirow{2}{*}{\textbf{block}} &\multirow{2}{*}{\textbf{registers}} &\multirow{2}{*}{\textbf{sh. mem.}} & \multicolumn{3}{c}{\textbf{Time ($\mu$s)}} \\ 
    &  & & &  & \textbf{ 4090} & \textbf{Orin MAXN} & \textbf{Orin 15W} \\
    \toprule
    Tensor Step-FW & 84 & 128 & 30 [36] & 0 & 2 & 13 & 26 \\
    Tensor Step-BW & 84 & 128 & 38 & 0 & 1 & 4 & 6 \\
    \midrule 
    Kinematics-FW & 48 & 128 & 33 [34]  & 28672 & 10 & 44 & 112 \\
    Kinematics-BW & 192 & 128 & 39 [40] & 7168 & 12 & 25 & 34 \\
    \midrule
    Self-Collision & 1536 & 400 & 33 [36] & 1280 & 7 & 77 & 208 \\ 
    OBB-Collision & 768 & 128 & 77 [78] & 0 & 8 & 64 & 157 \\
    \midrule
    Pose-Cost & 12 & 128 & 61 [58] & 0 & 2 & 5 & 6 \\  
    Bound-Smooth & 42 & 256 & 54 & 0 & 2 & 6 & 9 \\ 
    \midrule
    L-BFGS Step & 12 & 224 & 48 &  3728 & 4 & 11 & 13 \\ 
    Line Search & 12 & 224 & 42 &  260 & 3 & 8 & 9 \\
    Update Best & 21 & 128 & 16 &  0 & 1 & 3 & 4 \\
    \midrule
    Reduce (4) & &  &  & 0 & 7  & 21   & 28   \\ 
    Elementwise (3) &  &  &  & 0 & 4 & 28  &  41\\ 
    Concatenate & 
    &  512 & 21 [18] & 0 & 2 & 4 & 6 \\ 
    Jit  & 21 & 128 & 16 & 0 & 1 & 3 & 3 \\ 
    \midrule
    \textbf{All (20 kernels)} &  & &  &   & \textbf{66} &\textbf{316} & \textbf{662} \\
    \bottomrule
    \end{tabular}
    
    \caption{Time taken by kernels in 1 iteration of Trajectory Optimization(12 seeds, 32 timesteps).}
    \label{tab:kernel-parameters-10}
    
\end{table}
\begin{table} %
    \centering
    \begin{tabular}{l r r r r >{\columncolor{bargreenlight}}r >{\columncolor{bargreenlight2}}r  >{\columncolor{bargreenlight3}}r }
    \toprule
    \multirow{2}{*}{\textbf{ Kernel }} & \multirow{2}{*}{\textbf{grid}} & \multirow{2}{*}{\textbf{block}} &\multirow{2}{*}{\textbf{registers}} &\multirow{2}{*}{\textbf{sh. mem.}} & \multicolumn{3}{c}{\textbf{Time ($\mu$s)}} \\ 
    &  & & &  & \textbf{ 4090} & \textbf{Orin MAXN} & \textbf{Orin 15W} \\
    \toprule
    Tensor Step-FW & 7 & 128 & 30 [36] & 0 & 2 & 5 & 6 \\
    Tensor Step-BW & 7 & 128 & 38 & 0 & 1 & 2 & 2 \\
    \midrule 
    Kinematics-FW & 4 & 128 & 33 [34]  & 28672 & 9 & 16 & 17 \\
    Kinematics-BW & 16 & 128 & 39 [40] & 7168 & 4 & 7 & 7 \\
    \midrule
    Self-Collision & 128 & 400 & 33 [36] & 1280 & 2 & 9 & 20 \\ 
    OBB-Collision & 64 & 128 & 77 [78] & 0 & 4 & 10 & 19 \\
    \midrule
    Pose-Cost & 1 & 128 & 61 [58] & 0 & 2 & 3 & 4 \\  
    Bound-Smooth & 4 & 256 & 54 & 0 & 2 & 4 & 5 \\ 
    \midrule
    L-BFGS Step & 1 & 224 & 48 &  3728 & 4 & 7 & 7 \\ 
    Line Search & 1 & 224 & 42 &  260 & 3 & 5 & 5 \\
    Update Best & 2 & 128 & 16 &  0 & 1 & 2 & 2 \\
    \midrule
    Reduce (4) & &  &  & 0 & 7  & 13 &  15 \\ 
    Elementwise (3) &  &  &  & 0 & 3 & 6 & 7\\ 
    Concatenate & 
    &  512 & 21 [18] & 0 & 1 & 3 & 3 \\ 
    Jit  & 7 [2, 4] & 128 & 16 & 0 & 1 & 2 & 2 \\ 
    \midrule
    \textbf{All (20 kernels)} &  & &  &   & \textbf{46} &\textbf{94} & \textbf{121} \\
    \bottomrule
    \end{tabular}
    \caption{Time taken by kernels in 1 iteration of Trajectory Optimization(1 seed, 32 timesteps).}
    \label{tab:kernel-parameters}
    
\end{table}

\begin{table}
    \centering
    \begin{tabular}{l r r r >{\columncolor{bargreenlight}}r >{\columncolor{bargreenlight2}}r  >{\columncolor{bargreenlight3}}r}
    \toprule
    \multirow{2}{*}{\textbf{ Kernel }} & \multirow{2}{*}{\textbf{grid}} & \multirow{2}{*}{\textbf{block}} &\multirow{2}{*}{\textbf{registers}} & \multicolumn{3}{c}{\textbf{Time ($\mu$s)}} \\ 
    &  & &  & \textbf{ 4090} & \textbf{Orin MAXN} & \textbf{Orin 15W} \\
    \toprule
    OBB-Collision (1-TO) & 64 & 128 & 77 [78]  & 4 & 10 & 19 \\
    Mesh-Collision (1-TO) & 68 & 256 & 207 [210]  & 27 & 115 & 260 \\
    nvblox-Collision* (1-TO) & 64 & 128 & 56 &  2 & - & - \\
    \midrule
    OBB-Collision (12-TO) & 768 & 128 & 77 [78] & 8 & 64 & 157 \\
    Mesh-Collision (12-TO) & 340 & 256 & 207 [210] & 37 & 353 & 874 \\
    nvblox-Collision* (12-TO) & 768 & 128 & 56 &  5 & - & - \\ \bottomrule    
    \end{tabular}
    \caption{World Collision Checking Representation.}
    \label{tab:coll-kernel-time}
\end{table}

\begin{table*}
\centering
\begin{tabular}{l r r}
\toprule
\textbf{Joint Type} & \textbf{Joint Transformation} & \textbf{Full Link Transformation} \\ \toprule
Fixed Joint & $\begin{bmatrix}
    1 & 0 & 0 & 0\\
    0 & 1 & 0 & 0\\
    0 & 0 & 1 & 0\\
    0 & 0 & 0 & 1
         \end{bmatrix}$ & $\begin{bmatrix}
    f_0 & f_1 & f_2 & f_3\\
    f_4 & f_5 & f_6 & f_7\\
    f_8 & f_9 & f_{10} & f_{11}\\
    0 & 0 & 0 & 1
         \end{bmatrix}$ \\ \midrule
Prismatic X & $\begin{bmatrix}
    1 & 0 & 0 & d_x \\
    0 & 1 & 0 & 0\\
    0 & 0 & 1 & 0\\
    0 & 0 & 0 & 1
    \end{bmatrix}$ & $\begin{bmatrix}
        f_0 & f_1 & f_2 & f_0  d_x + f_3\\
        f_4 & f_5 & f_6 & f_4  d_x + f_7 \\
        f_8 & f_9 & f_{10} & f_8  d_x + f_{11} \\ 
        0 & 0 & 0 & 1 \\
    \end{bmatrix}$ \\ \midrule
Prismatic Y & $\begin{bmatrix}
    1 & 0 & 0 & 0 \\
    0 & 1 & 0 & d_y\\
    0 & 0 & 1 & 0\\
    0 & 0 & 0 & 1
    \end{bmatrix}$ 
    & $\begin{bmatrix}
        f_0 & f_1 & f_2 & f_1  d_y + f_3\\
        f_4 & f_5 & f_6 & f_5  d_y + f_7 \\
        f_8 & f_9 & f_{10} & f_9  d_y + f_{11} \\ 
        0 & 0 & 0 & 1 \\
    \end{bmatrix}$\\ \midrule
Prismatic Z & $\begin{bmatrix}
    1 & 0 & 0 & 0 \\
    0 & 1 & 0 & 0\\
    0 & 0 & 1 & d_z\\
    0 & 0 & 0 & 1
    \end{bmatrix}$ & 
    $ \begin{bmatrix}
    f_0 & f_1 & f_2 & f_2  d_z + f_3\\
    f_4 & f_5 & f_6 & f_6  d_z + f_7 \\
    f_8 & f_9 & f_{10} & f_{10}  d_z + f_{11} \\ 
    0 & 0 & 0 & 1 
    \end{bmatrix}$\\ \midrule
Revolute X & $\begin{bmatrix}
    1 & 0 & 0 & 0 \\
    0 & \cos \theta_x & -\sin \theta_x & 0\\
    0 & \sin \theta_x & \cos \theta_x & 0\\
    0 & 0 & 0 & 1
    \end{bmatrix}$ & $\begin{bmatrix}
        f_0 & f_1 \cos \theta_x + f_2 \sin \theta_x & -f_1 \sin \theta_x + f_2 \cos \theta_x & f_3 \\
        f_4 & f_5 \cos \theta_x + f_6 \sin \theta_x & -f_5 \sin \theta_x + f_6 \cos \theta_x &
        f_7 \\
        f_8 & f_9 \cos \theta_x + f_{10} \sin \theta_x & -f_9 \sin \theta_x + f_{10} \sin \theta_x & f_{11}\\
        0 & 0 & 0 & 1 \\
    \end{bmatrix}$ \\ \midrule
Revolute Y & $\begin{bmatrix}
    \cos \theta_y & 0 & \sin \theta_y & 0\\
    0 & 1 & 0 & 0 \\
    -\sin \theta_y & 0 & \cos \theta_y & 0\\
    0 & 0 & 0 & 1
    \end{bmatrix}$ & $\begin{bmatrix}
        f_0 \cos \theta_y - f_2 \sin \theta_y  & f_1 & f_0 \sin \theta_y + f_2 \cos \theta_y & f_3\\
        f_4 \cos \theta_y - f_6 \sin \theta_y  & f_5 & f_4 \sin \theta_y + f_6 \cos \theta_y & f_7\\
        f_8 \cos \theta_y - f_{10} \sin \theta_y  & f_9 & f_8 \sin \theta_y + f_{10} \cos \theta_y & f_{11}\\
        0 & 0 & 0 & 1 \\
    \end{bmatrix}$ \\ \midrule
Revolute Z & $\begin{bmatrix}
    \cos \theta_z & -\sin \theta_z & 0 & 0\\
    \sin \theta_z & \cos \theta_z & 0 & 0\\
    0 & 0 & 1 & 0 \\
    0 & 0 & 0 & 1
    \end{bmatrix}$& 
    $ \begin{bmatrix}
        f_0 \cos \theta_z + f_1 \sin \theta_z & -f_0 \sin \theta_z + f_1 \cos \theta_z & f_2 & f_3 \\
        f_4 \cos \theta_z + f_5 \sin \theta_z & -f_4 \sin \theta_z + f_5 \cos \theta_z & f_6 & f_7 \\
        f_8 \cos \theta_z + f_9 \sin \theta_z & -f_8 \sin \theta_z + f_9 \cos \theta_z & f_{10} & f_{11} \\
        0 & 0 & 0 & 1 \\
    \end{bmatrix}$\\ \bottomrule
\end{tabular}
\caption{Joint Transformation matrices based on axis of actuation.}
\label{tab:joint-transform}
\end{table*}

\begin{table}
    \centering
\begin{tabular}{l r r}
    \toprule
    \textbf{Joint Type} & \textbf{Position Gradient} & \textbf{Rotation Gradient} \\ \toprule
    Prismatic X & $\vec{g}_d^\top (\vec{x})$ & 0 \\ \midrule
    Prismatic Y & $\vec{g}_d^\top (\vec{y})$ & 0 \\ \midrule
    Prismatic Z & $\vec{g}_d^\top (\vec{z})$ & 0 \\ \midrule
    Revolute X & $\vec{g}_d^\top (\vec{x} \times (\vec{l} - \vec{d}))$  & $\vec{g}_r^\top \langle[\vec{x}]_{\times} R_p\rangle$ \\ \midrule
    Revolute Y & $\vec{g}_d^\top (\vec{y} \times (\vec{l} - \vec{d}))$  & $\vec{g}_r^\top \langle[\vec{y}]_{\times} R_p\rangle$ \\ \midrule
    Revolute Z & $\vec{g}_d^\top (\vec{z} \times (\vec{l} - \vec{d}))$  & $\vec{g}_r^\top \langle[\vec{z}]_{\times} R_p\rangle$ \\
    \toprule
\end{tabular}
\caption{Gradient for different joint types, where $\langle \cdot \rangle$ refers to flattening a matrix to a vector.}
\label{tab:gradient-transform}
\end{table}

\begin{algorithm}
\DontPrintSemicolon
\algokw
\caption{Forward Kinematics using 4 threads}\label{alg:fk}
\SetKwInput{Input}{Input}
\SetKwInput{Output}{Output}
\SetKwInput{KinData}{Kinematics Data}
\SetKwInput{Kernel}{Kernel Launch Data}
\Kernel{Launch 4 threads per batch, tid = thread id}
\Input{q$\in \mathbb{R}^{B\times D}$}
\Output{link\_pos$\in \mathbb{R}^{B \times N \times 3}$, link\_quat$\in \mathbb{R}^{B \times N \times 4}$, b\_robot\_spheres$\in \mathbb{R}^{B \times M \times 4}$, b\_cumul\_mat$\in \mathbb{R}^{B \times N \times 16}$}
\KinData{fixedTransform$\in \mathbb{R}^{L\times4\times 4}$, 
robotSpheres$\in \mathbb{R}^{M\times4}$,
linkMap$\in \mathbb{Z}^{L}$,
jointMap$\in \mathbb{Z}^{L}$,
jointMapType$\in \mathbb{Z}^{L}$,
storeLinkMap$\in \mathbb{Z}^{N}$,
linkSphereMap$\in \mathbb{Z}^M$,
B,M,N,L,D}
\tcc{B = batch size, M = number of spheres, N = number of links to write, L = number of links, D = number of actuated joints}
extern shared cumul\_mat \tcp*{store cumul matrix in shared memory $\in \mathbb{R}^{bpb \times l \times 16}$ (bpb= batches per block)}
col\_idx = tid \% 4  \tcp*{Using four threads per batch index}
b\_idx = tid / 4 \tcp*{batch index for current thread}
m\_base = b\_idx * L * 16 \tcp*{matrix index}
cumul\_mat[m\_base + col\_idx * 4] = fixedTransform[col\_idx * 4] \tcp*{read a column of the base link matrix}
\uIf{write\_cumul}{
 b\_cumul\_mat[b\_idx * L * 16 + col\_idx * 4] = cumul\_mat[m\_base + col\_idx * 4] \;
}
\For(\tcp*[f]{loop over links}){ l$\gets 1$\KwTo L}{
ft\_base = l * 16\;
in\_base = m\_base + linkMap[l] * 16\;
out\_base = m\_base + l * 16 \;
j\_type = jointMapType[l]\tcp*{read joint type}
angle = q[b\_idx * D * jointMap[l]] \tcp*{joint articulation value}
\tcc{compute local transformation matrix from articulation value using table~6}
j\_col = joint\_transform(j\_type, angle, fixedTransform[ft\_base + col\_idx], col\_idx)\;

\For(\tcp*[f]{multiply local transform with previous link transform to get global transform}){i $\gets 0$ \KwTo 3}{ 
cumul\_mat[out\_base + i * 4 + col\_idx] = dot(cumul\_mat[in\_base + i * 4], j\_col)\;
}

\uIf(\tcp*[f]{write out transforms for use in backward}){write\_cumul}{
b\_cumul\_mat[b\_idx * L * 16 + l * 16 + col\_idx * 4] = cumul\_mat[out\_base + col\_idx*4]\;
}
}
\tcc{compute sphere positions and write to memory}
mpt = (M + 3) / 4 \tcp*{spheres per thread in a batch index}
\For{i $\gets 0$ \KwTo mpt}{
m\_idx = i * 4 + col\_idx \;
\uIf{m\_idx $\geq$ M}
{
break\;
} 
read\_cumul\_idx = linkSphereMap[m\_idx] \tcp*{read link index for sphere}
transform\_sphere(robotSpheres[sph\_idx * 4], cumul\_mat[mat\_base + read\_cumul\_idx * 16], b\_robot\_spheres[b\_idx + m\_idx*4]) \;
}

\tcc{write link poses to memory}
\For{i $\gets 0$ \KwTo N}{
l\_map = storeLinkMap[i]\;
l\_base = b\_idx * N\;
out\_mbase = m\_base + l\_map * 16\;
quat = mat\_to\_quat(cumul\_mat[out\_mbase])\;

link\_quat[l\_base * 4 + i *4 + col\_idx] = quat[col\_idx] \tcp*{write one value per thread}

\uIf{col\_idx$<$3}
{
link\_pos[l\_base * 3 + i* 3 + col\_idx] = cumul\_mat[out\_mbase]\;
}
}
\end{algorithm}

\begin{algorithm}
\DontPrintSemicolon
\algokw
\caption{Backward Kinematics using 16 threads}\label{alg:backward-kinematics}
\SetKwInput{Input}{Input}
\SetKwInput{Output}{Output}
\SetKwInput{KinData}{Kinematics Data}
\SetKwInput{Kernel}{Kernel Launch Data}
\Kernel{Launch 16 threads per batch}
\Input{grad\_link\_pos, grad\_link\_quat, grad\_spheres, global\_cumul\_mat}
\KinData{
robotSpheres$\in \mathbb{R}^{M\times4}$,
linkMap$\in \mathbb{Z}^{L}$,
jointMap$\in \mathbb{Z}^{L}$,
jointMapType$\in \mathbb{Z}^{L}$,
storeLinkMap$\in \mathbb{Z}^{N}$,
linkSphereMap$\in \mathbb{Z}^M$,
B,M,N,L,D}
\Output{grad\_out\_q}
extern shared cumul\_mat\;
b\_idx = tid / 16\;
elem\_idx = tid \% 16 \;
\For(\tcp*[f]{read global cumul matrix to shared memory}){l $\gets$0 \KwTo N}{
cumul\_mat[e\_idx] = global\_cumul\_mat[b\_idx * L * 16 + l * 16 + elem\_idx] \;
}
psum\_grad = []\;
mpt = (M+15)/16 \;
\For(\tcp*[f]{project sphere gradients to joints}){i in M}{
m\_idx = elem\_idx * mpt + i\;
\uIf {m\_idx $\geq$ M}{
break\;
}
loc\_grad\_sph = grad\_spheres[(b\_idx * M + m\_idx) * 4]\; 
\uIf{loc\_grad\_sph == 0}
{
continue \;
}
read\_cumul\_idx = linkSphereMap[m\_idx] \;
sphere\_mem = transform\_sphere(robotSpheres[m * 4], cumul\_mat) \;
\tcc{assuming all joints affecting current sphere has an index below read\_cumul\_idx}
\For{j in [read\_cumul\_idx, 0]}{
\uIf{ \upshape linkChainMap[read\_cumul\_idx, j] == 0}{
continue \tcp*{skip links that are not in current serial chain}
}
j\_type = jointMapType[j]\;
psum\_grad[jointMap[j]] += point\_backward\_grad(cumul\_mat[j * 16], sphere\_mem, loc\_grad\_sph, j\_type)\;
}
}
\For(\tcp*[f]{project link gradients to joints}){i $\gets 0$ \KwTo N}{

g\_pos = grad\_link\_pos[(b\_idx + i)*3]\;

g\_quat = grad\_link\_quat[(b\_idx + i)*4]\;

\uIf{g\_pos == 0 \&\& g\_quat == 0}{
continue \tcp*{skip computation when gradients are zero}
}
l\_map = storeLinkMap[i]\;
l\_base = b\_idx * N\;
out\_mbase = m\_base + l\_map * 16\;
dpt = (l\_map + 15) / 16 \tcp*{number of links per thread}
\For{k in [dpt, 0]}{
j = k * 4 + elem\_idx\;
\uIf{\upshape j $>$ l\_map or j $< $ 0 or linkChainMap[l\_map, j] == 0}{
continue \tcp*{skip links that are not in current serial chain}
}
j\_idx = jointMap[j]\;
j\_type = jointMapType[j]\;
\tcc{compute gradient using table 7}
psum\_grad[j\_idx] += pose\_backward\_grad(cumul\_mat[], l\_pos, g\_pos, g\_quat)\;
}
}
psum\_grad = warpReduce(psum\_grad) \tcp*{sum gradient across warp}
grad\_out\_q[b\_idx * D] = psum\_grad \tcp*{write out gradients in a for loop on thread 0}
\end{algorithm}

\begin{algorithm}
\DontPrintSemicolon
\algokw
\caption{Robot Self-Collision Checking using sphere-pair threads}\label{alg:self-collision}
\SetKwInput{Input}{Input}
\SetKwInput{Output}{Output}
\SetKwInput{KinData}{Kinematics Data}
\SetKwInput{Kernel}{Kernel Launch Data}
\SetKwInput{Data}{Data}
\Input{b\_robot\_spheres$\in \mathbb{R}^{b \times M \times 4}$}
\Output{out\_distance$\in \mathbb{R}^{b}$, out\_grad$\in \mathbb{R}^{b\times M \times 4}$}
\Data{sparse\_index$\in \mathbb{B}^{b \times M}$}
\KinData{offsets$\in \mathbb{M}$, weight$\in \mathbb{R}$, locations$\in \mathbb{Z}^{M \times M}$}
\Kernel{NDPT, NBPB, B, M, Launch  (max\_pairs + NDPT )/NDPT threads}
b\_idx = block\_id * NBPB\;
nbpb = min(NBPB, B - b\_idx)\;
\uIf{nbpb == 0}{
return \;
}
extern shared rs\_shared\;
\uIf{tid$<$M}
{
\For{l$\in [0, nbpb]$}{
sph = b\_robot\_spheres[4 * ((b\_idx + l) * M + tid)] \tcp*{read sphere from global memory}
sph[3] += offsets[tid] \tcp*{add offset to sphere radius}
rs\_shared[NBPB * tid + l] = sph \tcp*{copy sphere to shared memory}
}
}
sync\_threads()\;

indices[NDPT * 2] \;
\For(\tcp*[f]{read indices of sphere pairs for this thread}){i $\in [0, NDPT*2]$}{
indices[i]  = locations[tid * 2 * NDPT + i]\;
}
max\_d = \{d:0, i:-1, j:-1\} \;
\For(\tcp*[f]{compute sphere pair distances and store the largest in this thread}
){\forloop{k}{0}{NDPT}}{
i = indices[k*2] \;
j = indices[k * 2 + 1] \;
\For{\forloop{l}{0}{nbpb}}{
sph1 = rs\_shared[NBPB * i + l]\;
sph2 = rs\_shared[NBPB * j + l]\;
\uIf{\upshape sph1.radius $\leq$ 0.0 or sph2.radius $\leq$ 0.0}
{
continue\;
}
dist = sphere\_distance(sph1, sph2)\;
\uIf{\upshape dist $>$ max\_d.d}
{
max\_d.d = dist\;
max\_d.i = i\;
max\_d.j = j\;
}
}
}
w\_max\_d = WarpMax(max\_d) \tcp*{Find the largest distance across threads in warp}

\uIf{tid$<$ M}{
\For{\forloop{l}{0}{nbpb}}{
\uIf{sparse\_index[(b\_idx + l) * M + tid]$!= 0$}{
out\_grad[(b\_idx + l) * M * 4 + tid*4] = 0  \tcp*{reset all gradients to zero}
sparse\_index[(b\_idx + l) * M + tid] = 0 \;
}
}
}
sync\_threads()\;
\uIf{tid == 0}{
\For{\forloop{l}{0}{nbpb}}{
max\_d = best\_d[l*32]\;
\For{\forloop{i}{1}{(blockid+31 )/ 32}}{
\uIf{w\_max\_d[l * 32 + i].d $>$ max\_d}
{
max\_d = w\_max\_d[l * 32 + i]\tcp*{Find the largest distance across different warps}
}
}
\uIf{max\_d.d $!= 0$}
{
write\_distance\_gradient(max\_d, b\_robot\_spheres[b\_idx * M * 4], sparse\_idx[b\_idx * M])\;
}
}
}
\end{algorithm}

\begin{algorithm}
\DontPrintSemicolon
\algokw
\caption{World Collision Distance}\label{alg:collision-kernel}
\SetKwInput{Input}{Input}
\SetKwInput{Output}{Output}
\SetKwInput{WorldData}{World Model Input}
\SetKwInput{CollData}{Collision Config Input}
\SetKwInput{Kernel}{Kernel Launch Data}
\SetKwInput{Data}{Data}
\Kernel{Launch 1 thread per sphere}
\WorldData{obb\_bounds, obb\_pose, obb\_enable, max\_nobs, nboxes}
\CollData{activation\_distance, weight}
\Input{b\_robot\_spheres, env\_idx, B, H, M}
\Output{out\_distance, out\_grad}
\Data{sparsity\_idx}

bid = tid / (H, M)\;
hid = (tid - bid * H * M) / M\;
sid = (tid - bid * H * M - hid * M)\;
sph\_idx = bid * H * M + hid * M + sid \tcp*{compute ids from thread indices}
sph = b\_robot\_spheres[sph\_idx] \tcp*{read sphere from global memory}

\uIf{sph.radius $ <$ 0.0}
{
return \tcp*{we use negative sphere radius to deactivate spheres (e.g., spheres for a grasped object)}
}
max\_dist = 0\;
sum\_grad = 0\;

eta = activation\_distance\;
sph.radius += eta \tcp*{add activation distance to sphere radius}
start\_box\_idx = env\_idx *  max\_nobs\;
\For(\tcp*[f]{loop over obstacles}){\forloop{box\_idx}{0}{nboxes}}{
\uIf{obb\_enable[start\_box\_idx + box\_idx] == 0}{
continue \tcp*{check if obstacle is enabled}
}
loc\_obb\_pose = obb\_pose[start\_box\_idx + box\_idx] \tcp*{read obstacle pose into register}
loc\_sph = transform\_sphere(loc\_obb\_pose, sph)\tcp*{transform sphere from world frame to obstacle frame}
loc\_bounds = obb\_bounds[start\_box\_idx + box\_idx] \tcp*{read obstacle data}
loc\_bounds = loc\_bounds / 2\;
\uIf(\tcp*[f]{check if sphere collides with obstacle}){\upshape check\_sphere\_aabb(loc\_bounds, loc\_sphere)}{
loc\_bounds += loc\_sphere.radius \;
cl = compute\_sphere\_gradient(loc\_bounds, loc\_sphere, eta)\;
max\_dist += cl.distance\;
sum\_grad += project\_gradient\_global\_frame(loc\_obb\_pose, cl)\;
}
}
\If{max\_dist == 0}{
\uIf{sparsity\_idx[sph\_idx] == 0}{
return \;
}
sparsity\_idx[sph\_idx] = 0\;
out\_grad[sph\_idx * 4] = 0\;
out\_distance[sph\_idx] = 0\;
}
max\_dist = weight * max\_dist\;
sum\_grad = weight * sum\_grad\;
out\_distance[sph\_idx] = max\_dist\;
out\_grad[sph\_idx * 4] = sum\_grad\;
sparsity\_idx[sph\_idx] = 1\;
\end{algorithm}

\begin{algorithm}
\DontPrintSemicolon
\algokw
\caption{World Continuous Collision Distance}\label{alg:continuous-collision}
\SetKwInput{Input}{Input}
\SetKwInput{Output}{Output}
\SetKwInput{WorldData}{World Model Input}
\SetKwInput{CollData}{Collision Config Input}
\SetKwInput{Kernel}{Kernel Launch Data}
\SetKwInput{Data}{Data}
\Kernel{Launch 1 thread per sphere}
\WorldData{obb\_bounds, obb\_pose, obb\_enable, max\_nobs, nboxes}
\CollData{activation\_distance, weight, steps, speed\_dt}
\Input{b\_robot\_spheres, env\_idx, B, H, M}
\Output{out\_distance, out\_grad}
\Data{sparsity\_idx}
bid = tid / (H, M)\;
hid = (tid - bid * H * M) / M\;
sid = (tid - bid * H * M - hid * M) \tcp*{compute ids from thread indices}
sph1 = b\_robot\_spheres[(b\_addrs + (hid * M ) + sid) * 4] \tcp*{read sphere from global memory}
\uIf{sph1.radius $ <$ 0.0}
{
return \tcp*{we use negative sphere radius to deactivate spheres (e.g., spheres for a grasped object)}
}

max\_dist = 0\;
sum\_grad = 0\;
sweep\_fwd = False\;
sweep\_bwd = False\;
eta = activation\_distance\;
dt = speed\_dt\;
start\_box\_idx = env\_idx *  max\_nobs\;
sph1.radius += eta\;
\uIf{hid $>$ 0}
{
sph0 = b\_robot\_spheres[(b\_addrs + ((hid-1) * M ) + sid) * 4]\;
sph0.radius += eta\;
sph0\_distance = sphere\_distance(sph0, sph1)\;
sph0\_len = sph0\_distance + sph0.radius * 2\;
\uIf(\tcp*[f]{read sphere position in previous time-step}){sph0\_distance $>$ 0.0}{
sweep\_bwd = True\;
}
}
\uIf(\tcp*[f]{read sphere position in next time-step}){hid $<$ horizon -1}
{
sph2 = b\_robot\_spheres[(b\_addrs + ((hid+1) * M ) + sid) * 4]\;
sph2.radius += eta\;
sph2\_distance = sphere\_distance(sph2, sph1)\;
sph2\_len = sph2\_distance + sph2.radius * 2\;
\uIf{sph2\_distance $>$ 0.0}{
sweep\_fwd = True\;
}
}
\tcc{Perform continuous collision computation using Algorithm 12}
max\_dist, sum\_grad = compute\_continuous\_collision\_distance()\;
\uIf(\tcp*[f]{check if collision cost is zero}){max\_dist == 0}{
\uIf{sparsity\_idx[sph\_idx] == 0}{
return \tcp*{use sparsity data to exit early if tensors are already zero}
}
sparsity\_idx[sph\_idx] = 0\;
out\_grad[sph\_idx * 4] = 0\;
out\_distance[sph\_idx] = 0\;
}
max\_dist = weight * max\_dist\;
sum\_grad = weight * sum\_grad\;
out\_distance[sph\_idx] = max\_dist\;
out\_grad[sph\_idx * 4] = sum\_grad\;
sparsity\_idx[sph\_idx] = 1\;
\end{algorithm}

\begin{algorithm}
\caption{Continuous Collision Distance}\label{alg:continuous-collision-distance}
\For(\tcp*[f]{loop over obstacles}){\forloop{box\_idx}{0}{nboxes}}{
\uIf{obb\_enable[start\_box\_idx + box\_idx] == 0}{
continue \tcp*{check if obstacle is enabled}
}
in\_obb\_pose = obb\_pose[start\_box\_idx + box\_idx] \tcp*{read obstacle pose}
loc\_sph = transform\_sphere(in\_obb\_pose, sph)\tcp*{transform sphere from world frame to obstacle frame}
loc\_bounds = obb\_bounds[start\_box\_idx + box\_idx]\;
loc\_bounds = loc\_bounds / 2\;

\eIf(\tcp*[f]{check if sphere collides with obstacle}){\upshape check\_sphere\_aabb(loc\_bounds, loc\_sphere)}{
loc\_bounds += loc\_sphere.radius\;
cl = compute\_sphere\_gradient(loc\_bounds, loc\_sphere, eta)\;
max\_dist += cl.distance\;
sum\_grad += project\_gradient\_global\_frame(in\_obb\_pose, cl)\;
jump\_distance = sph1.radius \tcp*{start with a jump distance of sphere radius}
}
{
jump\_distance = compute\_distance(loc\_bounds, loc\_sphere) \tcp*{compute distance to obstacle}
}
jump\_d = jump\_distance\;
\If{sweep\_bwd \&\& jump\_d $<$ sph0\_distance}
{
loc\_sph0 = transform\_sphere(in\_obb\_pose, sph0)\;
\For(\tcp*[f]{loop over sweep steps}){\forloop{j}{0}{steps}}{
\uIf{jump\_d $\geq$ sph0\_distance}
{
break \tcp*{jump distance is greater than half distance between current and previous time-steps}
}
k0 = 1 - jump\_d / (sph0\_len)\;
compute\_jump\_distance(loc\_sph, loc\_sph0, k0, eta, loc\_bounds,
grad\_loc\_bounds, sum\_pt, jump\_d)\;
}
}
\If{sweep\_fwd \&\& jump\_d $<$ sph2\_distance}
{
loc\_sph2 = transform\_sphere(in\_obb\_pose, sph2)\;
\For{\forloop{j}{0}{steps}}{
\uIf{jump\_d $\geq$ sph2\_distance}
{
break \tcp*{jump distance is greater than half distance between current and next time-steps}
}
k0 = 1 - jump\_d / (sph2\_len)\;
compute\_jump\_distance(loc\_sph, loc\_sph2, k0, eta, loc\_bounds,
grad\_loc\_bounds, sum\_pt, jump\_d)\;
}
}
\If{sum\_pt.w $>$ 0}
{
max\_dist += sum\_pt.w\;
project\_gradient\_global\_frame(in\_obb\_mat, sum\_pt, max\_grad)\;
}
}
\end{algorithm}

\clearpage
\section{Changes since ICRA 2023}
We have made significant improvements to this work since \emph{cuRobo}'s publication at ICRA 2023~\cite{curobo_icra23}. We added jerk minimization to our trajectory optimization as we found large jerks to trigger safety stops on the real robot and also lead to worse tracking. We also switched from using backward difference to five point stencil difference for computing derivatives of state from position. This increased the speed at which we could move the UR10, enabling acceleration, and velocity to reach the robot's limits. In addition, we now implicitly account for the zero velocity, acceleration, and jerk at the final timestep and start by duplicating states. This removed overshoot that often happens when using a central difference scheme. All these changes to our state representation reduced the number of iterations needed to converge by 50\%.

We found that our evaluation dataset had goals that were not very close to the start state of the robot. This led our weights to work only for medium and long range motions. To make our approach work for small motions, we had to increase the weights for smoothness cost terms. We were unable to find a set of weights that would work for any length trajectory. To overcome this problem, we added a time-step optimization step that will re-optimize the trajectory with a dt estimate from an initial trajectory optimization. We also made changes to the Tesseract planner to closely match the trajectory optimization problem we are trying to solve. Specifically, we found that Tesseract did not optimize for trajectories that stop at the last timestep. We hence added velocity constraints to Tesseract's trajopt implementation, which enabled us to obtain a cuRobo-like trajectory profile as seen in Fig.~\ref{fig:trajectory_profile_plot}-(b). This increased the planning time for Tesseract to 5.8 seconds on a desktop PC with an i7-7800x. We upgraded our desktop PC with a more recent CPU, an AMD Ryzen 9 7950x which reduced Tesseract's planning time to 2.9 seconds. The upgraded CPU also reduced our geometric planning time to 20ms. We introduced a procedure to tune weights in Appendix~\ref{app:traj-tune} which allowed us to solve collision-free IK with 30 IK seeds instead of 100. This retuning of weights also enabled us to improve on position and rotation accuracy. 

We also improved on our CUDA kernels, implementing higher performing versions of the algorithms. We reduced memory latency by packing vectors with 3 floats into padded float4, enabling our kernels to use vectorized loads~(\href{https://developer.nvidia.com/blog/cuda-pro-tip-increase-performance-with-vectorized-memory-access/}{cuda-blog}).

\section{Author Contributions}
{
\begin{description}[leftmargin=0cm,style=unboxed, labelindent=0cm]
   \item[Balakumar Sundaralingam] led the overall project, designed, and developed the \emph{cuRobo} library. He ran evaluations in the paper, including real world experiments and wrote the paper.
   \item[Siva Kumar Sastry Hari] led the effort on accelerating compute bottlenecks with high performance CUDA kernels, contributed to \emph{cuRobo} library, and provided insights on GPU acceleration that led to design changes in \emph{cuRobo} library. He also wrote parts of the paper.   
    \item[Adam Fishman] implemented the evaluator, metrics, and provided a format for loading datasets. He also evaluated the \emph{pybullet-RRTConnect} and \emph{pybullet-AITStar} baselines.
    \item[Caelan Garett] provided insights on geometric planning and implemented an API to post process path plans from pddlstream using cuRobo.
     \item[Karl Van Wyk] provided initial implementations of Jacobian computations, discussed cost shaping heuristics to improve trajectory optimization, especially when handling constraints and increasing pose reaching accuracy.
    \item[Valks Blukis] implemented the nvblox pyTorch wrapper library and also developed an interface for rendering depth images from ground truth world representations.
    \item[Alexander Millane and Helen Oleynikova] implemented the signed distance and closest point functions in nvblox and exposed CUDA kernels for easy use within cuRobo.
    \item[Ankur Handa] advised on using MPPI as a sampling-based solver, provided insights on stencil methods for smoothing higher order derivatives and edited the paper.
    \item[Fabio Ramos] provided insights on numerical optimization, discussed L-BFGS and sampling-based methods, and edited the paper.
    \item[Nathan Ratliff] helped set the research direction, advised on trajectory optimization techniques including collision cost shaping, provided implementation insights to efficiently compute derivatives, and wrote parts of the paper.  
    \item[Dieter Fox] advised on the project, helped set the research direction, and provided ideas for experiments. 
\end{description}
}
\section{Revision History}
For the most recent version of this report, see the PDF from \website. We track revisions made since the initial arxiv submission in this section.

\noindent \textbf{Arxiv v2}
\begin{itemize}
    \item Changed name from `CuRobo' to `cuRobo'.
    \item Added flow  chart illustration of collision checking algorithms (Figure~\ref{fig:collision_algorithm_flowchart}). 
    \item Fixed typo in Equation~\ref{eq:pose_cost_term}.
    \item Added citations~\cite{millane2023nvblox, task_constructor_moveit}.
\end{itemize}
\addcontentsline{toc}{section}{References}
\bibliography{references}

\begin{thebibliography}{96}
\providecommand{\natexlab}[1]{#1}
\providecommand{\url}[1]{#1}
\csname url@samestyle\endcsname
\providecommand{\newblock}{\relax}
\providecommand{\bibinfo}[2]{#2}
\providecommand{\BIBentrySTDinterwordspacing}{\spaceskip=0pt\relax}
\providecommand{\BIBentryALTinterwordstretchfactor}{4}
\providecommand{\BIBentryALTinterwordspacing}{\spaceskip=\fontdimen2\font plus
\BIBentryALTinterwordstretchfactor\fontdimen3\font minus \fontdimen4\font\relax}
\providecommand{\BIBforeignlanguage}[2]{{%
\expandafter\ifx\csname l@#1\endcsname\relax
\typeout{** WARNING: IEEEtranN.bst: No hyphenation pattern has been}%
\typeout{** loaded for the language `#1'. Using the pattern for}%
\typeout{** the default language instead.}%
\else
\language=\csname l@#1\endcsname
\fi
#2}}
\providecommand{\BIBdecl}{\relax}
\BIBdecl

\bibitem[Medeiros et~al.(2020)Medeiros, Jelavic, Bjelonic, Siegwart, Meggiolaro, and Hutter]{medeiros2020trajectory}
V.~S. Medeiros, E.~Jelavic, M.~Bjelonic, R.~Siegwart, M.~A. Meggiolaro, and M.~Hutter, ``Trajectory optimization for wheeled-legged quadrupedal robots driving in challenging terrain,'' \emph{IEEE Robotics and Automation Letters}, vol.~5, no.~3, pp. 4172--4179, 2020.

\bibitem[LaValle(1998)]{lavalle1998rapidly}
S.~LaValle, ``Rapidly-exploring random trees: A new tool for path planning,'' \emph{Research Report 9811}, 1998.

\bibitem[LaValle(2006)]{lavalle2006PlanningAlgorithms}
S.~M. LaValle, \emph{Planning Algorithms}.\hskip 1em plus 0.5em minus 0.4em\relax Cambridge, U.K.: Cambridge University Press, 2006, available at http://planning.cs.uiuc.edu/.

\bibitem[Kunz and Stilman(2012)]{kunz2012time}
T.~Kunz and M.~Stilman, ``Time-optimal trajectory generation for path following with bounded acceleration and velocity,'' \emph{Robotics: Science and Systems VIII}, pp. 1--8, 2012.

\bibitem[Kunz and Stilman(2014)]{kunz2014probabilistically}
------, ``Probabilistically complete kinodynamic planning for robot manipulators with acceleration limits,'' in \emph{2014 IEEE/RSJ International Conference on Intelligent Robots and Systems}.\hskip 1em plus 0.5em minus 0.4em\relax IEEE, 2014, pp. 3713--3719.

\bibitem[Toussaint(2009)]{toussaint2009trajopt}
M.~Toussaint, ``Robot trajectory optimization using approximate inference,'' in \emph{Proc{.} of the Int{.} Conf{.} on Machine Learning (ICML)}.\hskip 1em plus 0.5em minus 0.4em\relax ACM, 2009, pp. 1049--1056.

\bibitem[Ratliff et~al.(2009)Ratliff, Zucker, Bagnell, and Srinivasa]{ratliff2009chomp}
N.~Ratliff, M.~Zucker, J.~A. Bagnell, and S.~Srinivasa, ``Chomp: Gradient optimization techniques for efficient motion planning,'' in \emph{2009 IEEE International Conference on Robotics and Automation}.\hskip 1em plus 0.5em minus 0.4em\relax IEEE, 2009, pp. 489--494.

\bibitem[Kalakrishnan et~al.(2011)Kalakrishnan, Chitta, Theodorou, Pastor, and Schaal]{kalakrishnan2011stomp}
M.~Kalakrishnan, S.~Chitta, E.~Theodorou, P.~Pastor, and S.~Schaal, ``Stomp: Stochastic trajectory optimization for motion planning,'' in \emph{2011 IEEE international conference on robotics and automation}.\hskip 1em plus 0.5em minus 0.4em\relax IEEE, 2011, pp. 4569--4574.

\bibitem[Schulman et~al.(2014)Schulman, Duan, Ho, Lee, Awwal, Bradlow, Pan, Patil, Goldberg, and Abbeel]{schulman2014motion}
J.~Schulman, Y.~Duan, J.~Ho, A.~Lee, I.~Awwal, H.~Bradlow, J.~Pan, S.~Patil, K.~Goldberg, and P.~Abbeel, ``Motion planning with sequential convex optimization and convex collision checking,'' \emph{The International Journal of Robotics Research}, vol.~33, no.~9, pp. 1251--1270, 2014.

\bibitem[Toussaint(2014)]{toussaint2014komo}
M.~Toussaint, ``{KOMO}: {N}ewton methods for k-order {M}arkov constrained motion problems,'' e-Print arXiv:1407.0414, 2014.

\bibitem[Ratliff et~al.(2015)Ratliff, Toussaint, and Schaal]{ratliff2015riemo}
N.~Ratliff, M.~Toussaint, and S.~Schaal, ``Understanding the geometry of workspace obstacles in motion optimization,'' in \emph{Proc{.} of the IEEE Int{.} Conf{.} on Robotics and Automation (ICRA)}, 2015.

\bibitem[Mukadam et~al.(2018)Mukadam, Dong, Yan, Dellaert, and Boots]{mukadam2018continuous}
M.~Mukadam, J.~Dong, X.~Yan, F.~Dellaert, and B.~Boots, ``Continuous-time gaussian process motion planning via probabilistic inference,'' \emph{The International Journal of Robotics Research}, vol.~37, no.~11, pp. 1319--1340, 2018.

\bibitem[Posa and Tedrake(2013)]{posa2013direct}
M.~Posa and R.~Tedrake, ``Direct trajectory optimization of rigid body dynamical systems through contact,'' in \emph{Algorithmic foundations of robotics X}.\hskip 1em plus 0.5em minus 0.4em\relax Springer, 2013, pp. 527--542.

\bibitem[Toussaint(2015)]{toussaint2015logic}
M.~Toussaint, ``Logic-geometric programming: An optimization-based approach to combined task and motion planning,'' in \emph{Twenty-Fourth International Joint Conference on Artificial Intelligence}, 2015.

\bibitem[Apgar et~al.(2018)Apgar, Clary, Green, Fern, and Hurst]{apgar2018fast}
T.~Apgar, P.~Clary, K.~Green, A.~Fern, and J.~W. Hurst, ``Fast online trajectory optimization for the bipedal robot cassie.'' in \emph{Robotics: Science and Systems}, vol. 101, 2018, p.~14.

\bibitem[Sundaralingam and Hermans(2019)]{sundaralingam2019relaxed}
B.~Sundaralingam and T.~Hermans, ``Relaxed-rigidity constraints: kinematic trajectory optimization and collision avoidance for in-grasp manipulation,'' \emph{Autonomous Robots}, vol.~43, no.~2, pp. 469--483, 2019.

\bibitem[Ichnowski et~al.(2020)Ichnowski, Danielczuk, Xu, Satish, and Goldberg]{goldberg2020gomp}
J.~Ichnowski, M.~Danielczuk, J.~Xu, V.~Satish, and K.~Goldberg, ``Gomp: Grasp-optimized motion planning for bin picking,'' in \emph{2020 IEEE International Conference on Robotics and Automation (ICRA)}, 2020, pp. 5270--5277.

\bibitem[Bertsekas(2018)]{bertsekas2019RLAndOptControl}
D.~P. Bertsekas, \emph{Reinforcement Learning and Optimal Control}.\hskip 1em plus 0.5em minus 0.4em\relax Athena Scientific, 2018.

\bibitem[Hansen(2016)]{hansen2016cmaTutorial}
\BIBentryALTinterwordspacing
N.~Hansen, ``The {CMA} evolution strategy: {A} tutorial,'' \emph{CoRR}, vol. abs/1604.00772, 2016. [Online]. Available: \url{http://arxiv.org/abs/1604.00772}
\BIBentrySTDinterwordspacing

\bibitem[tes()]{tesseract}
``{Tesseract},'' https://github.com/tesseract-robotics/tesseract, accessed: 2022-09-05.

\bibitem[Zucker et~al.(2013)Zucker, Ratliff, Dragan, Pivtoraiko, Klingensmith, Dellin, Bagnell, and Srinivasa]{zucker2013chomp}
M.~Zucker, N.~Ratliff, A.~D. Dragan, M.~Pivtoraiko, M.~Klingensmith, C.~M. Dellin, J.~A. Bagnell, and S.~S. Srinivasa, ``Chomp: Covariant hamiltonian optimization for motion planning,'' \emph{The International Journal of Robotics Research}, vol.~32, no. 9-10, pp. 1164--1193, 2013.

\bibitem[Murray et~al.(2016)Murray, Floyd-Jones, Qi, Sorin, and Konidaris]{konidaris2016fpga}
S.~Murray, W.~Floyd-Jones, Y.~Qi, D.~Sorin, and G.~Konidaris, ``Robot motion planning on a chip,'' in \emph{Proceedings of Robotics: Science and Systems}, AnnArbor, Michigan, June 2016.

\bibitem[Beeson and Ames(2015)]{beeson2015trac}
P.~Beeson and B.~Ames, ``Trac-ik: An open-source library for improved solving of generic inverse kinematics,'' in \emph{2015 IEEE-RAS 15th International Conference on Humanoid Robots (Humanoids)}.\hskip 1em plus 0.5em minus 0.4em\relax IEEE, 2015, pp. 928--935.

\bibitem[Coumans and Bai(2016--2021)]{coumans2021}
E.~Coumans and Y.~Bai, ``Pybullet, a python module for physics simulation for games, robotics and machine learning,'' \url{http://pybullet.org}, 2016--2021.

\bibitem[Pan and Manocha(2012)]{pan2012gpu}
J.~Pan and D.~Manocha, ``Gpu-based parallel collision detection for fast motion planning,'' \emph{The International Journal of Robotics Research}, vol.~31, no.~2, pp. 187--200, 2012.

\bibitem[Pan et~al.(2012)Pan, Chitta, and Manocha]{pan2012fcl}
J.~Pan, S.~Chitta, and D.~Manocha, ``Fcl: A general purpose library for collision and proximity queries,'' in \emph{2012 IEEE International Conference on Robotics and Automation}.\hskip 1em plus 0.5em minus 0.4em\relax IEEE, 2012, pp. 3859--3866.

\bibitem[Montaut et~al.(2022)Montaut, Lidec, Petrík, Sivic, and Carpentier]{Montaut-RSS-22}
L.~Montaut, Q.~Lidec, V.~Petrík, J.~Sivic, and J.~Carpentier, ``{Collision Detection Accelerated: An Optimization Perspective},'' in \emph{Proceedings of Robotics: Science and Systems}, New York City, NY, USA, June 2022.

\bibitem[Tracy et~al.(2023)Tracy, Howell, and Manchester]{tracy2022differentiable}
K.~Tracy, T.~A. Howell, and Z.~Manchester, ``Differentiable collision detection for a set of convex primitives,'' in \emph{IEEE International Conference on Robotics and Automation (ICRA)}, 2023.

\bibitem[Greenspan and Burtnyk(1996)]{greenspan1996obstacle}
M.~Greenspan and N.~Burtnyk, ``Obstacle count independent real-time collision avoidance,'' in \emph{Proceedings of IEEE International Conference on Robotics and Automation}, vol.~2.\hskip 1em plus 0.5em minus 0.4em\relax IEEE, 1996, pp. 1073--1080.

\bibitem[Van~Wyk et~al.(2022)Van~Wyk, Xie, Li, Rana, Babich, Peele, Wan, Akinola, Sundaralingam, Fox, et~al.]{van2022geometric}
K.~Van~Wyk, M.~Xie, A.~Li, M.~A. Rana, B.~Babich, B.~Peele, Q.~Wan, I.~Akinola, B.~Sundaralingam, D.~Fox \emph{et~al.}, ``Geometric fabrics: Generalizing classical mechanics to capture the physics of behavior,'' \emph{IEEE Robotics and Automation Letters}, vol.~7, no.~2, pp. 3202--3209, 2022.

\bibitem[Redon et~al.(2005)Redon, Lin, Manocha, and Kim]{redonsweptcoll}
\BIBentryALTinterwordspacing
S.~Redon, M.~C. Lin, D.~Manocha, and Y.~J. Kim, ``{Fast Continuous Collision Detection for Articulated Models},'' \emph{Journal of Computing and Information Science in Engineering}, vol.~5, no.~2, pp. 126--137, 02 2005. [Online]. Available: \url{https://doi.org/10.1115/1.1884133}
\BIBentrySTDinterwordspacing

\bibitem[Kim et~al.(2003)Kim, Varadhan, Lin, and Manocha]{kim2003fast}
Y.~J. Kim, G.~Varadhan, M.~C. Lin, and D.~Manocha, ``Fast swept volume approximation of complex polyhedral models,'' in \emph{Proceedings of the eighth ACM symposium on Solid modeling and applications}, 2003, pp. 11--22.

\bibitem[Redon et~al.(2002)Redon, Kheddar, and Coquillart]{redon2002fast}
S.~Redon, A.~Kheddar, and S.~Coquillart, ``Fast continuous collision detection between rigid bodies,'' in \emph{Computer Graphics Forum}, vol.~21, no.~3, 2002, pp. 279--287.

\bibitem[der Merwe et~al.(2020)der Merwe, Lu, Sundaralingam, Matak, and Hermans]{vandermerwe-icra2020-reconstruction-grasping}
\BIBentryALTinterwordspacing
M.~V. der Merwe, Q.~Lu, B.~Sundaralingam, M.~Matak, and T.~Hermans, ``{Learning Continuous 3D Reconstructions for Geometrically Aware Grasping},'' in \emph{IEEE International Conference on Robotics and Automation (ICRA)}, 2020. [Online]. Available: \url{https://sites.google.com/view/reconstruction-grasp/home}
\BIBentrySTDinterwordspacing

\bibitem[Ortiz et~al.(2022)Ortiz, Clegg, Dong, Sucar, Novotny, Zollhoefer, and Mukadam]{Ortiz-RSS-22}
J.~Ortiz, A.~Clegg, J.~Dong, E.~Sucar, D.~Novotny, M.~Zollhoefer, and M.~Mukadam, ``{iSDF: Real-Time Neural Signed Distance Fields for Robot Perception},'' in \emph{Proceedings of Robotics: Science and Systems}, New York City, NY, USA, June 2022.

\bibitem[Oleynikova et~al.(2017)Oleynikova, Taylor, Fehr, Siegwart, and Nieto]{oleynikova2017voxblox}
H.~Oleynikova, Z.~Taylor, M.~Fehr, R.~Siegwart, and J.~Nieto, ``Voxblox: Incremental 3d euclidean signed distance fields for on-board mav planning,'' in \emph{IEEE/RSJ International Conference on Intelligent Robots and Systems (IROS)}, 2017.

\bibitem[Bruce(2006)]{bruce2006real}
J.~R. Bruce, ``Real-time motion planning and safe navigation in dynamic multi-robot environments,'' \emph{Ph. D. dissertation}, 2006.

\bibitem[Quinlan and Khatib(1993)]{quinlan1993elastic}
S.~Quinlan and O.~Khatib, ``Elastic bands: Connecting path planning and control,'' in \emph{[1993] Proceedings IEEE International Conference on Robotics and Automation}.\hskip 1em plus 0.5em minus 0.4em\relax IEEE, 1993, pp. 802--807.

\bibitem[Millane et~al.(2023)Millane, Oleynikova, Wirbel, Steiner, Ramasamy, Tingdahl, and Siegwart]{millane2023nvblox}
A.~Millane, H.~Oleynikova, E.~Wirbel, R.~Steiner, V.~Ramasamy, D.~Tingdahl, and R.~Siegwart, ``nvblox: Gpu-accelerated incremental signed distance field mapping,'' 2023.

\bibitem[Park et~al.(2019)Park, Florence, Straub, Newcombe, and Lovegrove]{park2019deepsdf}
J.~J. Park, P.~Florence, J.~Straub, R.~Newcombe, and S.~Lovegrove, ``Deepsdf: Learning continuous signed distance functions for shape representation,'' in \emph{Proceedings of the IEEE/CVF conference on computer vision and pattern recognition}, 2019, pp. 165--174.

\bibitem[Tang et~al.(2023{\natexlab{a}})Tang, Sundaralingam, Tremblay, Wen, Yuan, Tyree, Loop, Schwing, and Birchfield]{tang2022rgb}
Z.~Tang, B.~Sundaralingam, J.~Tremblay, B.~Wen, Y.~Yuan, S.~Tyree, C.~Loop, A.~Schwing, and S.~Birchfield, ``Rgb-only reconstruction of tabletop scenes for collision-free manipulator control,'' in \emph{2023 IEEE International Conference on Robotics and Automation}, 2023.

\bibitem[Lambert and Boots(2021)]{Lambert2021EntropyRM}
\BIBentryALTinterwordspacing
A.~Lambert and B.~Boots, ``Entropy regularized motion planning via stein variational inference,'' \emph{ArXiv}, vol. abs/2107.05146, 2021. [Online]. Available: \url{https://api.semanticscholar.org/CorpusID:235795323}
\BIBentrySTDinterwordspacing

\bibitem[Schmidt et~al.(2015)Schmidt, Newcombe, and Fox]{schmidt2015dart}
\BIBentryALTinterwordspacing
T.~Schmidt, R.~Newcombe, and D.~Fox, ``\BIBforeignlanguage{English}{{DART}: {D}ense {A}rticulated {R}eal-time {T}racking with consumer depth cameras},'' \emph{\BIBforeignlanguage{English}{Autonomous Robots}}, vol.~39, no.~3, pp. 239--258, 2015. [Online]. Available: \url{http://dx.doi.org/10.1007/s10514-015-9462-z}
\BIBentrySTDinterwordspacing

\bibitem[Dellaert(2021)]{dellaert2021factorGraphs}
F.~Dellaert, ``Factor graphs: Exploiting structure in robotics,'' \emph{Annual Review of Control; Robotics; and Autonomous Systems}, 2021.

\bibitem[Nocedal and Wright(1999)]{nocedal1999numerical}
J.~Nocedal and S.~J. Wright, \emph{Numerical optimization}.\hskip 1em plus 0.5em minus 0.4em\relax Springer, 1999.

\bibitem[Todorov and Li(2005)]{todorov2005iLQG}
E.~Todorov and W.~Li, ``A generalized iterative lqg method for locally-optimal feedback control of constrained nonlinear stochastic systems,'' in \emph{In proceedings of the American Control Conference}, vol.~1, 2005, pp. 300--306.

\bibitem[Welling and Teh(2011)]{Welling2011Lan}
M.~Welling and Y.~W. Teh, ``Bayesian learning via stochastic gradient langevin dynamics,'' in \emph{Proceedings of the 28th international conference on machine learning (ICML-11)}.\hskip 1em plus 0.5em minus 0.4em\relax Citeseer, 2011, pp. 681--688.

\bibitem[Ma et~al.(2015)Ma, Chen, and Fox]{Ma2015MCMC}
Y.-A. Ma, T.~Chen, and E.~Fox, ``A complete recipe for stochastic gradient mcmc,'' in \emph{Advances in Neural Information Processing Systems}, C.~Cortes, N.~Lawrence, D.~Lee, M.~Sugiyama, and R.~Garnett, Eds., vol.~28.\hskip 1em plus 0.5em minus 0.4em\relax Curran Associates, Inc., 2015.

\bibitem[Wagener et~al.(2019)Wagener, Cheng, Sacks, and Boots]{wagener2019online}
N.~Wagener, C.-A. Cheng, J.~Sacks, and B.~Boots, ``An online learning approach to model predictive control,'' \emph{arXiv preprint arXiv:1902.08967}, 2019.

\bibitem[Srinivasa et~al.(2016)Srinivasa, Johnson, Lee, Koval, Choudhury, King, Dellin, Harding, Butterworth, Velagapudi, et~al.]{srinivasa2016system}
S.~S. Srinivasa, A.~M. Johnson, G.~Lee, M.~C. Koval, S.~Choudhury, J.~E. King, C.~M. Dellin, M.~Harding, D.~T. Butterworth, P.~Velagapudi \emph{et~al.}, ``A system for multi-step mobile manipulation: Architecture, algorithms, and experiments,'' in \emph{International Symposium on Experimental Robotics}.\hskip 1em plus 0.5em minus 0.4em\relax Springer, 2016, pp. 254--265.

\bibitem[Gammell et~al.(2015)Gammell, Srinivasa, and Barfoot]{gammell2015batch}
J.~D. Gammell, S.~S. Srinivasa, and T.~D. Barfoot, ``Batch informed trees (bit): Sampling-based optimal planning via the heuristically guided search of implicit random geometric graphs,'' in \emph{2015 IEEE international conference on robotics and automation (ICRA)}.\hskip 1em plus 0.5em minus 0.4em\relax IEEE, 2015, pp. 3067--3074.

\bibitem[Chamzas et~al.(2021)Chamzas, Quintero-Pena, Kingston, Orthey, Rakita, Gleicher, Toussaint, and Kavraki]{chamzas2021motionbenchmaker}
C.~Chamzas, C.~Quintero-Pena, Z.~Kingston, A.~Orthey, D.~Rakita, M.~Gleicher, M.~Toussaint, and L.~E. Kavraki, ``Motionbenchmaker: A tool to generate and benchmark motion planning datasets,'' \emph{IEEE Robotics and Automation Letters}, vol.~7, no.~2, pp. 882--889, 2021.

\bibitem[Fishman et~al.(2022)Fishman, Murali, Eppner, Peele, Boots, and Fox]{fishman2022mpinets}
A.~Fishman, A.~Murali, C.~Eppner, B.~Peele, B.~Boots, and D.~Fox, ``Motion policy networks,'' in \emph{Proceedings of the 6th Conference on Robot Learning (CoRL)}, 2022.

\bibitem[Sucan et~al.(2012)Sucan, Moll, and Kavraki]{sucan2012open}
I.~A. Sucan, M.~Moll, and L.~E. Kavraki, ``The open motion planning library,'' \emph{IEEE Robotics \& Automation Magazine}, vol.~19, no.~4, pp. 72--82, 2012.

\bibitem[Kuffner and LaValle(2000)]{kuffner2000rrt}
J.~J. Kuffner and S.~M. LaValle, ``Rrt-connect: An efficient approach to single-query path planning,'' in \emph{Proceedings 2000 ICRA. Millennium Conference. IEEE International Conference on Robotics and Automation. Symposia Proceedings (Cat. No. 00CH37065)}, vol.~2.\hskip 1em plus 0.5em minus 0.4em\relax IEEE, 2000, pp. 995--1001.

\bibitem[Coleman et~al.(2014)Coleman, Sucan, Chitta, and Correll]{coleman2014reducing}
D.~Coleman, I.~Sucan, S.~Chitta, and N.~Correll, ``Reducing the barrier to entry of complex robotic software: a moveit! case study,'' \emph{arXiv preprint arXiv:1404.3785}, 2014.

\bibitem[Thomason* et~al.(2023)Thomason*, Kingston*, and Kavraki]{thomason2024vamp}
W.~Thomason*, Z.~Kingston*, and L.~E. Kavraki, ``Motions in microseconds via vectorized sampling-based planning,'' in \emph{arxiv}, 2023.

\bibitem[Starke et~al.(2018)Starke, Hendrich, and Zhang]{starke2018memetic}
S.~Starke, N.~Hendrich, and J.~Zhang, ``Memetic evolution for generic full-body inverse kinematics in robotics and animation,'' \emph{IEEE Transactions on Evolutionary Computation}, vol.~23, no.~3, pp. 406--420, 2018.

\bibitem[Bhardwaj et~al.(2021)Bhardwaj, Choudhury, Boots, and Srinivasa]{bhardwaj2021leveraging}
M.~Bhardwaj, S.~Choudhury, B.~Boots, and S.~Srinivasa, ``Leveraging experience in lazy search,'' \emph{Autonomous Robots}, vol.~45, no.~7, pp. 979--996, 2021.

\bibitem[Carpentier et~al.(2019)Carpentier, Saurel, Buondonno, Mirabel, Lamiraux, Stasse, and Mansard]{carpentier2019pinocchio}
J.~Carpentier, G.~Saurel, G.~Buondonno, J.~Mirabel, F.~Lamiraux, O.~Stasse, and N.~Mansard, ``The pinocchio c++ library -- a fast and flexible implementation of rigid body dynamics algorithms and their analytical derivatives,'' in \emph{IEEE International Symposium on System Integrations (SII)}, 2019.

\bibitem[Bhardwaj et~al.(2022)Bhardwaj, Sundaralingam, Mousavian, Ratliff, Fox, Ramos, and Boots]{storm2021}
\BIBentryALTinterwordspacing
M.~Bhardwaj, B.~Sundaralingam, A.~Mousavian, N.~D. Ratliff, D.~Fox, F.~Ramos, and B.~Boots, ``{STORM: An Integrated Framework for Fast Joint-Space Model-Predictive Control for Reactive Manipulation},'' in \emph{Proceedings of the 5th Conference on Robot Learning}, ser. Proceedings of Machine Learning Research, vol. 164.\hskip 1em plus 0.5em minus 0.4em\relax PMLR, 2022, pp. 750--759. [Online]. Available: \url{https://proceedings.mlr.press/v164/bhardwaj22a.html}
\BIBentrySTDinterwordspacing

\bibitem[Meier et~al.(2022)Meier, Wang, Sutanto, Lin, and Shah]{meier2022differentiable}
F.~Meier, A.~Wang, G.~Sutanto, Y.~Lin, and P.~Shah, ``Differentiable and learnable robot models,'' \emph{arXiv preprint arXiv:2202.11217}, 2022.

\bibitem[Makhal and Goins(2018)]{makhal2018reuleaux}
A.~Makhal and A.~K. Goins, ``Reuleaux: Robot base placement by reachability analysis,'' in \emph{2018 Second IEEE International Conference on Robotic Computing (IRC)}.\hskip 1em plus 0.5em minus 0.4em\relax IEEE, 2018, pp. 137--142.

\bibitem[Murali et~al.(2023)Murali, Mousavian, Eppner, Fishman, and Fox]{murali2023cabinet}
A.~Murali, A.~Mousavian, C.~Eppner, A.~Fishman, and D.~Fox, ``Cabinet: Scaling neural collision detection for object rearrangement with procedural scene generation,'' in \emph{International Conference on Robotics and Automation (ICRA)}, 2023.

\bibitem[Zhu et~al.(2021)Zhu, Tremblay, Birchfield, and Zhu]{zhu2021hierarchical}
Y.~Zhu, J.~Tremblay, S.~Birchfield, and Y.~Zhu, ``Hierarchical planning for long-horizon manipulation with geometric and symbolic scene graphs,'' in \emph{2021 IEEE International Conference on Robotics and Automation (ICRA)}.\hskip 1em plus 0.5em minus 0.4em\relax IEEE, 2021, pp. 6541--6548.

\bibitem[Garrett et~al.(2020)Garrett, Lozano-P{\'e}rez, and Kaelbling]{garrett2020pddlstream}
C.~R. Garrett, T.~Lozano-P{\'e}rez, and L.~P. Kaelbling, ``Pddlstream: Integrating symbolic planners and blackbox samplers via optimistic adaptive planning,'' in \emph{Proceedings of the International Conference on Automated Planning and Scheduling}, vol.~30, 2020, pp. 440--448.

\bibitem[{renishaw}()]{aksim}
{renishaw}, ``{Universal Robots Joint Encoder},'' \url{https://www.renishaw.com/en/aksim-supports-universal-robots-for-smart-factory-automation--40389}, accessed: 2023-08-23.

\bibitem[{NVIDIA}(2022{\natexlab{a}})]{nvblox}
{NVIDIA}, ``{GitHub - nvidia-isaac/nvblox: A GPU-accelerated TSDF and ESDF library for robots equipped with RGB-D cameras},'' \url{https://github.com/nvidia-isaac/nvblox}, 2022, accessed: 2022-09-14.

\bibitem[Berenson et~al.(2011)Berenson, Srinivasa, and Kuffner]{berenson2011task}
D.~Berenson, S.~Srinivasa, and J.~Kuffner, ``Task space regions: A framework for pose-constrained manipulation planning,'' \emph{The International Journal of Robotics Research}, vol.~30, no.~12, pp. 1435--1460, 2011.

\bibitem[Alatartsev et~al.(2015)Alatartsev, Stellmacher, and Ortmeier]{alatartsev2015robotic}
S.~Alatartsev, S.~Stellmacher, and F.~Ortmeier, ``Robotic task sequencing problem: A survey,'' \emph{Journal of intelligent \& robotic systems}, vol.~80, pp. 279--298, 2015.

\bibitem[Gentilini et~al.(2013)Gentilini, Margot, and Shimada]{gentilini2013travelling}
I.~Gentilini, F.~Margot, and K.~Shimada, ``The travelling salesman problem with neighbourhoods: Minlp solution,'' \emph{Optimization Methods and Software}, vol.~28, no.~2, pp. 364--378, 2013.

\bibitem[Garrett et~al.(2021)Garrett, Chitnis, Holladay, Kim, Silver, Kaelbling, and Lozano-P{\'e}rez]{garrett2021integrated}
C.~R. Garrett, R.~Chitnis, R.~Holladay, B.~Kim, T.~Silver, L.~P. Kaelbling, and T.~Lozano-P{\'e}rez, ``Integrated task and motion planning,'' \emph{Annual review of control, robotics, and autonomous systems}, vol.~4, pp. 265--293, 2021.

\bibitem[Görner et~al.(2019)Görner, Haschke, Ritter, and Zhang]{task_constructor_moveit}
M.~Görner, R.~Haschke, H.~Ritter, and J.~Zhang, ``Moveit! task constructor for task-level motion planning,'' in \emph{2019 International Conference on Robotics and Automation (ICRA)}, 2019, pp. 190--196.

\bibitem[Manchester and Kuindersma(2020)]{manchester2020variational}
Z.~Manchester and S.~Kuindersma, ``Variational contact-implicit trajectory optimization,'' in \emph{Robotics Research: The 18th International Symposium ISRR}.\hskip 1em plus 0.5em minus 0.4em\relax Springer, 2020, pp. 985--1000.

\bibitem[Plancher et~al.(2022)Plancher, Neuman, Ghosal, Kuindersma, and Reddi]{plancher2022grid}
B.~Plancher, S.~M. Neuman, R.~Ghosal, S.~Kuindersma, and V.~J. Reddi, ``Grid: Gpu-accelerated rigid body dynamics with analytical gradients,'' in \emph{2022 International Conference on Robotics and Automation (ICRA)}.\hskip 1em plus 0.5em minus 0.4em\relax IEEE, 2022, pp. 6253--6260.

\bibitem[Plancher and Kuindersma(2020)]{plancher2020performance}
B.~Plancher and S.~Kuindersma, ``A performance analysis of parallel differential dynamic programming on a gpu,'' in \emph{Algorithmic Foundations of Robotics XIII: Proceedings of the 13th Workshop on the Algorithmic Foundations of Robotics 13}.\hskip 1em plus 0.5em minus 0.4em\relax Springer, 2020, pp. 656--672.

\bibitem[Plancher and Kuindersma(2019)]{plancher2019realtime}
------, ``Realtime model predictive control using parallel ddp on a gpu,'' in \emph{Toward Online Optimal Control of Dynamic Robots Workshop at the 2019 International Conference on Robotics and Automation (ICRA), Montreal, Canada}, 2019.

\bibitem[Bambade et~al.(2022)Bambade, El-Kazdadi, Taylor, and Carpentier]{Bambade-RSS-22}
A.~Bambade, S.~El-Kazdadi, A.~Taylor, and J.~Carpentier, ``{PROX-QP: Yet another Quadratic Programming Solver for Robotics and beyond},'' in \emph{Proceedings of Robotics: Science and Systems}, New York City, NY, USA, June 2022.

\bibitem[Howell et~al.(2022)Howell, Tracy, Le~Cleac’h, and Manchester]{howell2022calipso}
T.~A. Howell, K.~Tracy, S.~Le~Cleac’h, and Z.~Manchester, ``Calipso: A differentiable solver for trajectory optimization with conic and complementarity constraints,'' in \emph{The International Symposium of Robotics Research}.\hskip 1em plus 0.5em minus 0.4em\relax Springer, 2022, pp. 504--521.

\bibitem[Schubiger et~al.(2020)Schubiger, Banjac, and Lygeros]{schubiger2020gpu}
M.~Schubiger, G.~Banjac, and J.~Lygeros, ``Gpu acceleration of admm for large-scale quadratic programming,'' \emph{Journal of Parallel and Distributed Computing}, vol. 144, pp. 55--67, 2020.

\bibitem[Pineda et~al.(2022)Pineda, Fan, Monge, Venkataraman, Sodhi, Chen, Ortiz, DeTone, Wang, Anderson, Dong, Amos, and Mukadam]{pineda2022theseus}
L.~Pineda, T.~Fan, M.~Monge, S.~Venkataraman, P.~Sodhi, R.~Chen, J.~Ortiz, D.~DeTone, A.~Wang, S.~Anderson, J.~Dong, B.~Amos, and M.~Mukadam, ``{Theseus: A Library for Differentiable Nonlinear Optimization},'' \emph{arXiv preprint arXiv:2207.09442}, 2022.

\bibitem[Yamada et~al.(2023)Yamada, Hung, Collins, Havoutis, and Posner]{yamada2023leveraging}
J.~Yamada, C.-M. Hung, J.~Collins, I.~Havoutis, and I.~Posner, ``Leveraging scene embeddings for gradient-based motion planning in latent space,'' in \emph{International Conference on Robotics and Automation (ICRA)}, 2023.

\bibitem[Shah et~al.(2023)Shah, Yang, and Aamodt]{shahisca2023}
D.~Shah, N.~Yang, and T.~M. Aamodt, ``Energy-efficient realtime motion planning,'' in \emph{International Symposium on Computer Architecture}, 2023.

\bibitem[Hsiao et~al.(2023)Hsiao, Hari, Sundaralingam, Yik, Tambe, Sakr, Keckler, and Reddi]{vapriros2023}
Y.-S. Hsiao, S.~Hari, B.~Sundaralingam, J.~Yik, T.~Tambe, C.~Sakr, S.~Keckler, and V.~Reddi, ``Vapr: Variable-precision tensors to accelerate robot motion planning,'' in \emph{IEEE/RSJ International Conference on Intelligent Robots and Systems (IROS)}, 2023.

\bibitem[Ichnowski and Alterovitz(2019)]{ichnowski2019motion}
J.~Ichnowski and R.~Alterovitz, ``Motion planning templates: A motion planning framework for robots with low-power cpus,'' in \emph{2019 International Conference on Robotics and Automation (ICRA)}.\hskip 1em plus 0.5em minus 0.4em\relax IEEE, 2019, pp. 612--618.

\bibitem[Wedel et~al.(2009)Wedel, Pock, Zach, Bischof, and Cremers]{wedel2009improved}
A.~Wedel, T.~Pock, C.~Zach, H.~Bischof, and D.~Cremers, ``An improved algorithm for tv-l 1 optical flow,'' in \emph{Statistical and Geometrical Approaches to Visual Motion Analysis: International Dagstuhl Seminar, Dagstuhl Castle, Germany, July 13-18, 2008. Revised Papers}.\hskip 1em plus 0.5em minus 0.4em\relax Springer, 2009, pp. 23--45.

\bibitem[Berscheid and Kr{\"o}ger(2021)]{berscheid2021jerk}
L.~Berscheid and T.~Kr{\"o}ger, ``Jerk-limited real-time trajectory generation with arbitrary target states,'' \emph{Robotics: Science and Systems XVII}, 2021.

\bibitem[Strub and Gammell(2020)]{strub2020adaptively}
M.~P. Strub and J.~D. Gammell, ``Adaptively informed trees (ait*): Fast asymptotically optimal path planning through adaptive heuristics,'' in \emph{2020 IEEE International Conference on Robotics and Automation (ICRA)}.\hskip 1em plus 0.5em minus 0.4em\relax IEEE, 2020, pp. 3191--3198.

\bibitem[{NVIDIA}(2022{\natexlab{b}})]{warp}
{NVIDIA}, ``{GitHub - NVIDIA/warp: A Python framework for high performance GPU simulation and graphics},'' \url{https://github.com/NVIDIA/warp}, 2022, accessed: 2022-09-14.

\bibitem[Fuji~Tsang et~al.(2022)Fuji~Tsang, Shugrina, Lafleche, Takikawa, Wang, Loop, Chen, Jatavallabhula, Smith, Rozantsev, Perel, Shen, Gao, Fidler, State, Gorski, Xiang, Li, Li, and Lebaredian]{KaolinLibrary}
C.~Fuji~Tsang, M.~Shugrina, J.~F. Lafleche, T.~Takikawa, J.~Wang, C.~Loop, W.~Chen, K.~M. Jatavallabhula, E.~Smith, A.~Rozantsev, O.~Perel, T.~Shen, J.~Gao, S.~Fidler, G.~State, J.~Gorski, T.~Xiang, J.~Li, M.~Li, and R.~Lebaredian, ``Kaolin: A pytorch library for accelerating 3d deep learning research,'' \url{https://github.com/NVIDIAGameWorks/kaolin}, 2022.

\bibitem[Wang et~al.(2023)Wang, Gao, Xu, Geng, Hu, Qiu, Li, Yang, Moon, Pandey, et~al.]{wang2023pypose}
C.~Wang, D.~Gao, K.~Xu, J.~Geng, Y.~Hu, Y.~Qiu, B.~Li, F.~Yang, B.~Moon, A.~Pandey \emph{et~al.}, ``Pypose: A library for robot learning with physics-based optimization,'' in \emph{Proceedings of the IEEE/CVF Conference on Computer Vision and Pattern Recognition}, 2023, pp. 22\,024--22\,034.

\bibitem[Tang et~al.(2023{\natexlab{b}})Tang, Lin, Akinola, Handa, Sukhatme, Ramos, Fox, and Narang]{tang2023industreal}
B.~Tang, M.~A. Lin, I.~Akinola, A.~Handa, G.~S. Sukhatme, F.~Ramos, D.~Fox, and Y.~Narang, ``Industreal: Transferring contact-rich assembly tasks from simulation to reality,'' \emph{arXiv preprint arXiv:2305.17110}, 2023.

\bibitem[Calli et~al.(2017)Calli, Singh, Bruce, Walsman, Konolige, Srinivasa, Abbeel, and Dollar]{calli2017yale}
B.~Calli, A.~Singh, J.~Bruce, A.~Walsman, K.~Konolige, S.~Srinivasa, P.~Abbeel, and A.~M. Dollar, ``Yale-cmu-berkeley dataset for robotic manipulation research,'' \emph{The International Journal of Robotics Research}, vol.~36, no.~3, pp. 261--268, 2017.

\bibitem[Dantam(2021)]{dual_quaternion}
\BIBentryALTinterwordspacing
N.~T. Dantam, ``Robust and efficient forward, differential, and inverse kinematics using dual quaternions,'' \emph{The International Journal of Robotics Research}, vol.~40, no. 10-11, pp. 1087--1105, 2021. [Online]. Available: \url{https://doi.org/10.1177/0278364920931948}
\BIBentrySTDinterwordspacing

\bibitem[{NVIDIA}(2022{\natexlab{c}})]{CUDAProgrammingGuide}
{NVIDIA}, ``{Programming Guide :: CUDA Toolkit Documentation},'' \url{https://docs.nvidia.com/cuda/cuda-c-programming-guide/index.html}, 2022, accessed: 2022-09-14.

\bibitem[Sundaralingam et~al.(2023)Sundaralingam, Hari, Fishman, Garrett, Van~Wyk, Blukis, Millane, Oleynikova, Handa, Ramos, Ratliff, and Fox]{curobo_icra23}
B.~Sundaralingam, S.~K.~S. Hari, A.~Fishman, C.~Garrett, K.~Van~Wyk, V.~Blukis, A.~Millane, H.~Oleynikova, A.~Handa, F.~Ramos, N.~Ratliff, and D.~Fox, ``Curobo: Parallelized collision-free robot motion generation,'' in \emph{2023 IEEE International Conference on Robotics and Automation (ICRA)}, 2023, pp. 8112--8119.

\end{thebibliography}

\end{document}